\documentclass{article}
\usepackage{ijcai25}

\usepackage{times}
\usepackage{soul}
\usepackage{url}
\usepackage[hidelinks]{hyperref}
\usepackage[utf8]{inputenc}
\usepackage[small]{caption}
\usepackage{graphicx}
\usepackage{amsmath}
\usepackage{amsthm}
\usepackage{booktabs}
\usepackage{algorithm}
\usepackage[switch]{lineno}

\usepackage{amsmath,amsfonts,bm}

\def\eqref#1{equation~\ref{#1}}

\def\1{\bm{1}}

\def\vc{{\bm{c}}}

\def\vm{{\bm{m}}}

\def\vw{{\bm{w}}}

\DeclareMathAlphabet{\mathsfit}{\encodingdefault}{\sfdefault}{m}{sl}
\SetMathAlphabet{\mathsfit}{bold}{\encodingdefault}{\sfdefault}{bx}{n}

\DeclareMathOperator*{\argmin}{arg\,min}

\usepackage{natbib}
\usepackage{amssymb}

\newtheorem{definition}{Definition} 
 
\newtheorem{assumption}{Assumption} 

\usepackage[noend]{algpseudocode}

\usepackage{subcaption}

\usepackage{enumitem}
\setlist[itemize]{align=parleft,left=10pt..1.6em}

\usepackage{multirow}

\usepackage{pifont}
\newcommand{\cmark}{\ding{51}}%
\newcommand{\xmark}{\ding{55}}%

\title{TinyProto: Communication-Efficient Federated Learning with Sparse Prototypes in Resource-Constrained Environments}

\author{
Gyuejeong Lee$^1$\And
Daeyoung Choi$^2$\\
\affiliations
$^1$SAKAK Inc.\\
$^2$The Cyber University of Korea\\
\emails
regulation.lee@sakak.co.kr,
choidy@cuk.edu
}

\begin{document}

\maketitle

\begin{abstract}
Communication efficiency in federated learning (FL) remains a critical challenge for resource-constrained environments. While prototype-based FL reduces communication overhead by sharing class prototypes—mean activations in the penultimate layer—instead of model parameters, its efficiency decreases with larger feature dimensions and class counts. We propose TinyProto, which addresses these limitations through Class-wise Prototype Sparsification (CPS) and adaptive prototype scaling. CPS enables structured sparsity by allocating specific dimensions to class prototypes and transmitting only non-zero elements, while adaptive scaling adjusts prototypes based on class distributions. Our experiments show TinyProto reduces communication costs by up to 4× compared to existing methods while maintaining performance. Beyond its communication efficiency, TinyProto offers crucial advantages: achieving compression without client-side computational overhead and supporting heterogeneous architectures, making it ideal for resource-constrained heterogeneous FL.
\end{abstract}

\section{Introduction}
Federated learning (FL) has emerged as a transformative framework for collaborative model training without the need for centralized data aggregation \citep{mcmahan2017communication}. It offers substantial advantages, including enhanced data privacy, optimized computational resource utilization, and improved communication efficiency \citep{zhang2021survey}. Among these benefits, reducing communication overhead is particularly critical for the scalability of FL \citep{sattler2019robust}. Traditional FL frameworks transmit model weights or gradients, rather than raw data, between the central server and clients, thereby conserving communication resources. However, as deep networks continue to grow in size, the communication overhead associated with transmitting model parameters increases significantly, undermining these advantages. This presents substantial challenges for large-scale deployments in resource-constrained environments, such as Internet of Things (IoT) networks \citep{nguyen2021federated}.

To address this challenge, various strategies have been proposed to optimize either communication frequency \citep{karimireddy2020scaffold, wang2019adaptivecomm, wang2019adaptive} or communication volume, with this work focusing specifically on the latter—reducing communication volume. Some approaches achieve this by transmitting only a subset of deep network parameters \citep{chen2021bridging, wang2024fullelevatingfederatedlearning} or gradients \citep{sattler2019robust, han2020adaptive}, while others utilize low-rank representations of parameter matrices \citep{wang2023svdfed, wu2024decoupling, wu2022communication}. Additionally, techniques such as pruning model weights have proven effective in mitigating communication overhead \citep{jiang2022model, zhang2022fedduap, wu2023efficient}.
In contrast to transmitting model parameters, alternative methods exchange knowledge between the server and clients by sharing logits \citep{jeong2018communication, li2019fedmd}, intermediate features \citep{tan2022fedproto, zhang2024fedtgp}, auxiliary networks \citep{shen2020federated}, or data generators \citep{zhu2021data}. Although these methods were initially developed to address challenges such as data and model heterogeneity, they also contribute to improving communication efficiency.

\begin{figure}[bt]
    \centering
    \includegraphics[width=0.990\columnwidth]{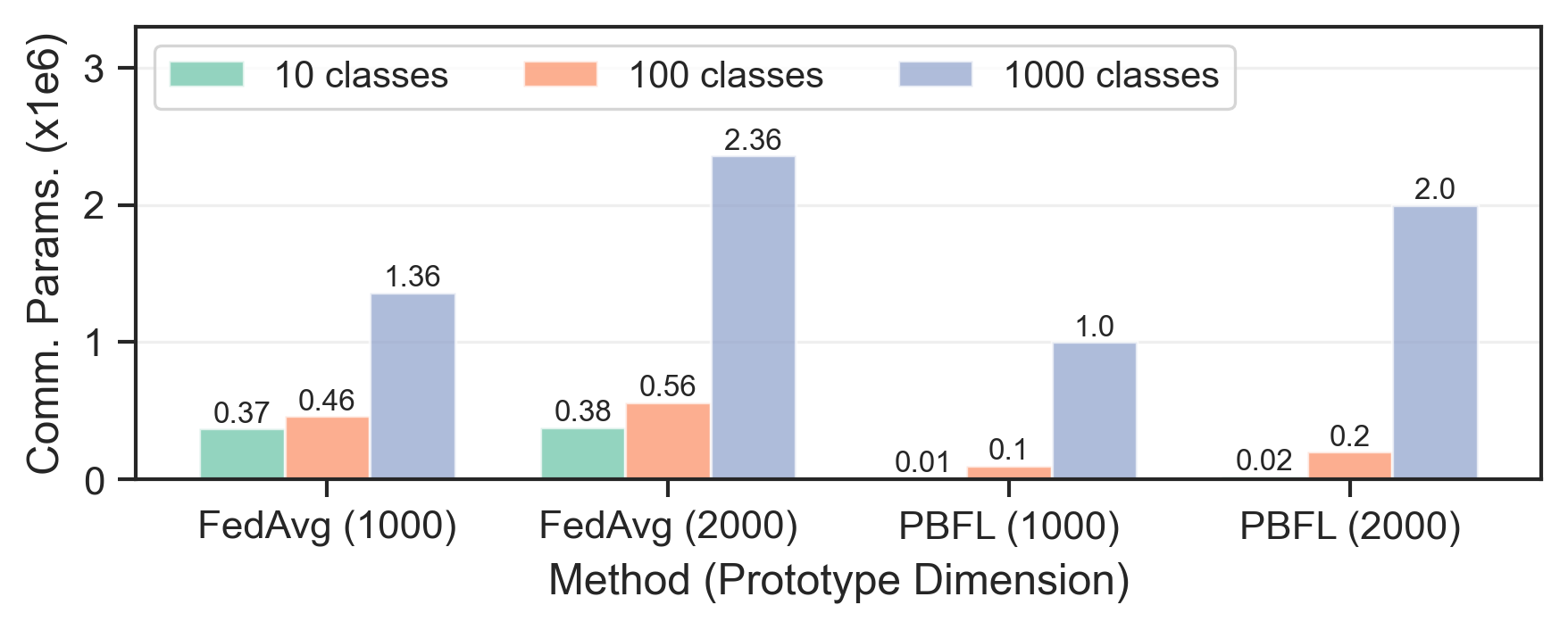}
    \vspace{-8pt}
    \caption{Comparison of communication costs between FedAvg and PBFL approaches for ShuffleNet v2, showing parameter counts across different numbers of classes and prototype dimensions.}
    \label{fig:comparison_comm_params}
    \vspace{-15pt}
\end{figure}

Among these approaches, prototype-based FL (PBFL) stands out for its communication costs being less sensitive to model size and its straightforward implementation, while providing natural privacy protection. PBFL's core mechanism involves sharing prototypes—represented as mean activations in the penultimate layer for each class—instead model parameters. While PBFL offers notable advantages in communication efficiency, our analysis reveals significant scalability limitations. Consider a classification task using ShuffleNet v2 (approximately 1.36M parameters for 1000 classes, with 1.0M parameters between the penultimate and output layers) as illustrated in Figure~\ref{fig:comparison_comm_params}. PBFL's communication costs, determined by the product of class count and prototype dimensions, are significantly lower than FedAvg \citep{mcmahan2017communication} when these factors are small. However, this advantage diminishes rapidly as either the number of classes or prototype dimensions increases, limiting PBFL's effectiveness in resource-constrained networks.

In this work, we address the challenge of mitigating the communication overhead caused by scalability issues in the PBFL approach, aiming to make it more practical for resource-constrained environments.  We propose TinyProto, a novel framework that combines prototype sparsification and prototype scaling. 
The Class-wise Prototype Sparsification (CPS) method reduces communication overhead by introducing structured sparsity, where each class prototype uses only specific dimensions while zeroing out the others. Beyond reducing communication costs, CPS improves classification performance by creating more discriminative prototypes through structured sparsification.
While PBFL methods like FedProto and FedTGP show promise for heterogeneous FL (HtFL), they often underperform compared to other HtFL approaches, making the direct application of CPS impractical. To address this limitation, we introduce a prototype scaling method that adjusts prototypes based on class-wise sample sizes.
By integrating CPS with prototype scaling, TinyProto achieves substantial communication efficiency without compromising model performance. Experimental results demonstrate that TinyProto reduces communication costs by up to 10× compared to the original PBFL approaches and 4× compared to the most efficient baselines, all while delivering superior performance. 

In summary, this work addresses a key challenge in PBFL: \textbf{reducing communication overhead while maintaining model performance}. We make three key contributions:
\begin{itemize}
\item We introduce CPS and adaptive prototype scaling—a novel method that achieves structured sparsity while enhancing classification through discriminative prototypes and class-wise adjustments.
\item We propose the TinyProto framework, integrating these methods to achieve up to 10× communication cost reduction while enhancing performance.
\item TinyProto distinguishes itself by eliminating client-side computation and fine-tuning requirements for model pruning while supporting heterogeneous architectures, making it particularly effective for resource-constrained, real-world deployments.
\end{itemize}

\section{Related Work}

\subsection{Heterogeneous Federated Learning}
HtFL addresses challenges arising from heterogeneity in data and client model architectures. Existing HtFL approaches can be broadly categorized based on the type of information exchanged: model parameters, auxiliary model or data generator parameters, and logits or intermediate features. To accommodate varying model architectures, LG-FedAvg~\citep{liang2020think} allows clients to share upper model layers while retaining architecture-specific lower layers.
Auxiliary model-based methods, such as FML~\citep{shen2020federated}, use mutual distillation~\citep{zhang2018deep} to train and share compact models. Similarly, sharing a data generator improves generalization~\citep{zhu2021data}.

Other approaches avoid sharing model parameters by using logits or intermediate features. FedMD~\citep{li2019fedmd}, for instance, uses a global public dataset to enable knowledge transfer among heterogeneous clients. FedDistill~\citep{jeong2018communication} transmits globally averaged class-wise logits from clients to the server, but this method risks exposing the number of classes and the logit distribution, leading to potential privacy concerns. To address these risks, FedProto~\citep{tan2022fedproto} takes a privacy-preserving approach by exchanging local prototypes of the penultimate layer.
Building on these advancements, recent works have refined PBFL techniques further. FedTGP~\citep{zhang2024fedtgp} enhances performance by employing adaptive-margin-enhanced contrastive learning (ACL), which improves prototype separability. Similarly, our work builds upon PBFL but focuses on enhancing communication efficiency by introducing sparsity in prototypes.

\subsection{Sparsity in Federated Learning}
Connection pruning \citep{han2015deep, han2015learning} and structured weight sparsification \citep{wen2016learning} have been widely studied in centralized deep learning to reduce memory usage and improve computational efficiency. These techniques have been adapted to federated learning to address communication constraints.
Gradient sparsification enhances communication efficiency by transmitting only the most significant updates. \cite{stich2018sparsified} demonstrate that k-sparsification in stochastic gradient descent can significantly reduce communication costs. Sparse Ternary Compression (STC) \citep{sattler2019robust} extends this top-k gradient sparsification for federated learning, while \cite{han2020adaptive} propose a fairness-aware method to ensure equitable updates across clients.
Model weight pruning offers another approach to reduce parameter transmission. PruneFL \citep{jiang2022model} employs adaptive pruning to reduce model size, while FedDUAP \citep{zhang2022fedduap} prunes based on layer dimensions and importance.
Several methods use masks to avoid transmitting unnecessary weights. \cite{wu2023efficient} employ masks to dynamically prune neurons or kernels, and FedMask \citep{li2021fedmask} transmits sparse binary masks instead of full parameters. However, these approaches often require additional client-side computation or support only homogeneous models.
Our framework also employs sparsification with masks but applies it to prototypes in PBFL, achieving lower communication costs while eliminating the need for extra client-side computation and supporting heterogeneous models.

\section{Problem Setting} \label{sec:problem}
We consider a system comprising $M$ clients and a server for a $K$ classes classification task. The clients interact with the server to jointly develop personalized models without sharing their private data directly. Each client $i$ has its data distribution $P_i$ with $K$ classes. These distributions can differ between clients, reflecting the typical scenario in FL. We define a loss function $\ell$ that evaluates the performance of each client's local model $\vw_i$ on data points from their respective distributions. 
The aim of the system can be described as minimizing the mean expected loss across all clients:
\begin{equation}
\min_{\mathbf{W}} \left\{ F(\mathbf{W}) := \frac{1}{M}\sum_{i=1}^{M}\mathbb{E}_{(x,y) \sim P_i}\left[\ell(\bm{w}_i;x,y)\right] \right\},
\end{equation}
where $\mathbf{W} = [\bm{w}_1, \bm{w}_2, ..., \bm{w}_M]$ represents a matrix containing all individual client models.
Given that we only have a limited set of data points, we estimate this expected loss using the empirical risk calculated on each client's local training dataset $\mathcal{D}_i = {(x_i^{(l)}, y_i^{(l)})}_{l=1}^{n_i}$, with its corresponding empirical distribution $\hat{P}_i$.
Thus, the training objective becomes finding the optimal set of local models that minimizes the average empirical risk across all clients:
\begin{equation}
\mathbf{W}^{*} = \arg\min_{\mathbf{W}}\frac{1}{M}\sum_{i=1}^M \mathcal{L}_i(\bm{w}_i).
\end{equation}
Here, $\mathcal{L}_i(\bm{w}_i) = \frac{1}{n_i} \sum_{l=1}^{n_i} \ell(\bm{w}_i;x_i^{(l)}, y_i^{(l)})$ represents the average loss for each client, calculated over their private data.

In this work, we split each client's deep network $\bm{w}_i$ into two components: representation layers (feature extractor) and a decision layer (classifier).
The feature extractor $f_i$ of client $i$, parameterized by $\bm{\theta}_i$, transforms input data from $\mathbb{R}^D$ into a feature space $\mathbb{R}^d$. For the sample $x$, it produces a feature vector $\vc=f_i(\bm{\theta}_i;x)$.
The classifier $g_i$, parameterized by $\bm{\phi}_i$, maps these features to the final output space $\mathbb{R}^K$.

\subsubsection{Prototype-Based Federated Learning} 
In PBFL, clients exchange prototypes (the mean of decision layer activations per class) rather than model parameters. Thus the key process handling prototypes consists of three steps. At the initial round, each client trains its model without any regularization before following these steps:

\vspace{0.5em}  
\noindent \textbf{Step 1} (Local Prototype Generation)\textbf{.}
    Each client generates local prototypes from its private dataset and sends them to the server. For each class $j$ on client $i$, the local prototype $\bar \vc_{i, j}^{L}$ is computed as:
\begin{equation}
    \bar \vc_{i,j}^{L} = \frac{1}{n_{i,j}} \sum_{(x,y) \in \mathcal{D}_{i,j}} f_i(\bm{\theta}_i;x),
    \label{eq:local_prototype_aggregation}
\end{equation}
where $n_{i,j} = |\mathcal{D}_{i,j}|$ is the number of samples from class $j$ on client $i$, and $\mathcal{D}_{i,j} \subseteq \mathcal{D}_i$ is the subset of client $i$'s local dataset containing samples from class $j$.

\vspace{0.5em}  
\noindent \textbf{Step 2} (Global Prototype Generation)\textbf{.}
    The server aggregates local prototypes to create global prototypes and sends them to the clients. For each class $j$, the global prototype $\bar{\vc}_{j}^{G}$ is computed using weighted averaging \citep{tan2022fedproto}:
\begin{equation}
\bar{\vc}_{j}^{G}=\frac{1}{\left|\mathcal{N}_j\right|} \sum_{i\in{\mathcal{N}_j}} \frac{n_{i,j}}{\sum_{i=1}^M n_{i,j}}\bar\vc_{i,j}^{L},
\label{eq:global_prototype_aggregation}
\end{equation}
where $\mathcal{N}_j$ is the set of clients with class $j$, and $\sum_{i=1}^M n_{i,j}$ is the total number of $j$-th class samples in the system.

\vspace{0.5em}  
\noindent \textbf{Step 3} (Local Model Training)\textbf{.}
Each client trains its model using a combined loss function:
\begin{equation}
\Tilde{\mathcal{L}}_i(\bm{w}_i) = \mathcal{L}_i(\bm{w}_i) + \lambda\mathcal{R}_i,
\label{eq:loss_function}
\end{equation}
where $\mathcal{R}_i$ is the regularization term for knowledge distillation, and $\lambda$ is a hyperparameter controlling regularization strength. The term $\mathcal{R}_i$ is formulated as:
\begin{equation}
\mathcal{R}_i = \sum_{j} \rho(\bar{\vc}_{i,j}^{L}, \bar{\vc}_{j}^{G}),
\label{eq:regularization_term}
\end{equation}
where the function $\rho(\cdot,\cdot)$ computes the Euclidean distance between the local and global prototypes.

This process iterates until convergence, with clients continually updating their local models while maintaining communication efficiency through prototype exchange.

\section{Methods}
This section presents two components of TinyProto and concludes with its integration algorithm for PBFL approaches.

\subsection{Prototype Sparsification for Communication Efficiency} 
\subsubsection{Motivation from Decision Layer Activation Analysis} 
\begin{figure}[t]
    \centering
    \begin{subfigure}[t]{0.495\columnwidth}
        \centering
        \includegraphics[width=\textwidth]{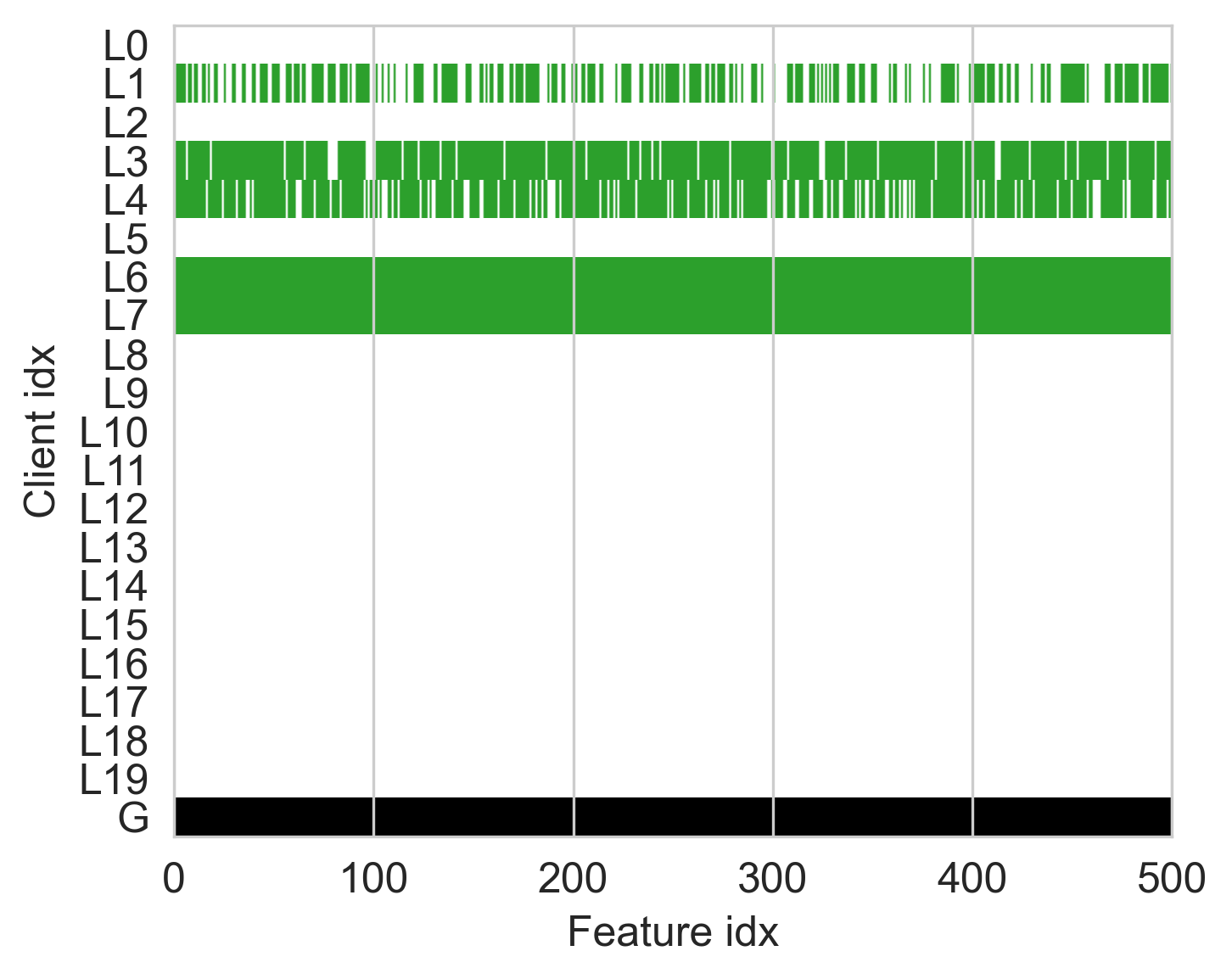}
        \caption{Class \#2's prototypes w/o CPS}
        \label{fig:heatmap_fedproto_without_cps_class2}
    \end{subfigure}
    \begin{subfigure}[t]{0.495\columnwidth}
        \centering
        \includegraphics[width=\textwidth]{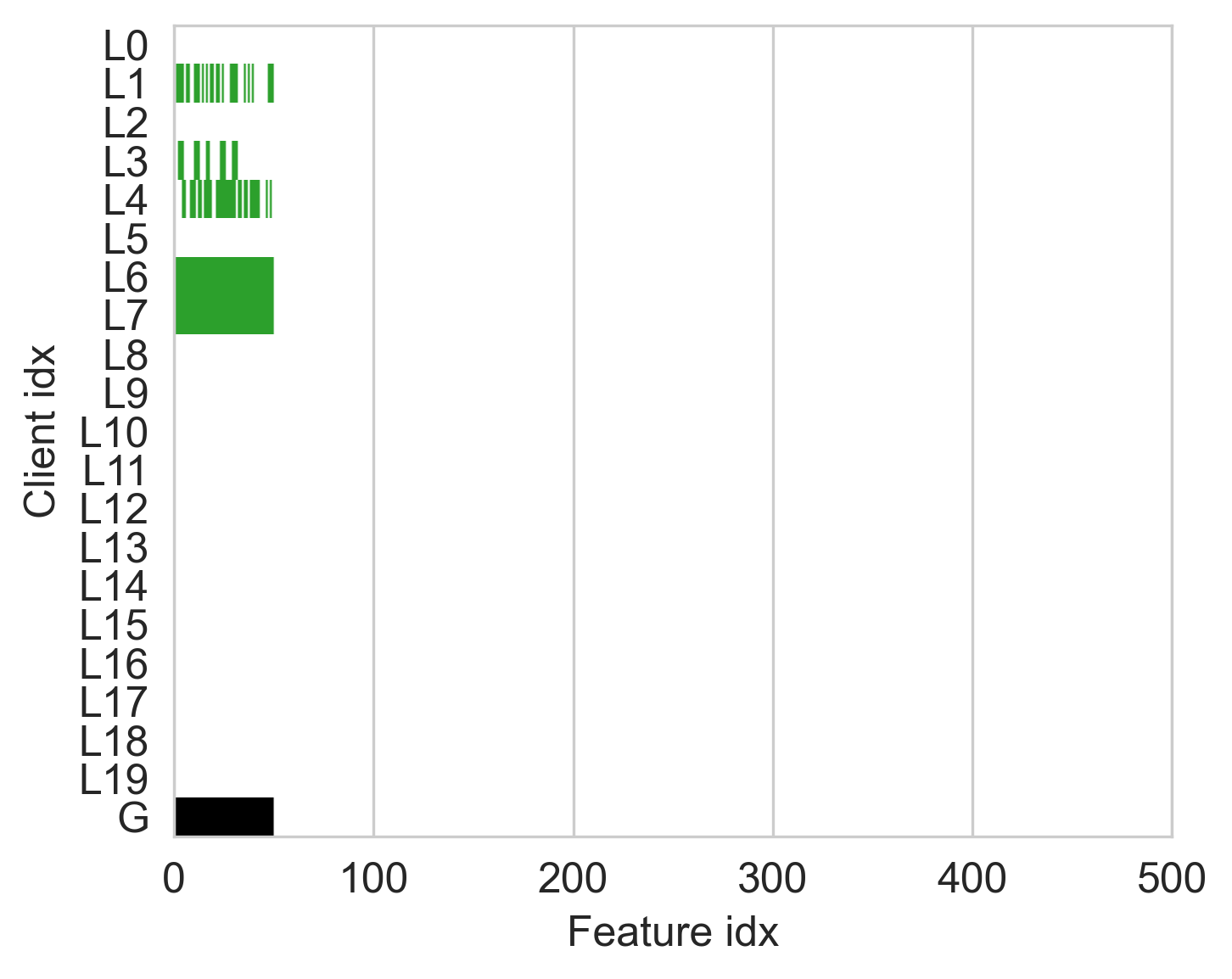}
        \caption{Class \#2's prototypes w/ CPS}
        \label{fig:heatmap_fedproto_with_cps_class2}
    \end{subfigure}
    
    \vspace{0.6em}
    
    \begin{subfigure}[t]{0.495\columnwidth}
        \centering
        \includegraphics[width=\textwidth]{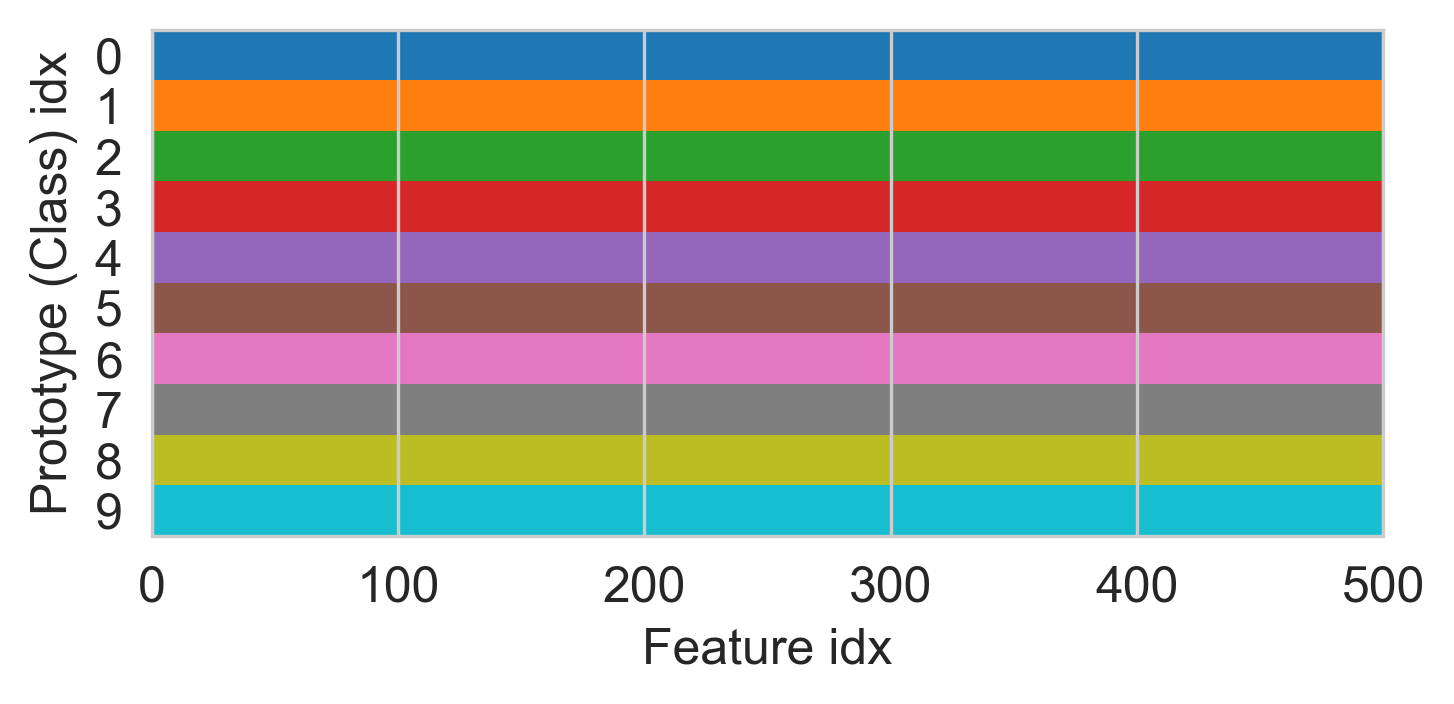}
        \caption{Global prototypes w/o CPS}
        \label{fig:heatmap_fedproto_without_cps_global}
    \end{subfigure}
    \begin{subfigure}[t]{0.495\columnwidth}
        \centering
        \includegraphics[width=\textwidth]{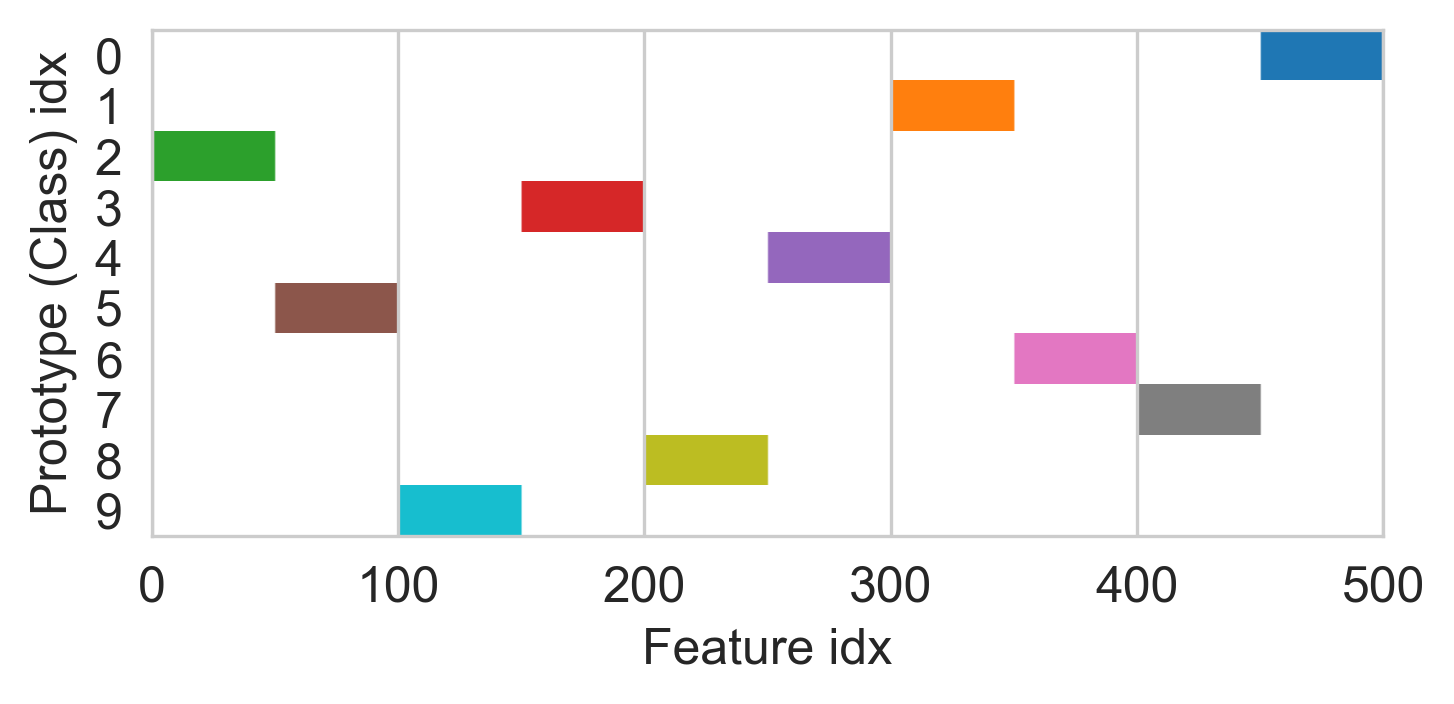}
        \caption{Global prototypes w/ CPS}
        \label{fig:heatmap_fedproto_with_cps_global}
    \end{subfigure}
    
    \caption{Prototype comparison of FedProto with and without CPS for the CIFAR-10 dataset. Each row in the heatmaps corresponds to a prototype, where colored cells indicate non-zero values. 
    The CPS dimension $s$ set to 50.}
    \label{fig:heatmaps_comparison}
    \vspace{-10pt}
\end{figure}

The decision layer frequently demonstrates sparse activation patterns when implementing the ReLU activation function, potentially resulting in dead units. A \textit{dead unit} refers to a hidden unit that outputs zero for all input patterns in the training set \citep{lu2019dying}, effectively not contributing to learning or inference. Our empirical analysis reveals that approximately 50\% of hidden units in the network's decision layer become dead units for each class.
Figure \ref{fig:heatmap_fedproto_without_cps_class2} illustrates the sparse activation phenomenon via a heatmap of 500-dimensional local prototypes (mean decision layer activation) across 20 clients and the global prototype for class \#2 in CIFAR-10. In this visualization, colored features indicate non-zero values, while blank areas represent dead units per class. Several clients (L1, L3, L4) show partial prototype utilization, while some clients display zero prototype vectors due to the absence of class \#2 in their local datasets.

Deep networks maintain high performance despite this sparse activation phenomenon, attributable to their substantial capacity and robust generalization capabilities \citep{arpit2017closer,zhang2021understanding,kawaguchi2022generalization}. Based on these characteristics, sparse prototypes could reduce server-client communication if the sparse locations were consistent across clients. However, in PBFL, model and data heterogeneity cause these sparse locations to vary among clients, as evidenced in Figure \ref{fig:heatmap_fedproto_without_cps_class2}. The resulting inconsistency leads to full utilization of global prototype dimensions, shown in Figure \ref{fig:heatmap_fedproto_without_cps_global}, thereby creating communication inefficiency.

\subsubsection{Class-Wise Prototype Sparsification}
To leverage sparsity in PBFL, we propose Class-wise Prototype Sparsification (CPS), which imposes structured sparsity per class to ensure consistent zero locations across clients. The method uses class-specific binary masking vectors to share predetermined sparse locations in prototypes. Through these masks, only a subset of elements where mask values are 1s are selected and communicated.

Let $\vm = (m_1, m_2, \ldots, m_d) \in \{0,1\}^d$ be a masking vector and $\bar{\vc} = (\bar{c}_1, \bar{c}_2, \ldots, \bar{c}_d) \in \mathbb{R}^d$ be a prototype. For notational simplicity, we omit class-specific subscripts $j$ from $\vm_j$ and $\bar{\vc}_j$ when the context is clear. With these vectors, we define the structured sparse prototype for updating local models and the compressed prototype for communicating prototypes.

\begin{definition}[Structured Sparse Prototype]
A structured sparse prototype $\tilde{\vc} \in \mathbb{R}^d$ is obtained by applying the sparsification operator $S: \mathbb{R}^d \rightarrow \mathbb{R}^d$ to an original prototype $\bar{\vc} \in \mathbb{R}^d$ with a binary mask $\vm \in \{0,1\}^d$ as:
\begin{equation}
\tilde{\vc} = S(\bar{\vc}; \vm) = \bar{\vc} \odot \vm,
\end{equation}
where $\odot$ denotes the Hadamard product.
\label{df:sparse_prototype}
\end{definition}

\begin{definition}[Compressed Prototype]
The compressed prototype $\hat{\vc} \in \mathbb{R}^s$ is obtained by applying the compression operator $C: \mathbb{R}^d \rightarrow \mathbb{R}^s$ to an original prototype $\bar{\vc} \in \mathbb{R}^d$ with a binary mask $\vm \in \{0,1\}^d$ as:
\begin{equation}
\hat{\vc} = C(\bar{\vc}; \vm) = (\bar{c}_i : m_i = 1),
\end{equation}
where $s = \sum_{i=1}^d m_i$ is the number of non-zero elements.
\label{df:compact_prototype}
\end{definition}

\begin{figure}[bt]
    \centering
    \includegraphics[width=0.99\columnwidth]{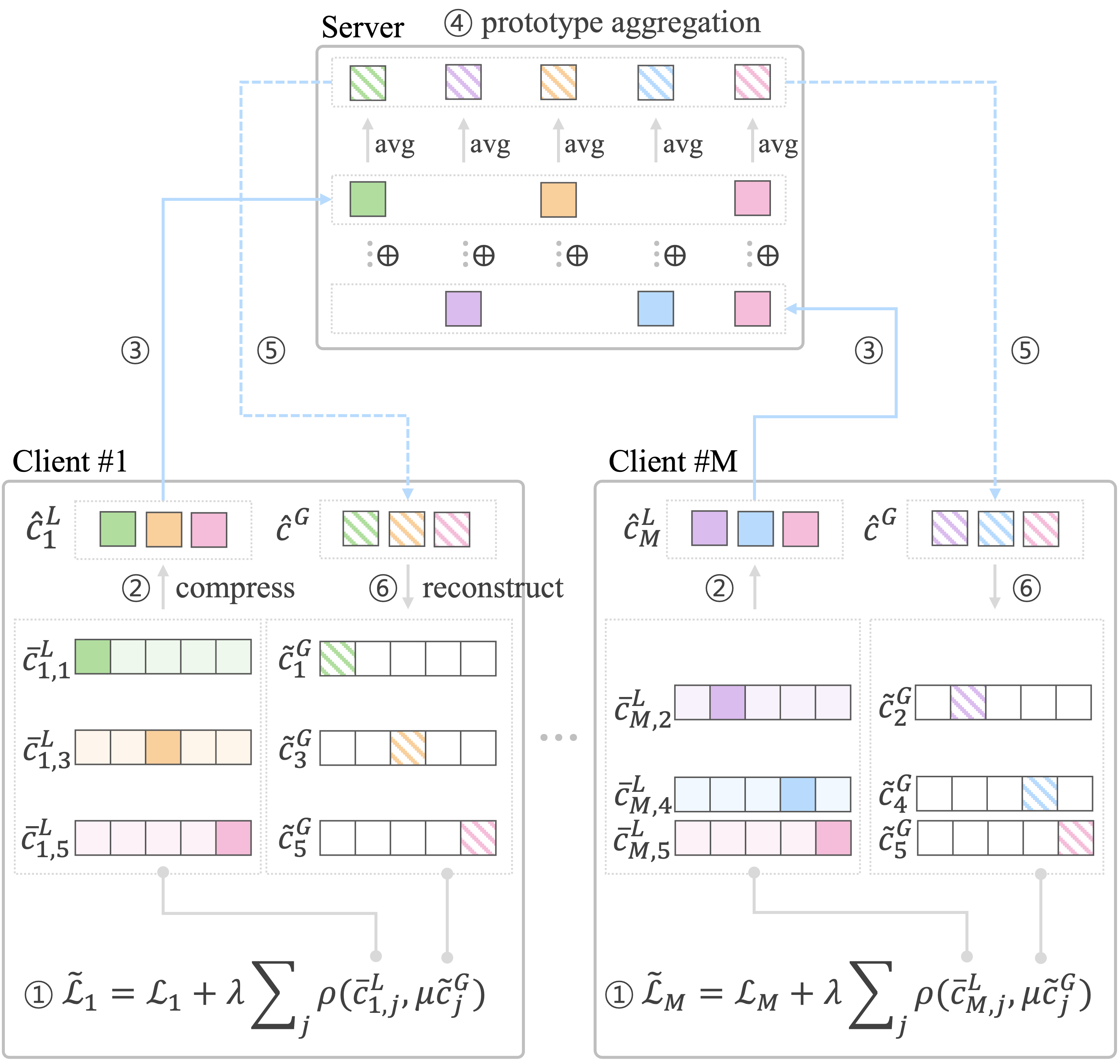}
    \caption{
        Process flow of TinyProto illustrated using a 5-class classification example with 5-dimensional prototypes and \( M \) clients. 
        Masking vectors are \( \vm_1 = (1, 0, 0, 0, 0), \vm_2 = (0, 1, 0, 0, 0), \dots, \vm_5 = (0, 0, 0, 0, 1) \). 
        The flow includes: 
        \textcircled{1} Each client trains its local model with regularization (initially without regularization) and generates local prototypes. 
        \textcircled{2} Local prototypes are compressed. 
        \textcircled{3} Compressed prototypes are transmitted to the server. 
        \textcircled{4} The server aggregates the compressed local prototypes. 
        \textcircled{5} The server distributes the compressed global prototypes to clients. 
        \textcircled{6} Clients reconstruct the compressed global prototypes into structured sparse prototypes, repeating the process.
    }
    \label{fig:tinyproto_diagram}
\end{figure}

The binary mask $\vm$ is shared between the server and clients to enable efficient communication. Rather than transmitting complete prototypes ($\bar{\vc}$), only compressed prototypes ($\hat{\vc}$) are exchanged. This compression corresponds to the colored dimensions where mask values are 1s, as illustrated in Figures \ref{fig:heatmap_fedproto_with_cps_class2} and \ref{fig:heatmap_fedproto_with_cps_global}. To ensure inter-class distinctiveness, we typically maximize pairwise Hamming distances between the masking vectors. 
The structured sparse global prototypes ($\tilde{\vc}^{G}$) guide local prototype sparsification during model training, as formulated in Eq. (\ref{eq:loss_function}) and (\ref{eq:regularization_term}).
The CPS can be seamlessly integrated into the standard PBFL process described in Section \ref{sec:problem}, as illustrated in the process flow of TinyProto using CPS in Figure \ref{fig:tinyproto_diagram}.

\subsubsection{Convergence Analysis of the CPS method}
Our convergence analysis builds upon FedProto \citep{tan2022fedproto}. Theorem 2 in FedProto establishes its non-convex convergence rate under specific hyperparameter selections. By extending Lemma 2 of FedProto, we show that CPS achieves the same convergence rate. FedProto's Lemma 2 proves that the loss function for any client can be bounded after server-side prototype aggregation. To adapt this lemma to our framework, we introduce an additional assumption: the sparsification operator $S$ (Definition 1) must maintain a consistent sparsity pattern per class across iterations and satisfy non-expansiveness. This extension preserves the lemma's bounded conditions for CPS, maintaining the convergence guarantees from FedProto's Theorem 2 within our framework. The complete details of this additional assumption and modified lemma are provided in Appendix.

\subsection{Prototype Scaling for Performance and Privacy Enhancement}

In this subsection, we address two critical challenges when CPS is applied for PBFL. First, while global prototypes can be computed using Eq.~($\ref{eq:global_prototype_aggregation}$) to incorporate each client's contribution in PBFL, this approach raises privacy concerns due to the transmission of local class distributions $n_{i,j}$ \citep{zhang2024fedtgp}. The method requires transmitting $n_{i,j}$ to calculate $\sum_{i=1}^M n_{i,j}$.
Consequently, simple averaging is commonly adopted in practice. For instance, the implementations in \cite{tan2022fedproto} and \cite{zhang2024fedtgp} use: 
\begin{equation}
\bar{\vc}_{j}^{G} = \frac{1}{|\mathcal{N}_j|} \sum_{i \in \mathcal{N}_j} \bar{\vc}_{i,j}^{L}, 
\label{eq:simple_averaging}
\end{equation}
where $\mathcal{N}_j$ represents the set of clients containing class $j$.
Second, when implemented with simple averaging (Eq. (\ref{eq:simple_averaging})), PBFL methods typically underperform compared to existing HtFL approaches, as they fail to reflect each client's class-specific importance denoted by $n_{i,j}$. This parallels FedAvg's performance degradation when simple averaging omits $n_i$ during aggregation, failing to account for client importance.

To address these issues, we propose a simple yet effective prototype scaling method that improves performance without directly transmitting $n_{i,j}$ to the server. The method scales both local and global prototypes. Local scaling is performed before transmitting prototypes to the server during \textbf{Step 1} (Local Prototype Generation) of PBFL, while global scaling is applied at the start of \textbf{Step 3} (Local Model Training).

\vspace{0.5em}   
\noindent \textbf{Local Prototype Scaling.} We scale the local prototypes by the sample count for the $j$-th class of the $i$-th client. This adjustment preserves each client's importance in federated learning as in Eq.~($\ref{eq:global_prototype_aggregation}$). Instead of transmitting $n_{i,j}$ directly, we send the scaled compressed prototype $n_{i,j} \hat{\vc}_{i,j}^{L}$, computed on the client side. This approach prevents privacy leakage by obfuscating $n_{i,j}$ during communication and at the server level. At the server, the global prototype for each class is computed as the mean of the received scaled prototypes, normalized by the number of clients contributing to that class:
\begin{equation}
\hat{\vc}_{j}^{G} = \frac{1}{|\mathcal{N}_j|} \sum_{i \in \mathcal{N}_j} n_{i,j} \hat{\vc}_{i,j}^{L}.
\label{eq:modified_global_prototype_aggregation}
\end{equation}

\noindent \textbf{Global Prototype Scaling.} We rescale the global prototypes using a hyperparameter-defined scaling constant to ensure stable training of local models. This adjustment mitigates potential stability issues caused by the varying scales of prototypes across clients and classes. The adjusted prototype regularization term for TinyProto is then defined as:
\begin{equation}
\mathcal{R}_i = \sum_{j} \rho(\bar{\vc}_{i,j}^{L}, \mu \tilde{\vc}_{j}^{G}),
\label{eq:modified_regularization_term}
\end{equation}
where $\mu$ represents the scaling constant, and $\tilde{\vc}_{j}^{G}$ denotes the structured sparse prototype for class $j$ reconstructed from compressed global prototype $\hat{\vc}_{j}^{G}$.

Despite its simplicity, the proposed scaling method achieves significant performance improvements while maintaining privacy, as demonstrated by our empirical results in Section \ref{sec:experiment}.
While \cite{zhang2024fedtgp} highlights that FedTGP avoids using $n_{i,j}$ during aggregation to address privacy concerns, our experimental results reveal that our proposed scaling method securely delivers significant performance improvements for FedTGP too.
Details on its application to FedTGP are provided in Appendix.

\subsection{TinyProto Algorithm}
The strength of TinyProto lies in its compatibility with existing PBFL algorithms. Integrating CPS into vanilla PBFL (FedProto) requires modifications such as creating and sharing masking vectors, and sparsifying and reconstructing prototypes using these vectors. Similarly, our scaling method can be applied to FedProto by multiplying the local prototype by $n_{i,j}$ and modifying the loss function. The detailed integration process is presented in Algorithm~\ref{algo:integration}.

\begin{algorithm}[bt]
\caption{TinyProto-FP \small (TinyProto integrated into FedProto)}
\hspace*{0.02in} {\bf Input:} $\mathcal{D}_i$, $\bm{w}_i$, $i=1, ..., M$ \\
\hspace*{0.02in} {\bf Server executes:}
\begin{algorithmic}[1]
\State Initialize masking vector set $\left\{\vm_j\right\}$ and compressed prototype set $\left\{\hat{\vc}_{j}^{G}\right\}$ for all classes.
\State Initialize set $\mathcal{S}^0=\{\}$ for clients selected up to the current iteration 
\For{iteration $t = 1, \ldots, T$}
    \State Sample a client subset $\mathcal{C}^t$
    \If{$i \notin \mathcal{S}^{t-1}$} 
        \State Send $\vm_j$ to client $i \in \mathcal{C}^t$
    \EndIf
    \For{client $i \in \mathcal{C}^t$ in parallel}
        \State $n_{i,j}\hat{\vc}^{L}_{i,j} \leftarrow$ LocalUpdate$\left(i, \hat{\vc}_{j}^{G}\right)$
    \EndFor
    \State Update $\hat{\vc}_{j}^{G}$ with Eq. (\ref{eq:modified_global_prototype_aggregation})
    \State Update $\mathcal{S}^t = \mathcal{S}^{t-1} \cup \mathcal{C}^t$
\EndFor
\end{algorithmic}
\hspace*{0.02in} \\
\hspace*{0.02in} {\bf LocalUpdate}$\left(i, \hat{\vc}_{j}^{G}\right)$:
\begin{algorithmic}[1]
\State Reconstruct $\tilde{\vc}_{j}^{G}$ from $\hat{\vc}_{j}^{G}$
\For{each local epoch}
    \For{batch $(x_{i}^{(l)}, y_{i}^{(l)}) \in \mathcal{D}_i$}
        \State Update model using the loss in Eq. (\ref{eq:loss_function}) and (\ref{eq:modified_regularization_term})
    \EndFor
\EndFor
\State Compute $\bar{\vc}^{L}_{i,j}$ by Eq. (\ref{eq:local_prototype_aggregation}) and convert it to $\hat{\vc}^{L}_{i,j}$ 
\State \Return $n_{i,j}\hat{\vc}^{L}_{i,j}$
\end{algorithmic}
\label{algo:integration}
\end{algorithm}

\begin{table*}[t]
\centering
{\fontsize{9}{11}\selectfont
\begin{tabular}{lrrrrrr}
\toprule
\multirow{2}{*}{Algorithm (CPS dim.) } & \multicolumn{2}{c}{CIFAR-10} & \multicolumn{2}{c}{CIFAR-100} & \multicolumn{2}{c}{TinyImageNet} \\
\cmidrule(lr){2-3} \cmidrule(lr){4-5} \cmidrule(lr){6-7}
 & Acc. (\%) & Comm. (M) & Acc. (\%) & Comm. (M) & Acc. (\%) & Comm. (M) \\
\midrule
LG-FedAvg & 86.91 $\pm$ 0.14 & 0.20 & 38.54 $\pm$ 0.21 & 2.00 & 22.30 $\pm$ 0.37 & 4.00 \\
FML & 86.59 $\pm$ 0.15 & 34.32 & 37.83 $\pm$ 0.03 & 36.12 & 22.03 $\pm$ 0.12 & 38.12 \\
FedKD & 87.10 $\pm$ 0.02 & 30.66 & 39.74 $\pm$ 0.42 & 32.26 & 23.08 $\pm$ 0.17 & 34.05 \\
FedDistill & \underline{86.93 $\pm$ 0.12} & \underline{$<$0.01} & 39.52 $\pm$ 0.33 & 0.29 & 22.98 $\pm$ 0.15 & 1.17 \\
FedProto & 82.90 $\pm$ 0.46 & 0.15 & 29.97 $\pm$ 0.18 & 1.46 & 13.30 $\pm$ 0.06 & 2.93 \\
FedTGP & 86.32 $\pm$ 0.49 & 0.15 & 36.92 $\pm$ 0.16 & 1.46 & 19.44 $\pm$ 0.12 & 2.93 \\
\midrule
TinyProto-FP (50) & 84.52 $\pm$ 0.06 & 0.02 & 31.82 $\pm$ 0.24 & 0.15 & 16.01 $\pm$ 0.21 & 0.29 \\
TinyProto-FT (50) & \textbf{88.47 $\pm$ 0.21} & \textbf{0.02} & \underline{\textbf{45.94 $\pm$ 0.40}} & \underline{\textbf{0.15}} & \underline{\textbf{27.29 $\pm$ 0.21}} & \underline{\textbf{0.29}} \\
\bottomrule
\end{tabular}
}
\caption{Classification test accuracy (Acc.) and communication cost (Comm.) across datasets with prototype dimension $d=500$. Communication cost is measured by the number of parameters shared per FL round (M: millions). TinyProto-FP and TinyProto-FT denote TinyProto integrated into FedProto and FedTGP, respectively. The CPS dim. of 50 corresponds to a 90\% compression rate.
}
\label{table:total_result}
\vspace{-5pt}
\end{table*}

\section{Experiments} \label{sec:experiment}

This section evaluates the performance and communication efficiency of our proposed methods, with particular emphasis on CPS and prototype scaling techniques.

\subsection{Experimental Setup}

\noindent {\bf Datasets and Models.}
We conduct our experiments on three widely used federated learning datasets: CIFAR-10, CIFAR-100 \citep{krizhevsky2009learning}, and TinyImageNet \citep{le2015tiny}. Each dataset is divided into training (75\%) and test (25\%) subsets. To simulate non-IID data distributions across clients, we partition the data using a Dirichlet distribution ($\text{Dir}(\alpha)$) with $\alpha = 0.1$ \citep{lin2020ensemble}.  
We employ lightweight models designed for resource-constrained environments, including ResNet-8 \citep{zhong2017deep}, EfficientNet \citep{tan2019efficientnet}, ShuffleNet v2 \citep{ma2018shufflenet}, and MobileNet v2 \citep{sandler2018mobilenetv2}. Each model integrates a global average pooling layer \citep{szegedy2015going}, with the default prototype dimension fixed at $d = 500$.

\begin{table}[tbh]
    \centering
    {\fontsize{9}{11}\selectfont
    \begin{tabular}{ll}
    \toprule
    Algorithm & Communication cost formulation \\
    \midrule
    LG-FedAvg & $\sum^M_{i=1} |\bm{\phi}_i|\times 2$ \\
    FML & $M\times (|\bm{\theta}_{aux}| + |\bm{\phi}_{aux}|)\times 2$ \\
    FedKD & $M\times (|\bm{\theta}_{aux}| + |\bm{\phi}_{aux}|)\times 2\times r$ \\
    FedDistill & $\sum^M_{i=1} (K_i + K) \times K $ \\
    FedProto and FedTGP & $\sum^M_{i=1} (K_i + K) \times d$ \\
    \midrule
    TinyProto & $\sum^M_{i=1} (K_i + K) \times s$ \\
    \bottomrule
    \end{tabular}
    }
    \caption{
    Per-round communication cost formulation. For FML and FedKD, $|\bm{\theta}_{aux}|$ and $|\bm{\phi}_{aux}|$ represent auxiliary feature extractor and classifier parameter sizes, with FedKD using dimensionality reduction factor $r$. TinyProto reduces prototype dimension from $d$ to $s$. For FedDistill and PBFL algorithms, only logits or prototypes for classes present in local datasets ($K_i$) are transmitted to the server.
    }
    \label{table:comm_cost_theory}
    \vspace{-5pt}
\end{table}

\noindent {\bf Federated Learning Configuration.}
Our federated learning setup involves 20 clients, all actively participating in each of the 300 communication rounds. The client-side training configuration consists of a learning rate of 0.01, a batch size of 32, and one local training epoch per round. For the prototype regularization term, we set $\lambda=1$ following FedProto and determine $\mu$ through grid search: $\mu=1.5 \times 10^{-4}$ for CIFAR-10 and $\mu=1.5 \times 10^{-3}$ for CIFAR-100 and TinyImageNet. We evaluate the proposed approaches, TinyProto integrated into FedProto and FedTGP (referred to as TinyProto-FP and TinyProto-FT), against six state-of-the-art data-free federated learning algorithms: LG-FedAvg, FML, FedKD,  FedDistill, FedProto, and FedTPG.

\noindent {\bf Evaluation Protocol.}
For PBFL methods (FedProto, FedTGP, and TinyProto), predictions $\hat{y}$ are determined by minimizing the L2 distance between the decision-layer activation $f_i(\bm{\theta}_i;x)$ and local prototypes $\bar{\vc}_{i,j}^{L}$ \citep{tan2022fedproto}:
\begin{equation}
\hat{y} = \argmin_j \| f_i(\bm{\theta}_i;x) - \bar{\vc}_{i,j}^{L} \|_2.
\end{equation}
The primary evaluation metric is the highest mean test accuracy achieved across all communication rounds, which is a standard measure in federated learning research \citep{mcmahan2017communication}. Each experiment is repeated three times with different random seeds, and the average results are reported to ensure statistical robustness. No hyperparameter schedulers are applied during training to maintain fairness.  
Detailed configurations, including hyperparameter settings and additional results, are provided in Appendix. Code is available at: https://github.com/regulationLee/TinyProto

\begin{figure}[t]
   \centering
   \includegraphics[width=0.9\columnwidth]{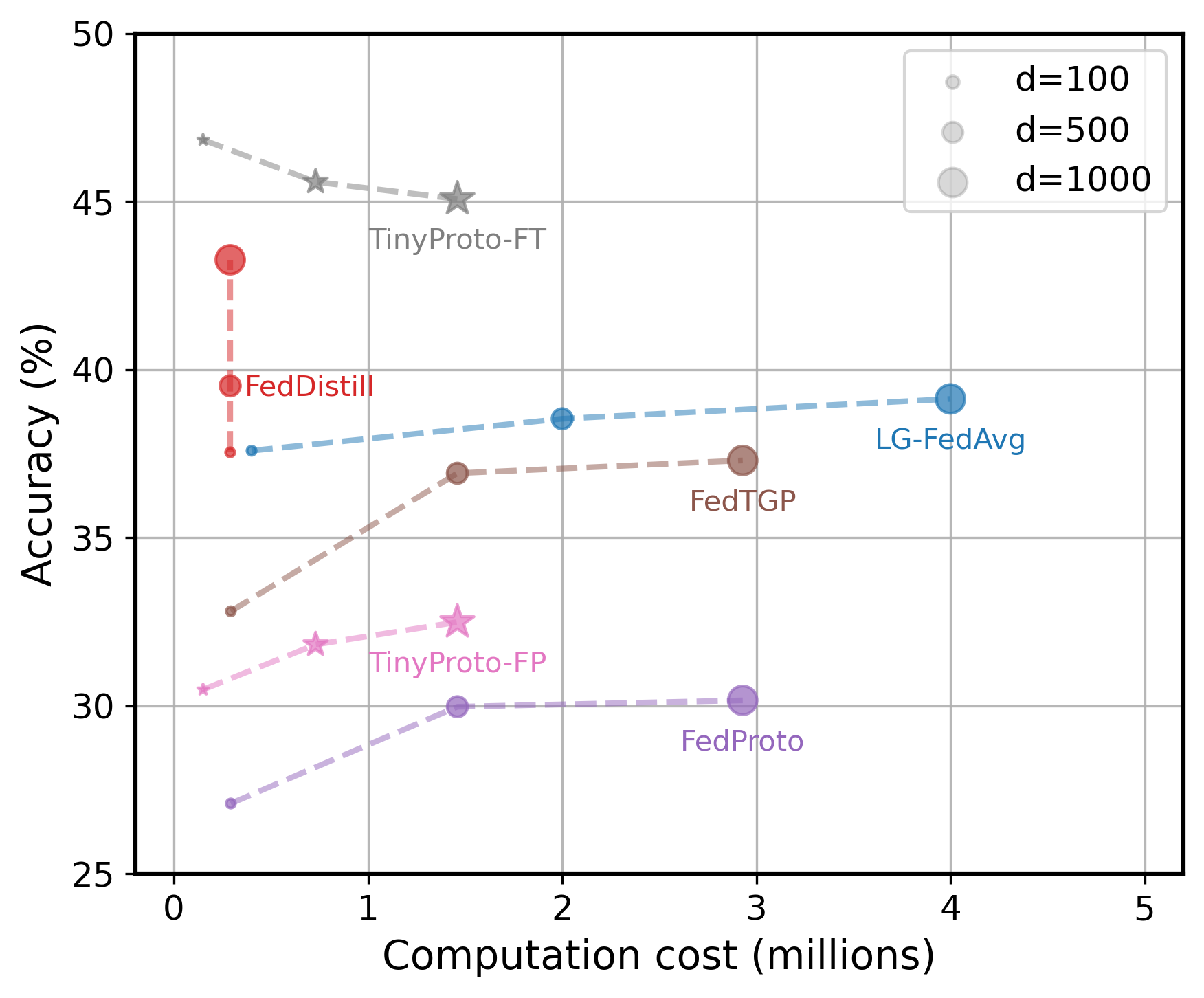}
   \caption{Impact of decision layer neurons (prototype dimension) on accuracy and communication cost for CIFAR-100. The CPS dimension $s$ is set to 10\% of prototype dimension $d$ with scaling method applied. We exclude FML and FedKD from visualization due to their substantially higher communication costs, as shown in Table \ref{table:total_result}.}
   \label{fig:impact_num_neurons}
\end{figure}

\subsection{Communication Cost and Performance}
\noindent {\bf Communication Cost Reduction.} 
The communication costs of different approaches are compared in Table~\ref{table:total_result}, with their theoretical formulations presented in Table~\ref{table:comm_cost_theory}. Note that our FedProto baseline implementation follows \cite{tan2022fedproto}, utilizing Eq. (\ref{eq:simple_averaging}).
TinyProto-FT with the prototype scaling method ($s=50$) demonstrates superior communication efficiency and higher accuracy compared to baseline methods (\underline{underlined}), except for datasets with few classes such as CIFAR-10. This achieves up to 4× reduction in communication costs compared to FedDistill, the most communication-efficient baseline with significantly higher performance. While FedDistill achieves the lowest communication cost for CIFAR-10 due to its fewer classes relative to prototype dimensions, its transmission of averaged logits introduces higher privacy risks compared to prototype sharing. Moreover, FedDistill cannot leverage CPS-style compression since logits must maintain their dense representation. Notably, TinyProto-FT consistently outperforms various algorithms across different settings, as highlighted in \textbf{bold} as shown in Table \ref{table:total_result}. 

\noindent {\bf Sensitivity to Class and Neuron Counts.} 
The impact of class count and decision layer neurons (prototype dimension $s$) is analyzed in Table {\ref{table:total_result}} and Figure {\ref{fig:impact_num_neurons}}. All algorithms exhibit increased communication costs as the number of classes grows, attributed to the expansion of output layer neurons (Table \ref{table:total_result}). However, TinyProto with CPS dimension $s=50$ is less affected by this increase since it is already compressed by 90\% compared to FedProto and FedTGP while achieving higher performance. For the impact of decision layer neurons from 100 to 500 and 1000, FedDistill remains unaffected as it only depends on class count as shown in Figure \ref{fig:impact_num_neurons}. While LG-FedAvg and PBFL are impacted by this parameter, TinyProto shows minimal sensitivity.

\noindent {\bf Scalability and Data Heterogeneity.} 
Table \ref{table:additional_result} demonstrates our method's robustness. When increasing client count ($M$) from 20 to 50, our method maintains superior performance with significantly less communication cost. This performance advantage holds across data distributions, from more heterogeneous ($\alpha=0.01$) to more homogeneous ($\alpha=0.5$) than the default setting ($\alpha=0.1$).

\begin{table}[t]
\centering
\setlength{\tabcolsep}{4.pt}
{\fontsize{9}{11}\selectfont
\begin{tabular}{lrrrrr}
\toprule
\multirow{2}{*}{Algo. (CPS dim.)} & \multicolumn{1}{c}{Scalability} & \multicolumn{2}{c}{Data heterogeneity} \\
\cmidrule(lr){2-2} \cmidrule(lr){3-4}
 & $M=50$ & $\alpha=0.01$ & $\alpha=0.5$ \\
\midrule
LG-FedAvg         & 37.47 $\pm$ 0.19    & 66.75 $\pm$ 0.36  & 20.87 $\pm$ 0.18 \\
FML               & 37.61 $\pm$ 0.12    & 64.29 $\pm$ 0.19  & 20.66 $\pm$ 0.20 \\
FedKD             & 38.39 $\pm$ 0.15    & 66.08 $\pm$ 0.24  & 20.94 $\pm$ 0.21 \\
FedDistill        & 40.70 $\pm$ 0.40    & 60.11 $\pm$ 0.21  & 22.28 $\pm$ 0.12 \\
FedProto          & 29.36 $\pm$ 0.57    & 58.98 $\pm$ 0.31  & 12.88 $\pm$ 0.36 \\
FedTGP            & 36.69 $\pm$ 0.15    & 64.02 $\pm$ 0.34  & 16.62 $\pm$ 0.28 \\
\midrule
TinyProto-FP (50) & 32.43 $\pm$ 0.31    & 61.52 $\pm$ 0.54  & 17.01 $\pm$ 0.45 \\
TinyProto-FT (50) & \textbf{43.51 $\pm$ 0.35}    & \textbf{72.74 $\pm$ 0.59}  & \textbf{24.35 $\pm$ 0.11} \\
\bottomrule
\end{tabular}
}
\caption{Classification test accuracy (\%) on CIFAR-100 under varying scalability and data heterogeneity. CPS dimension $s=50$ with the scaling method is applied to both TinyProto methods.}
\label{table:additional_result}
\end{table}

\begin{figure}[t]
    \centering
    \begin{subfigure}{0.48\columnwidth}
        \centering
        \includegraphics[width=\textwidth]{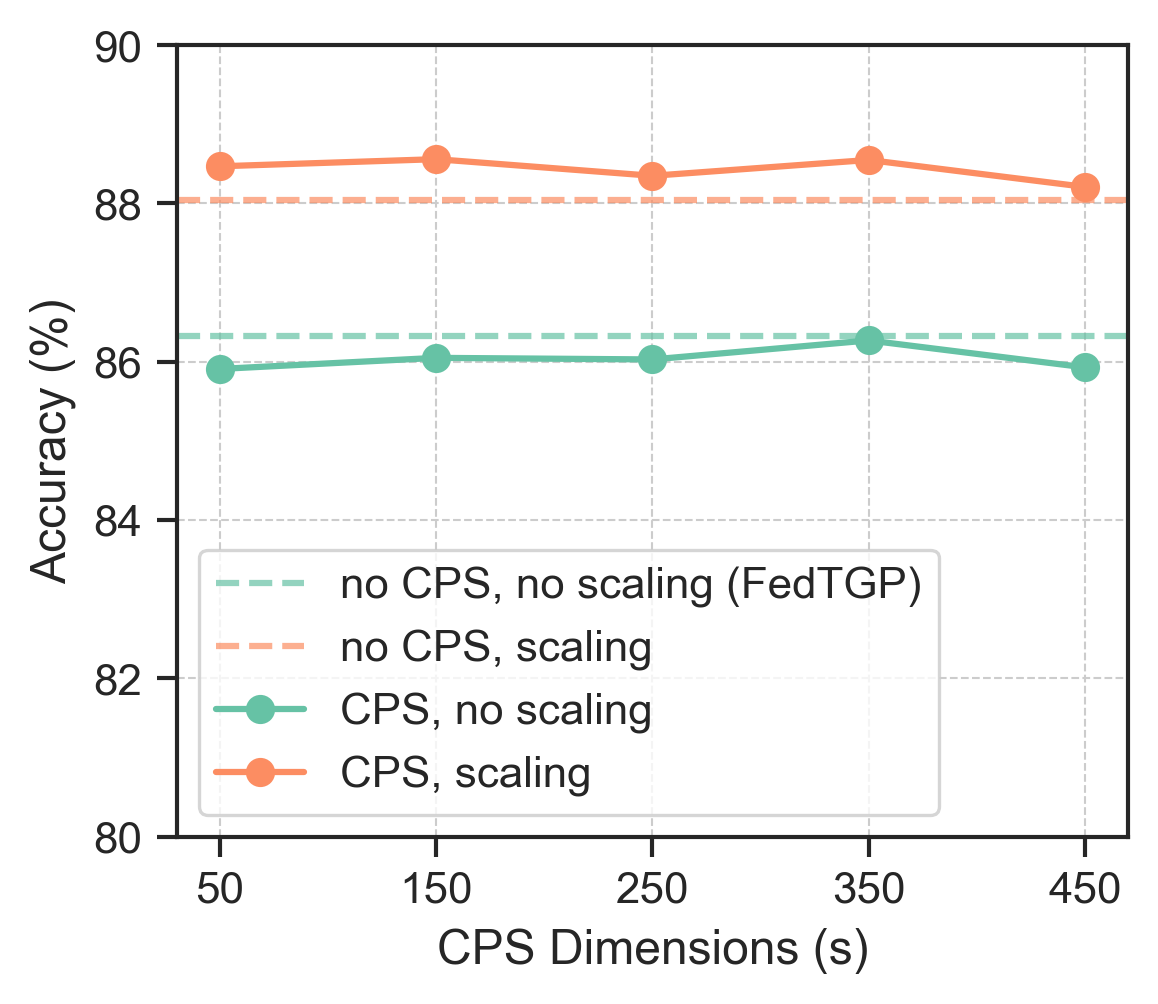}
        \caption{CIFAR-10}
    \end{subfigure}
    \hfill
    \begin{subfigure}{0.48\columnwidth}
        \centering
        \includegraphics[width=\textwidth]{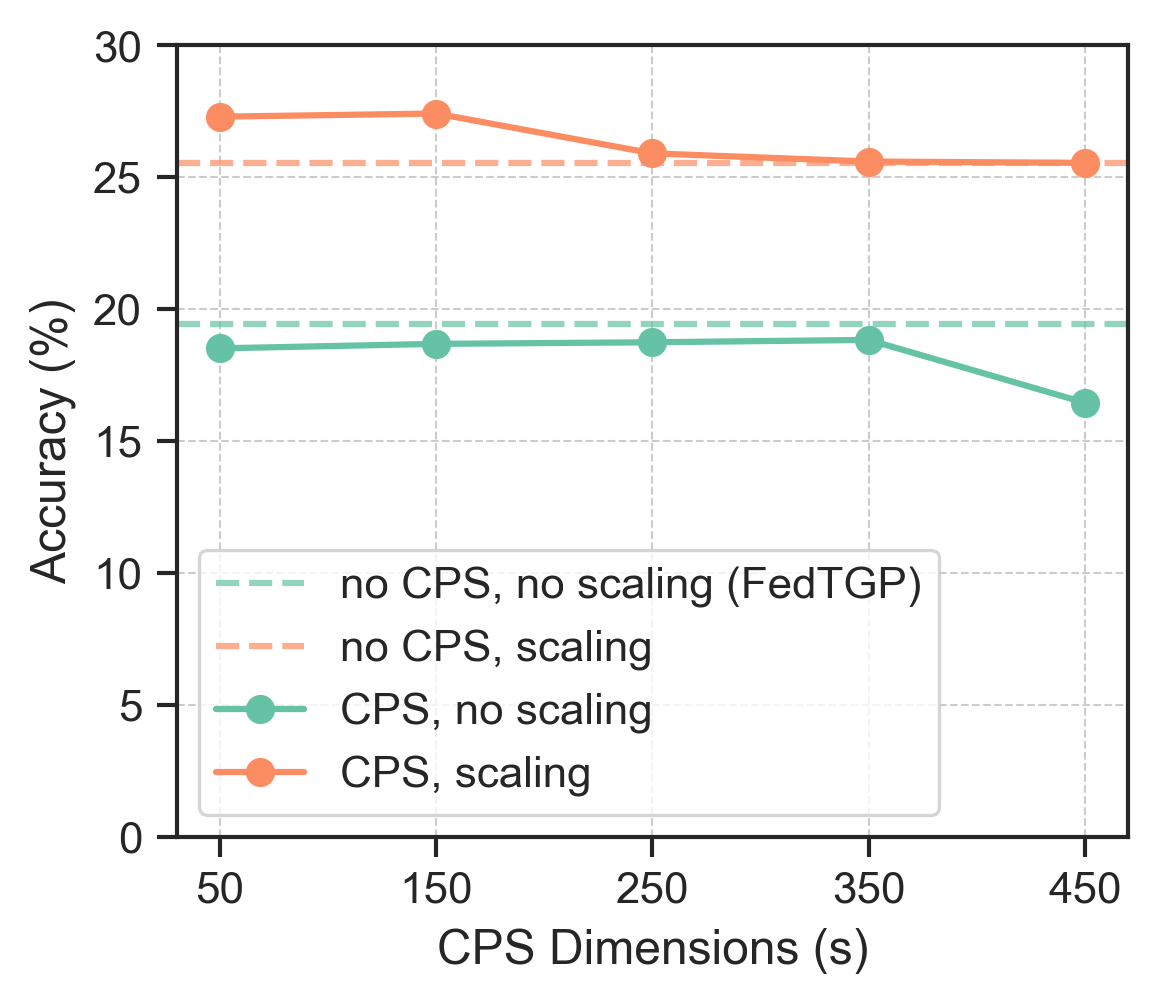}
        \caption{TinyImageNet}
    \end{subfigure}
    \caption{Classification accuracy (\%) comparison for TinyProto-FT with varying conditions. The CPS method varies CPS dimension $s$.}
    \label{fig:ablation}
\end{figure}

\begin{figure}[t]
    \centering
    \begin{subfigure}{0.48\columnwidth}
        \centering
        \includegraphics[width=\textwidth]{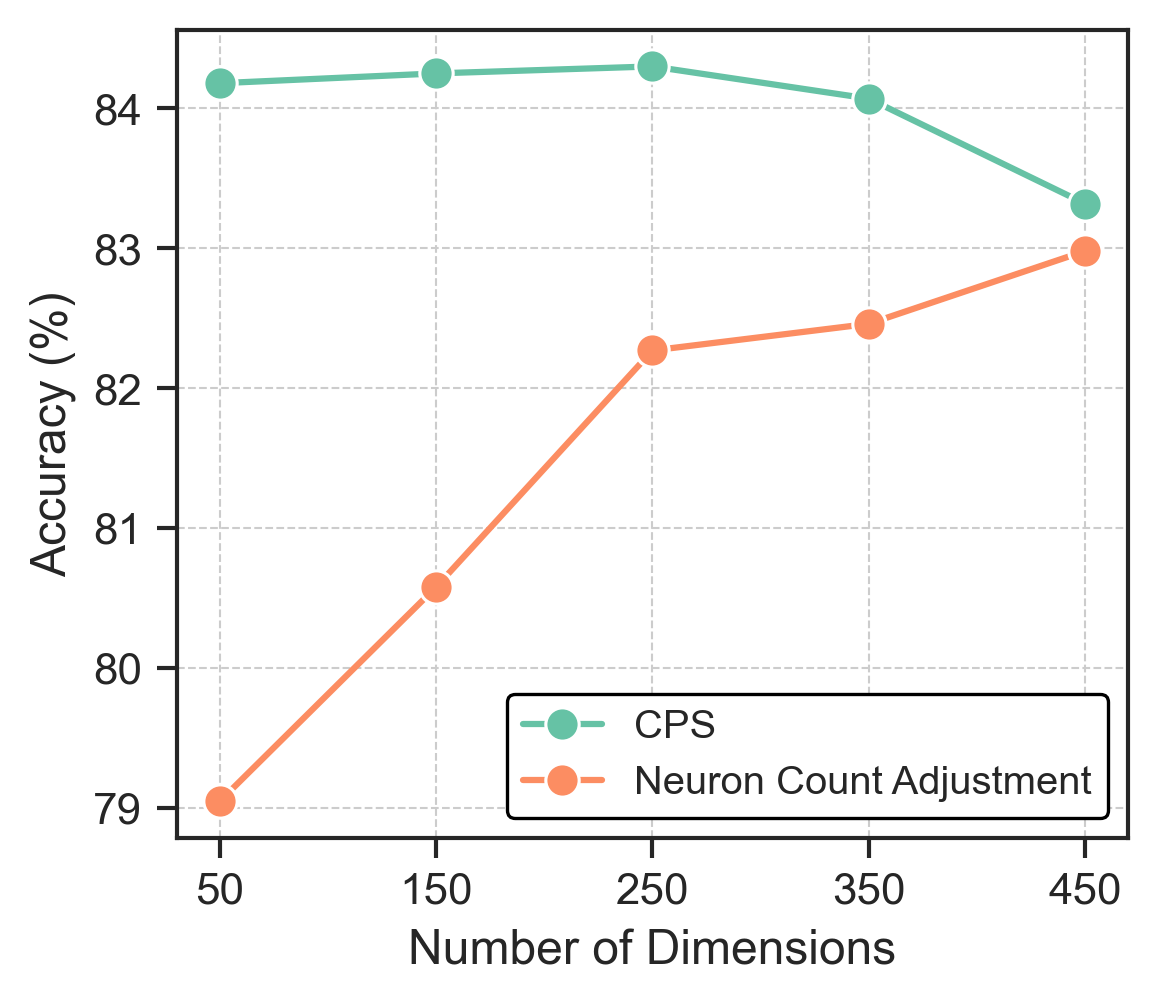}
        \caption{CIFAR-10}
    \end{subfigure}
    \hfill
    \begin{subfigure}{0.48\columnwidth}
        \centering
        \includegraphics[width=\textwidth]{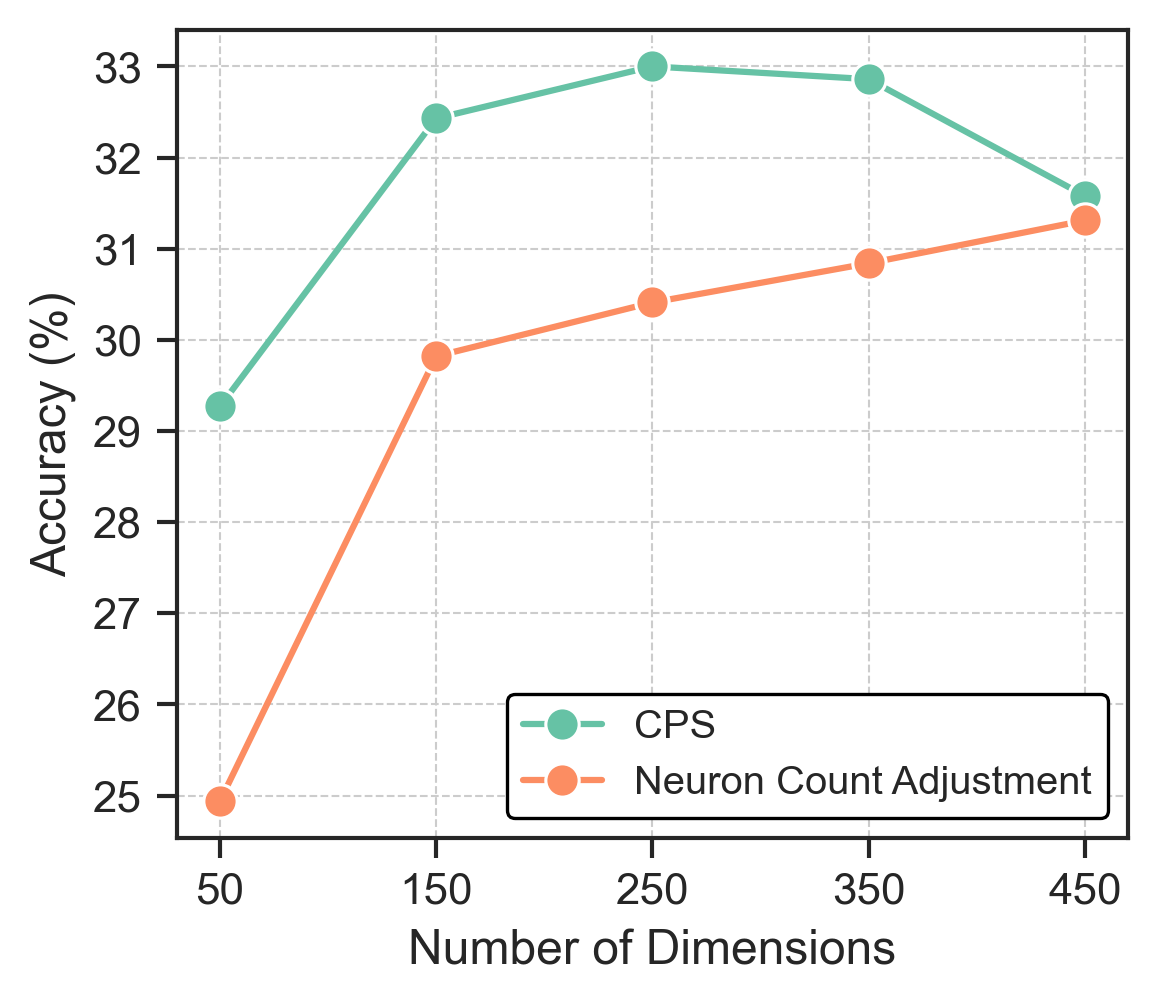}
        \caption{CIFAR-100}
    \end{subfigure}
    \caption{Classification accuracy (\%) comparison for TinyProto-FP. The CPS method varies CPS dimension $s$, while Neuron Count Adjustment modifies the number of decision layer neurons $d$.}
    \label{fig:compression}
\end{figure}

\subsection{Analysis of CPS and Prototype Scaling Effects}
\noindent {\bf Ablation Study.} 
TinyProto-FT with prototype scaling (orange dotted line) significantly outperforms FedTGP (green dotted line) in Figure \ref{fig:ablation}, likely due to effective incorporation of local prototype importance during server-side aggregation. Despite reducing dimensionality for sharing information through prototype sparsification, CPS maintains or slightly improves performance due to high capacity of deep networks, validating TinyProto's ability to reduce communication overhead without performance degradation. Similar trends are observed between TinyProto-FP and FedProto, with detailed ablation results available in Table \ref{table:supp_total_result} in Appendix.

\noindent {\bf CPS Dimension vs. Decision Layer Dimension}
Figure \ref{fig:compression} shows performance comparisons between CPS with varying CPS dimensions and models without CPS but with varying decision layer neurons at the same communication costs. CPS demonstrates consistently superior performance across different settings by maintaining the full decision layer capacity during training while achieving the same communication efficiency. Specifically, CPS retains the complete dimensionality during training and selectively sparsifies prototypes for communication, whereas reducing decision layer neurons permanently constrains model capacity through a smaller dimensionality in both training and inference phases.

\section{Limitation and Conclusion}
In this work, we introduce TinyProto, a framework leveraging structured sparsity and adaptive prototype scaling to address PBFL's inherent scalability challenges. While the introduction of CPS and prototype scaling requires additional hyperparameters that may increase deployment complexity, our framework effectively reduces communication overhead caused by high-dimensional prototypes and large class counts. Our extensive experiments demonstrate that TinyProto achieves a 10× reduction in communication costs compared to the original PBFL approach and 4× reduction versus existing efficient baselines, while maintaining or improving accuracy across diverse datasets. Beyond these efficiency gains, TinyProto offers distinct advantages over existing communication-efficient FL methods: it eliminates the need for additional computation such as fine-tuning and provides native support for heterogeneous models. These characteristics make TinyProto particularly well-suited for real-world deployments involving resource-constrained and heterogeneous clients, where both computational efficiency and architectural flexibility are essential.






\bibliographystyle{named}
\bibliography{0_main}

\onecolumn

\newpage

\appendix
\section*{Appendix}
\section{Visualization of Structured Sparse Prototypes} \label{sec:visualization_prototype}
We provide visualizations of structured sparse prototypes in both their original and binary forms for structured sparse prototype dimension ($s$) and datasets. The heatmaps are presented in pairs: those on the left depict the original values of the prototypes, while those on the right show the same prototypes with values converted to 1 when larger than 0, and 0 otherwise.
\begin{figure}[ht]
    \centering
    \begin{subfigure}{0.33\textwidth}
        \centering
        \includegraphics[width=\textwidth]{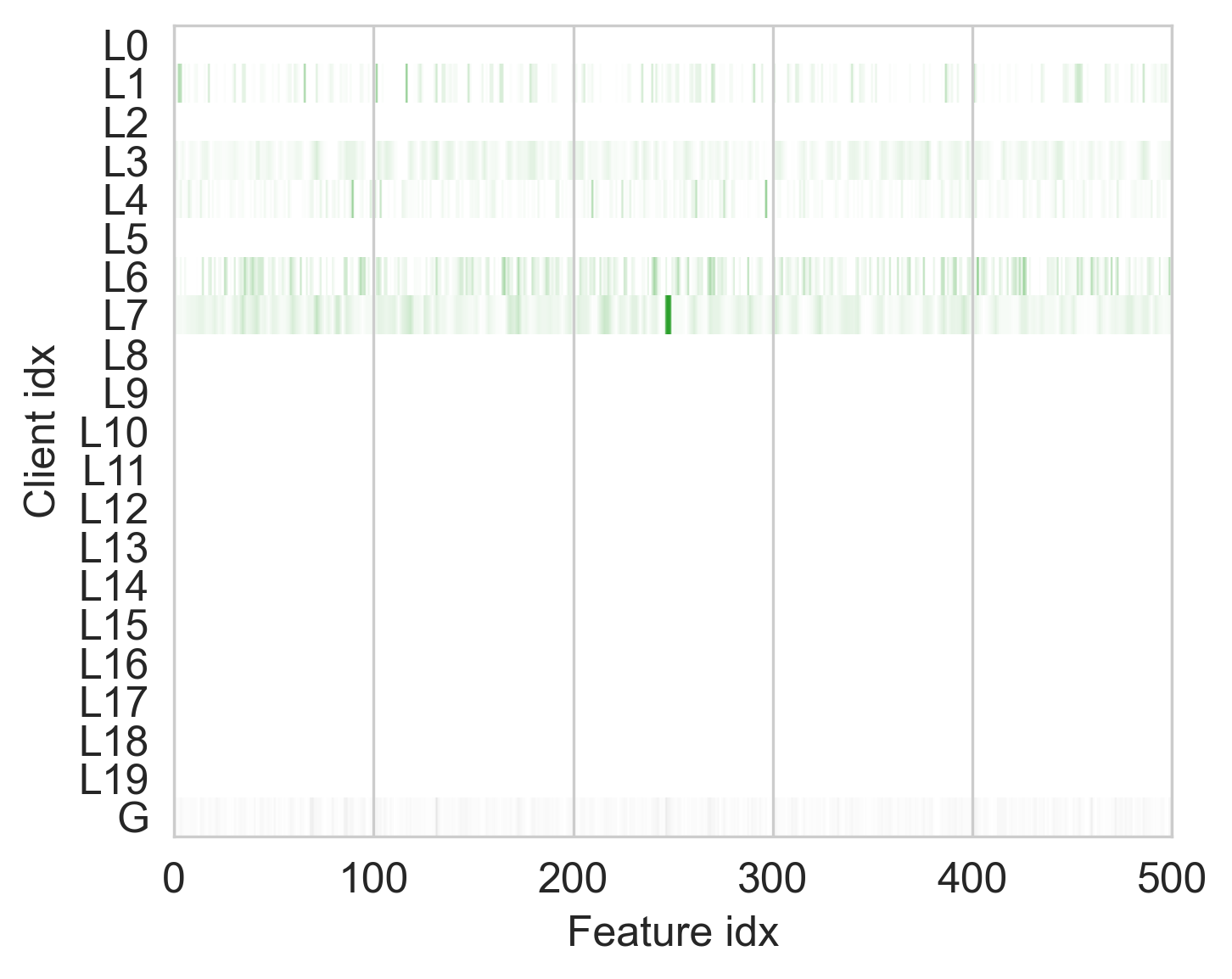}
        \caption{Prototypes of class \#2 (original)}
    \end{subfigure}
    \quad
    \begin{subfigure}{0.33\textwidth}
        \centering
        \includegraphics[width=\textwidth]{figures/FedProto_CIFAR10_Original_global_prototypes_2_binary.png}
        \caption{Prototypes of class \#2 (binary)}
    \end{subfigure}

    \begin{subfigure}{0.33\textwidth}
        \centering
        \includegraphics[width=\textwidth]{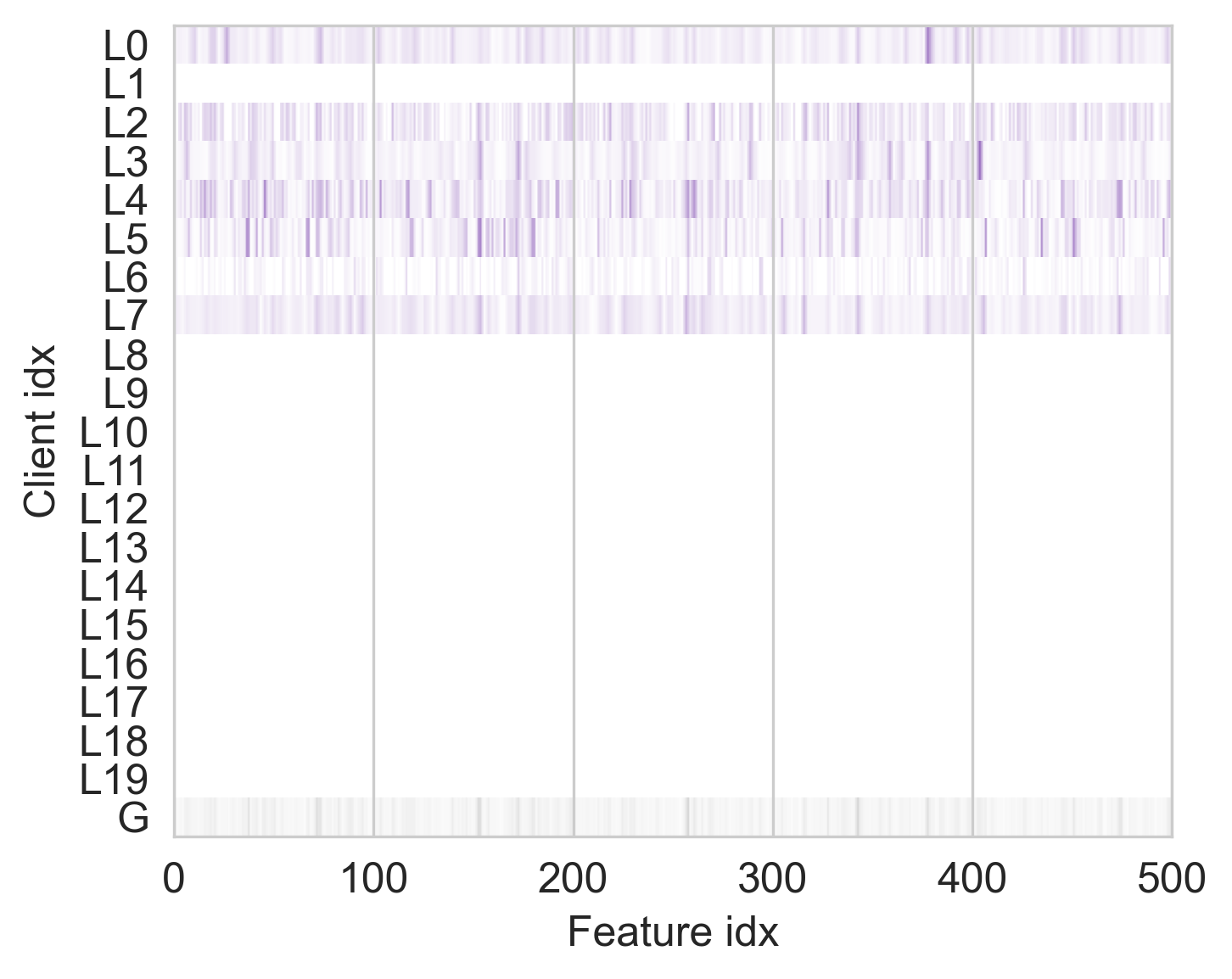}
        \caption{Prototypes of class \#4 (original)}
    \end{subfigure}
    \quad
    \begin{subfigure}{0.33\textwidth}
        \centering
        \includegraphics[width=\textwidth]{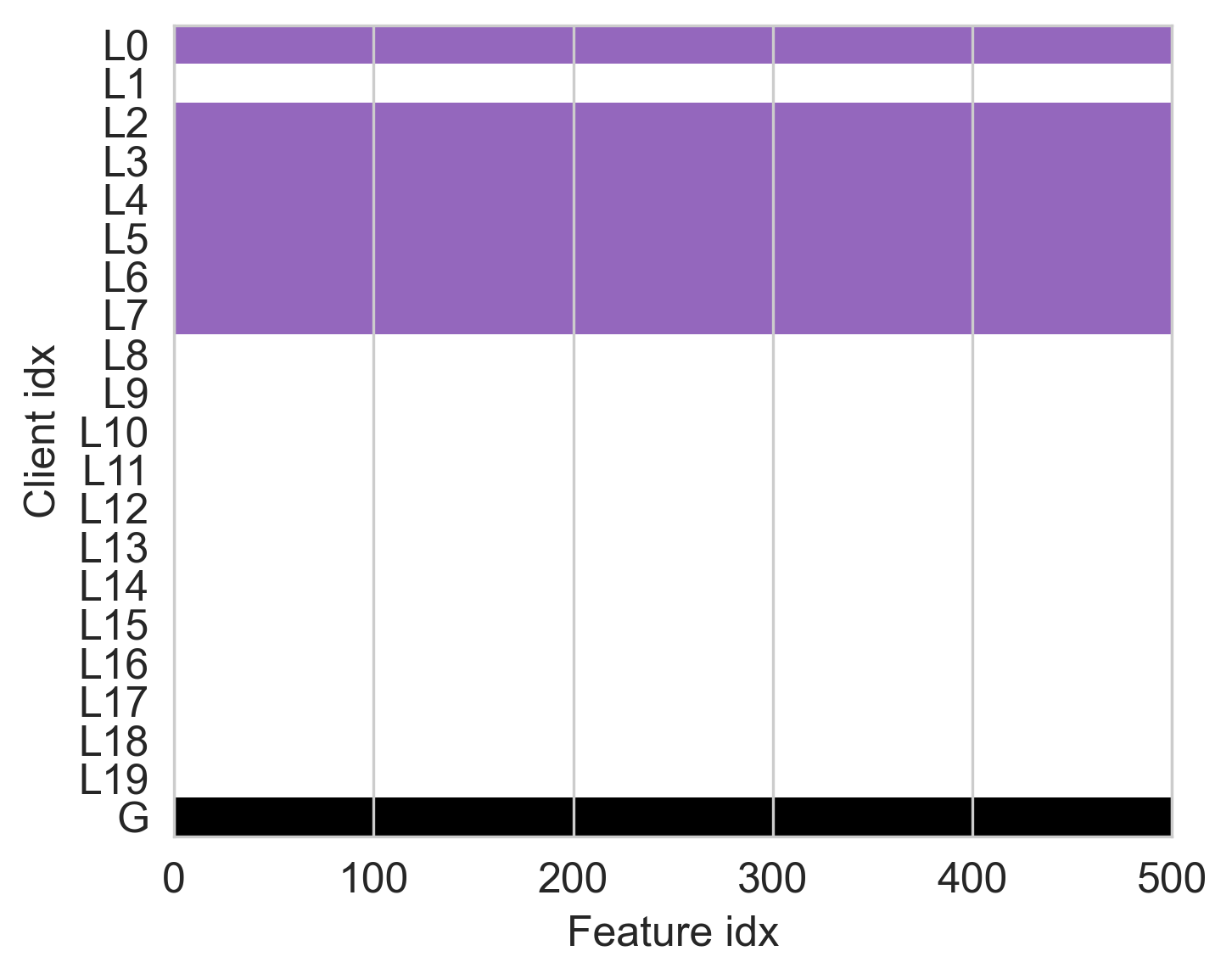}
        \caption{Prototypes of class \#4 (binary)}
    \end{subfigure}
    
    \begin{subfigure}{0.33\textwidth}
        \centering
        \includegraphics[width=\textwidth]{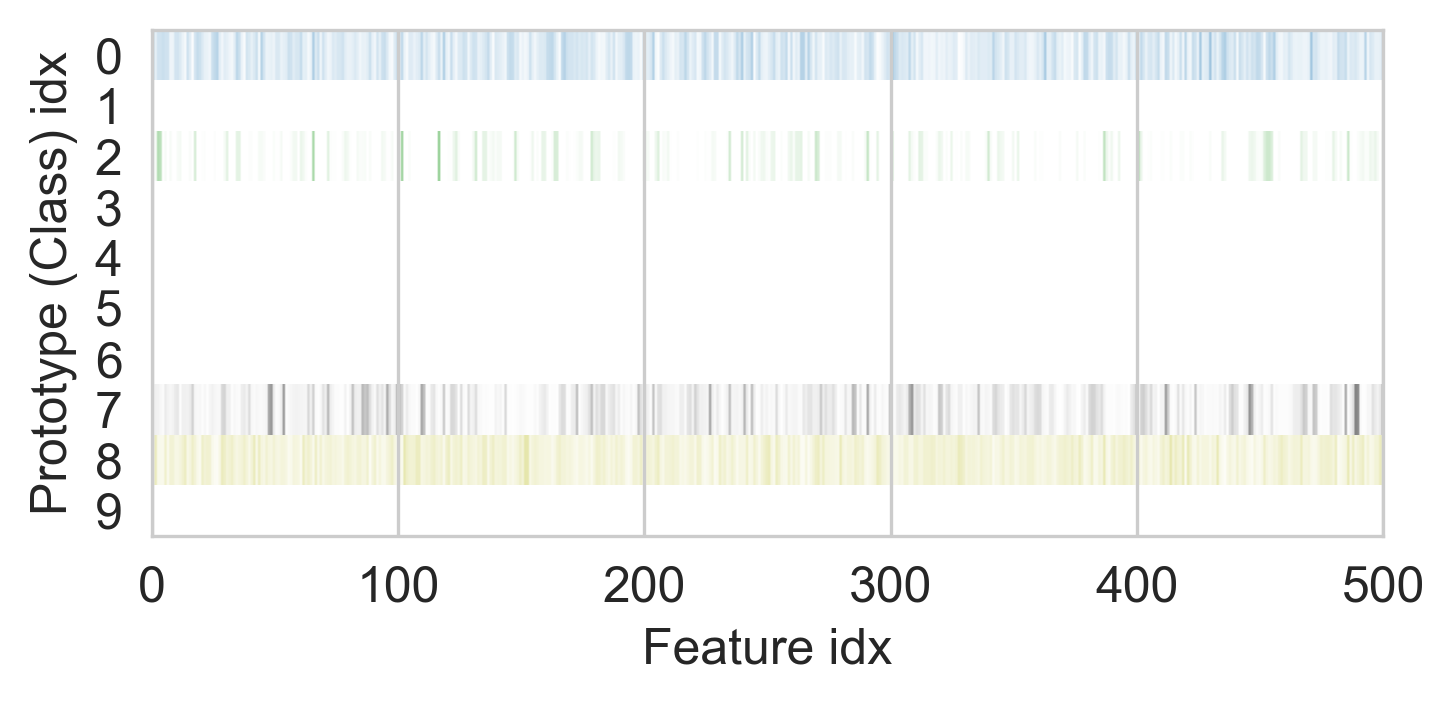}
        \caption{Prototypes of client \#10 (original)}
    \end{subfigure}
    \quad
    \begin{subfigure}{0.33\textwidth}
        \centering
        \includegraphics[width=\textwidth]{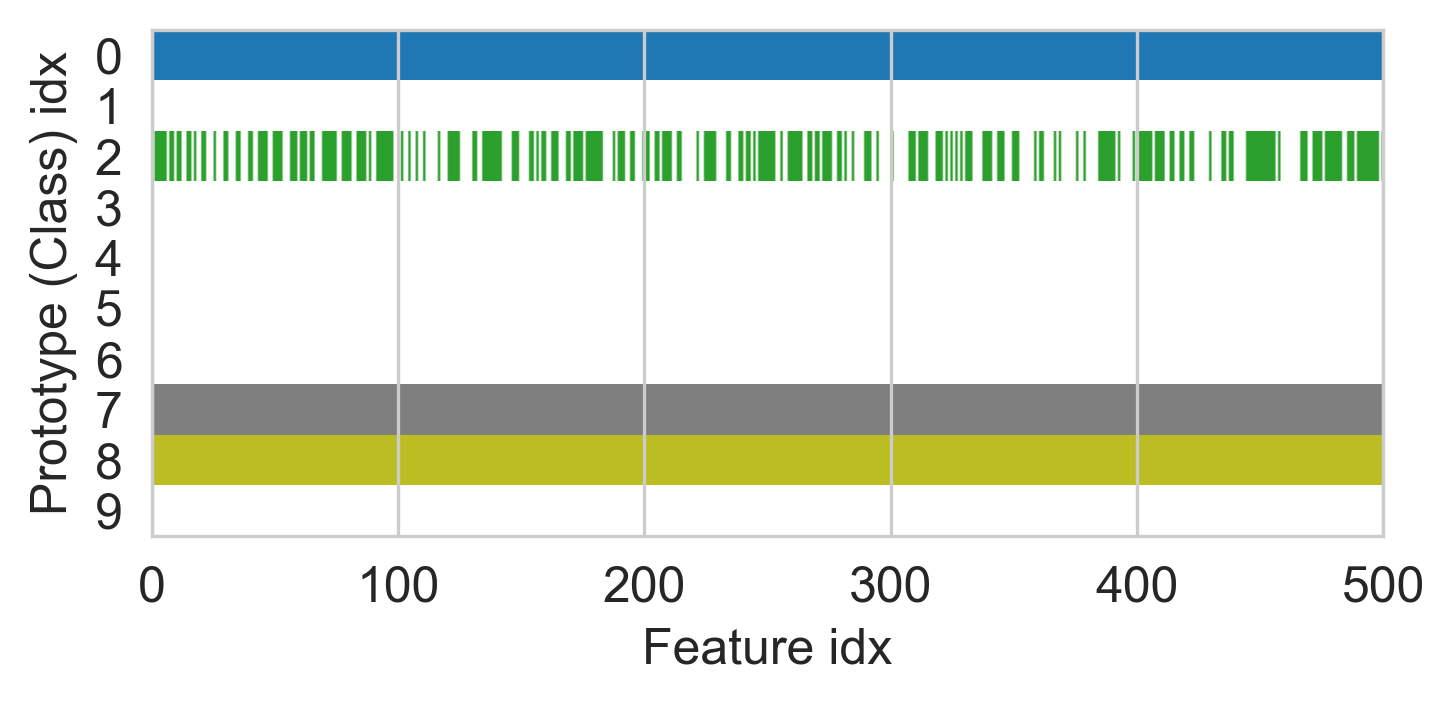}
        \caption{Prototypes of client \#10 (binary)}
    \end{subfigure}

    \begin{subfigure}{0.33\textwidth}
        \centering
        \includegraphics[width=\textwidth]{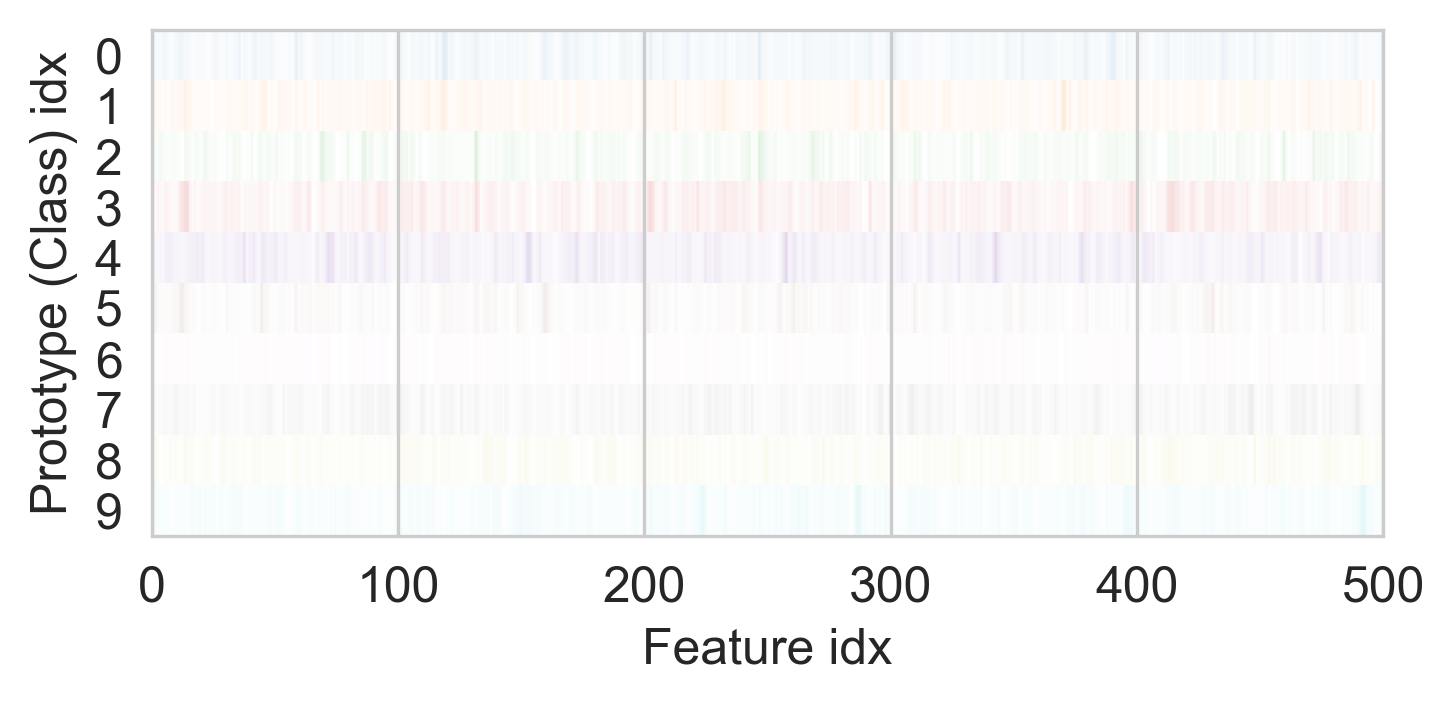}
        \caption{Global prototypes (original)}
    \end{subfigure}
    \quad
    \begin{subfigure}{0.33\textwidth}
        \centering
        \includegraphics[width=\textwidth]{figures/FedProto_CIFAR10_Original_global_prototypes_binary.png}
        \caption{Global prototypes (binary)}
    \end{subfigure}
    \caption{Prototype comparison of FedProto without Class-wise Prototype Sparsification (CPS) for the CIFAR-10 dataset.}
    \label{fig:heatmaps_prototype_cifar10_original}
    \vspace{-10pt}
\end{figure}

\begin{figure}[ht]
    \centering
    \begin{subfigure}{0.33\textwidth}
        \centering
        \includegraphics[width=\textwidth]{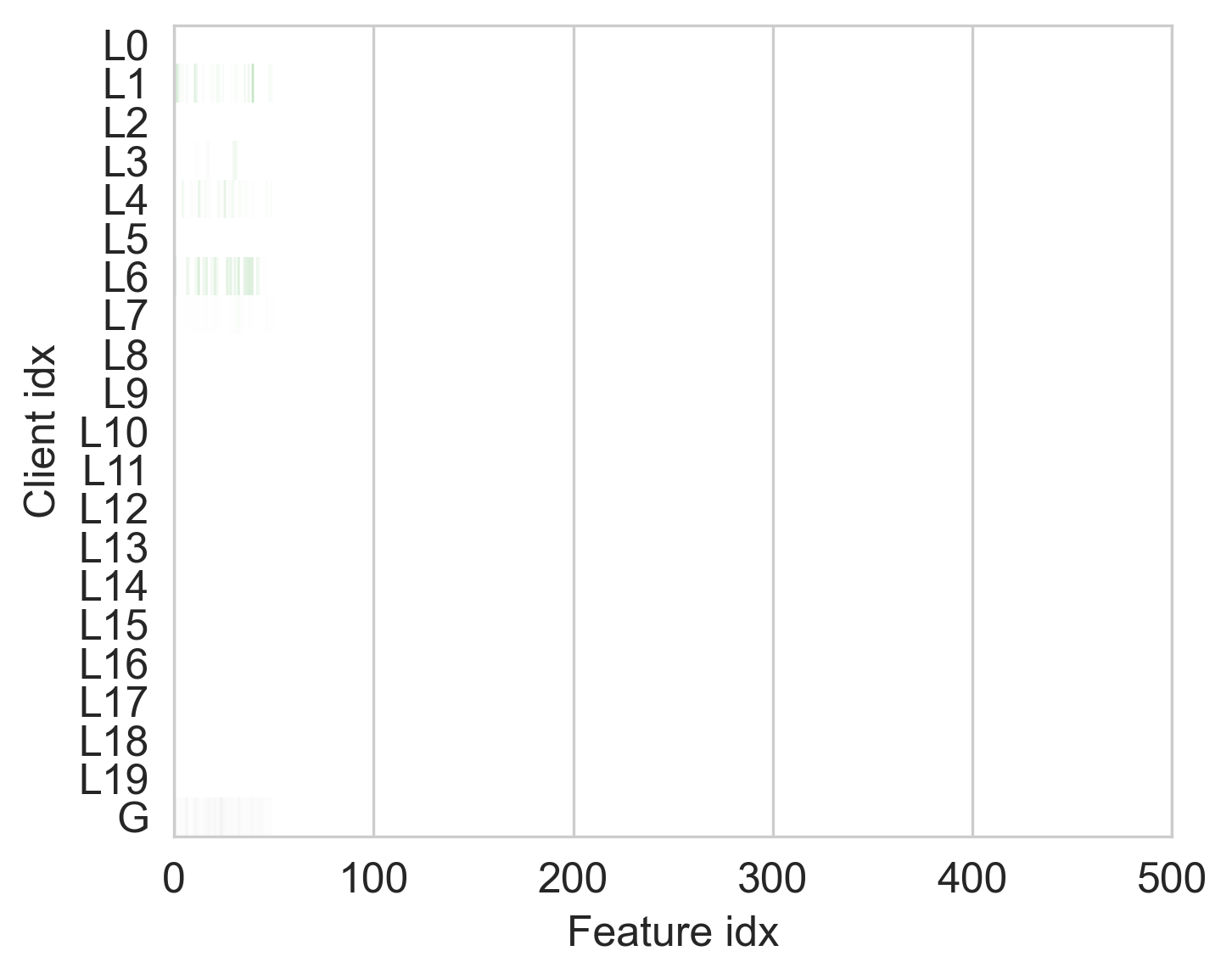}
        \caption{Prototypes of class \#2 (original)}
    \end{subfigure}
    \quad
    \begin{subfigure}{0.33\textwidth}
        \centering
        \includegraphics[width=\textwidth]{figures/FedProto_CIFAR10_CSR_1_global_prototypes_2_binary.png}
        \caption{Prototypes of class \#2 (binary)}
    \end{subfigure}

    \begin{subfigure}{0.33\textwidth}
        \centering
        \includegraphics[width=\textwidth]{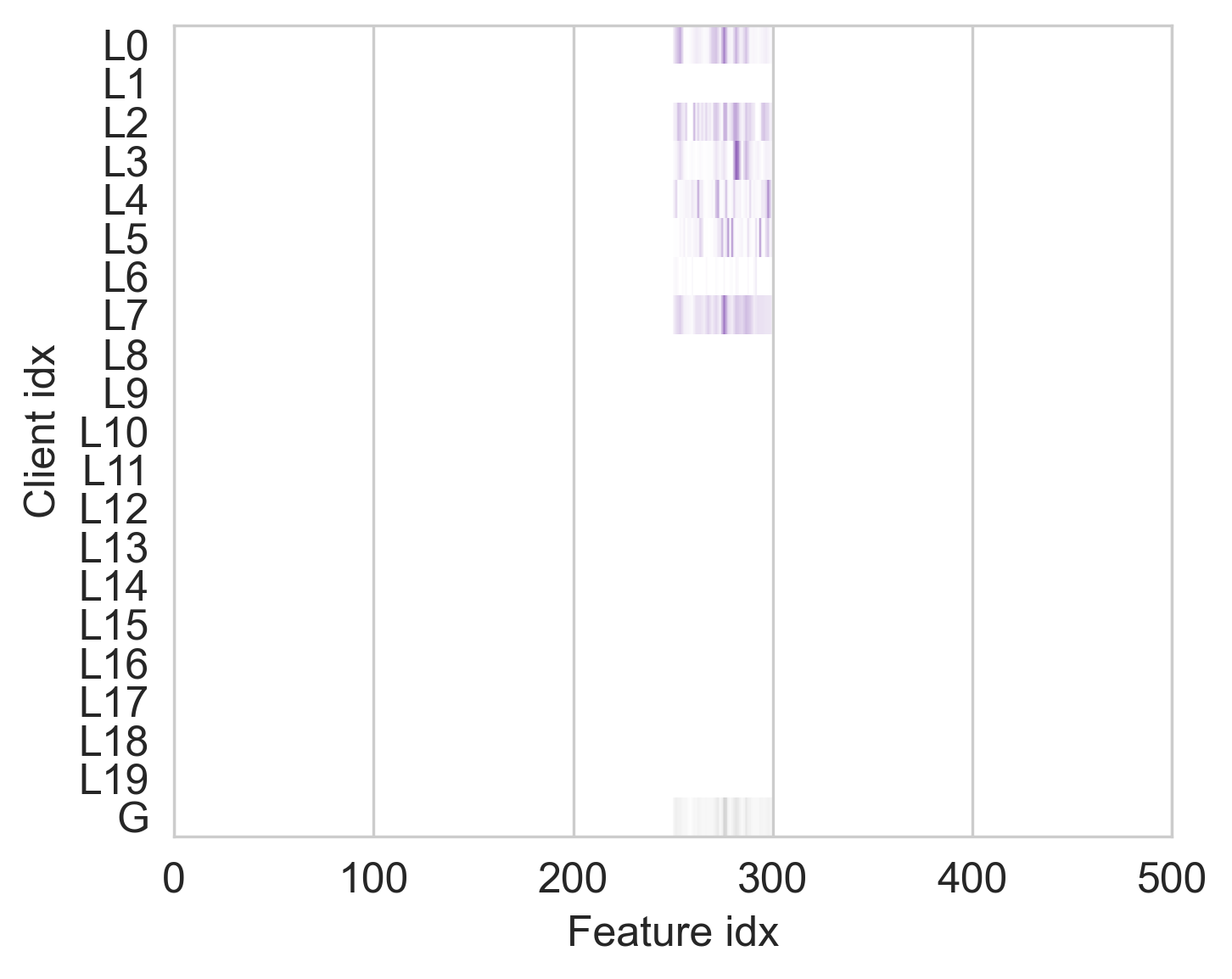}
        \caption{Prototypes of class \#4 (original)}
    \end{subfigure}
    \quad
    \begin{subfigure}{0.33\textwidth}
        \centering
        \includegraphics[width=\textwidth]{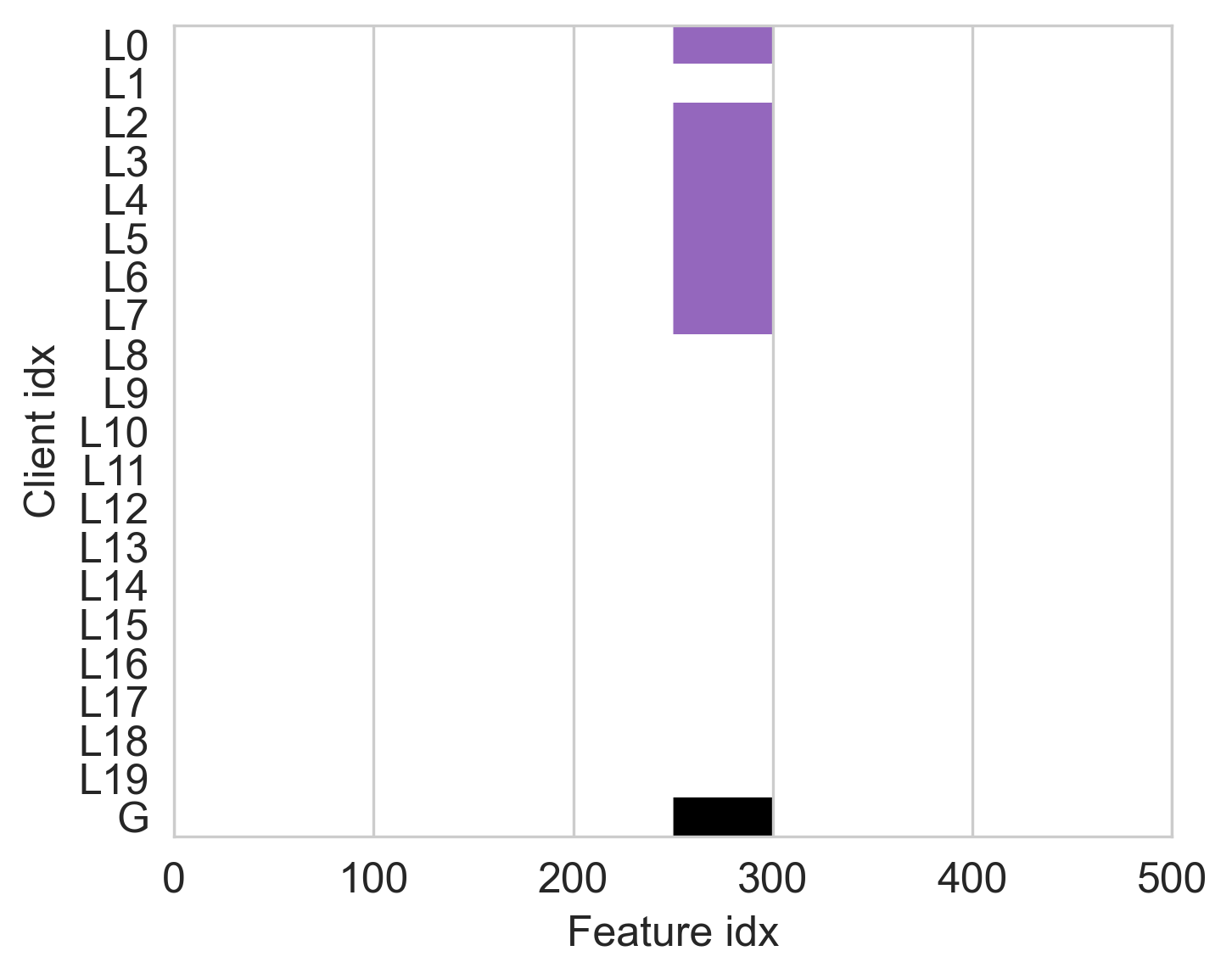}
        \caption{Prototypes of class \#4 (binary)}
    \end{subfigure}
    
    \begin{subfigure}{0.33\textwidth}
        \centering
        \includegraphics[width=\textwidth]{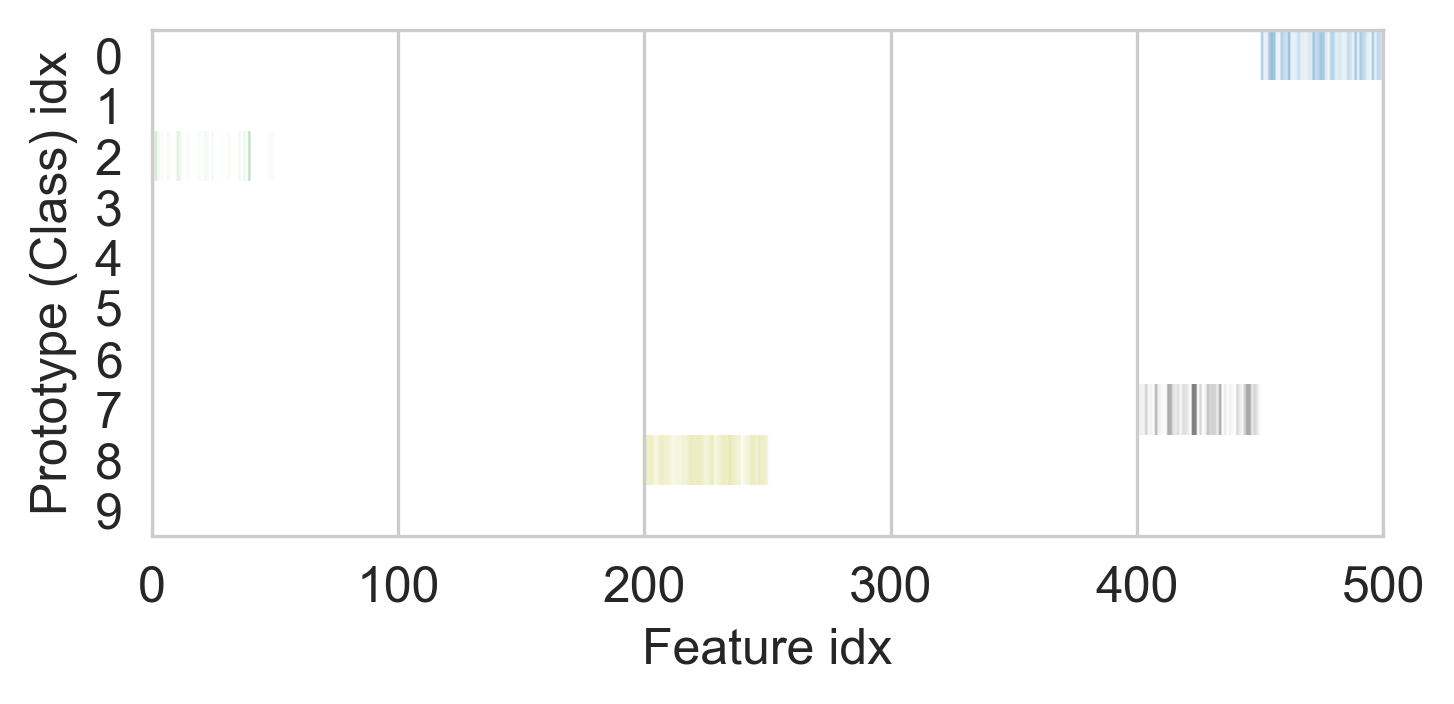}
        \caption{Prototypes of client \#10 (original)}
    \end{subfigure}
    \quad
    \begin{subfigure}{0.33\textwidth}
        \centering
        \includegraphics[width=\textwidth]{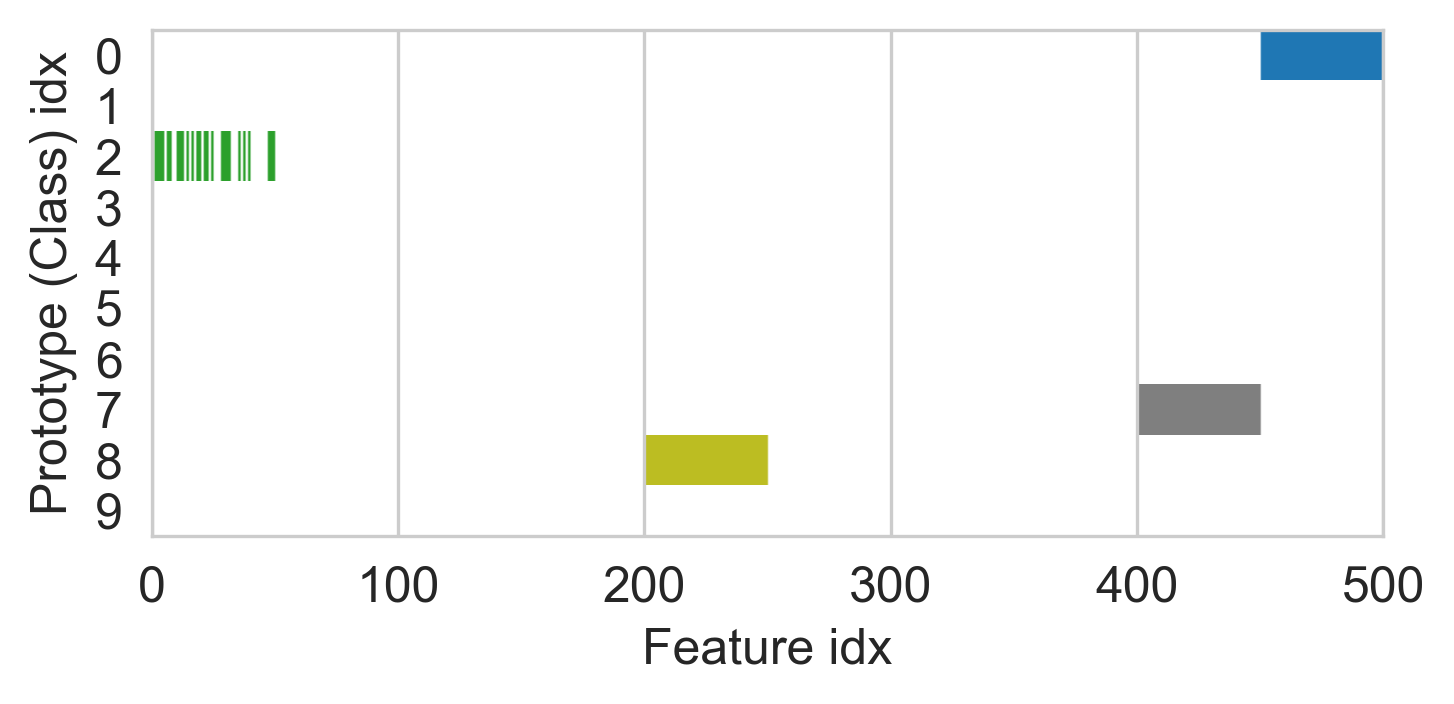}
        \caption{Prototypes of client \#10 (binary)}
    \end{subfigure}

    \begin{subfigure}{0.33\textwidth}
        \centering
        \includegraphics[width=\textwidth]{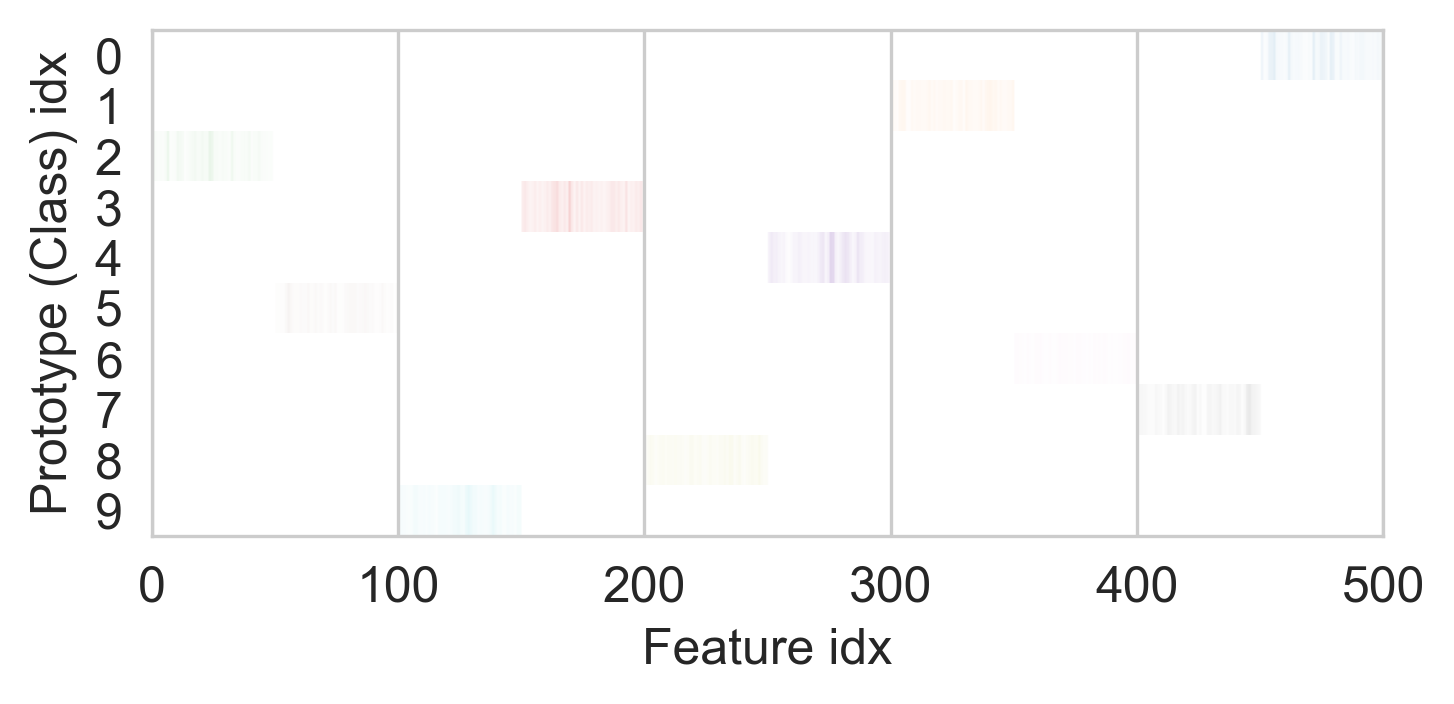}
        \caption{Global prototypes (original)}
    \end{subfigure}
    \quad
    \begin{subfigure}{0.33\textwidth}
        \centering
        \includegraphics[width=\textwidth]{figures/FedProto_CIFAR10_CSR_1_global_prototypes_binary.png}
        \caption{Global prototypes (binary)}
    \end{subfigure}
    \caption{Prototype comparison of FedProto with Class-wise Prototype Sparsification (CPS) for the CIFAR-10 dataset. The dimension $s$ is 50 for CPS.}
    \label{fig:heatmaps_prototype_cifar10_cps50}
    \vspace{-10pt}
\end{figure}

\begin{figure}[ht]
    \centering
    \begin{subfigure}{0.33\textwidth}
        \centering
        \includegraphics[width=\textwidth]{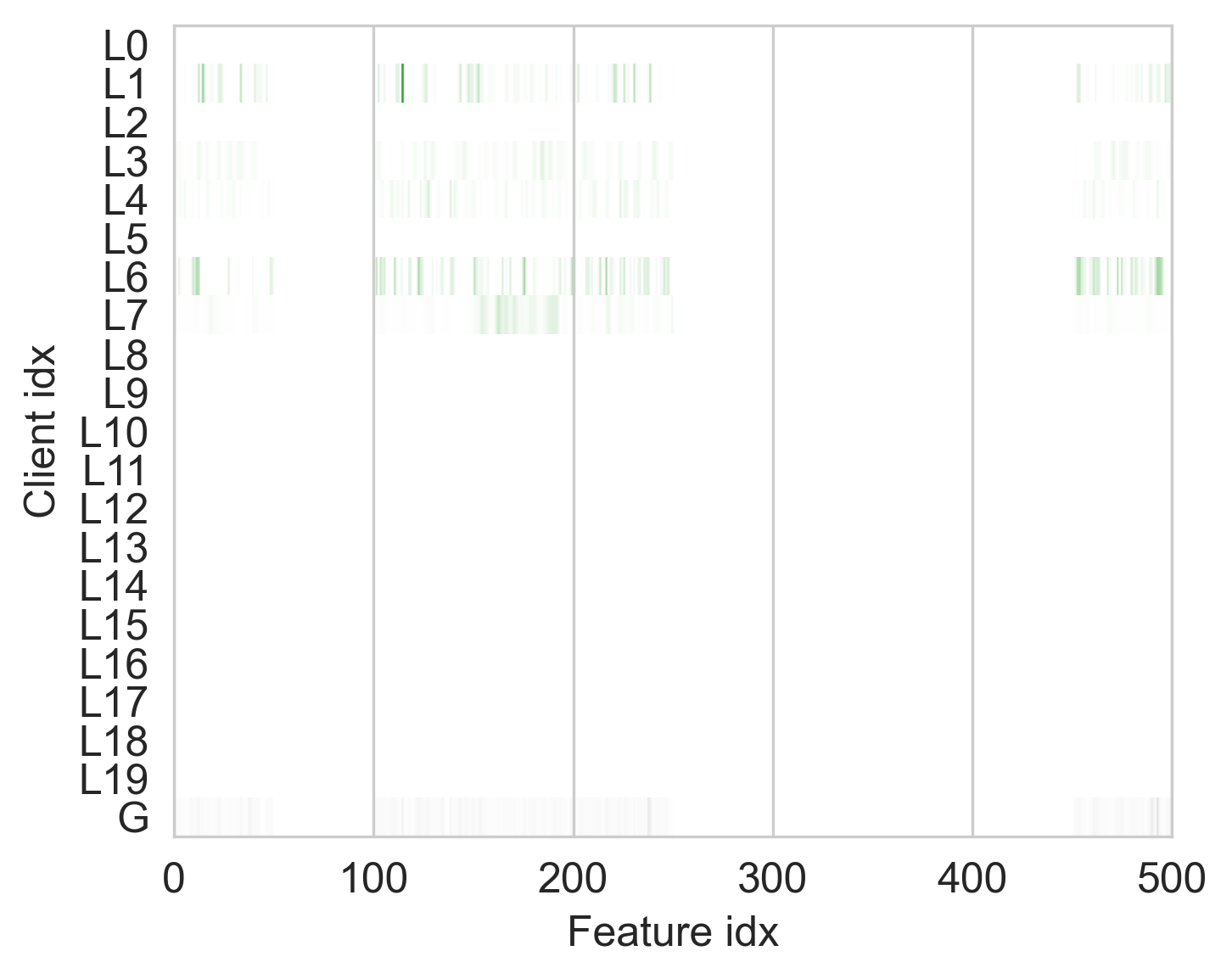}
        \caption{Prototypes of class \#2 (original)}
    \end{subfigure}
    \quad
    \begin{subfigure}{0.33\textwidth}
        \centering
        \includegraphics[width=\textwidth]{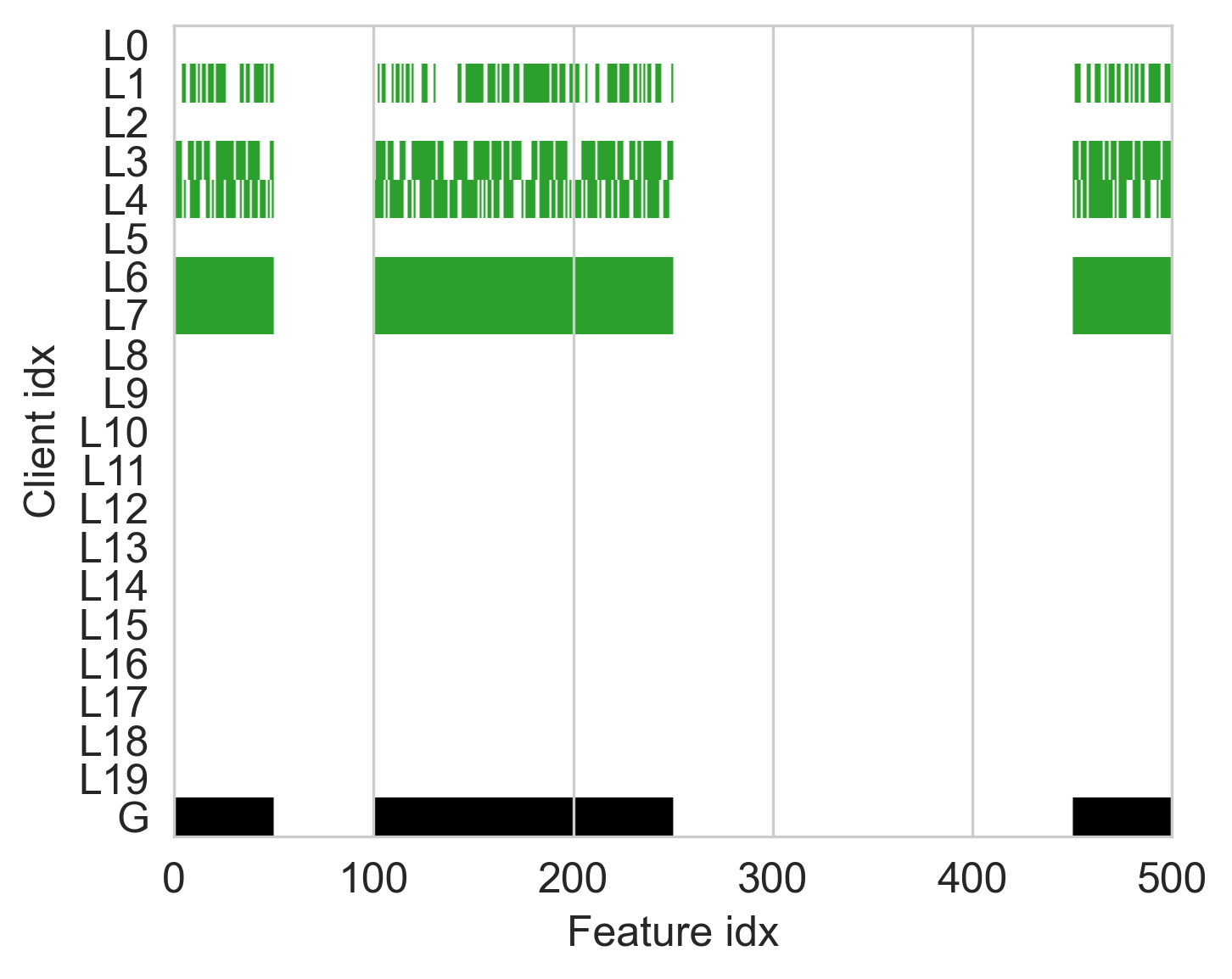}
        \caption{Prototypes of class \#2 (binary)}
    \end{subfigure}

    \begin{subfigure}{0.33\textwidth}
        \centering
        \includegraphics[width=\textwidth]{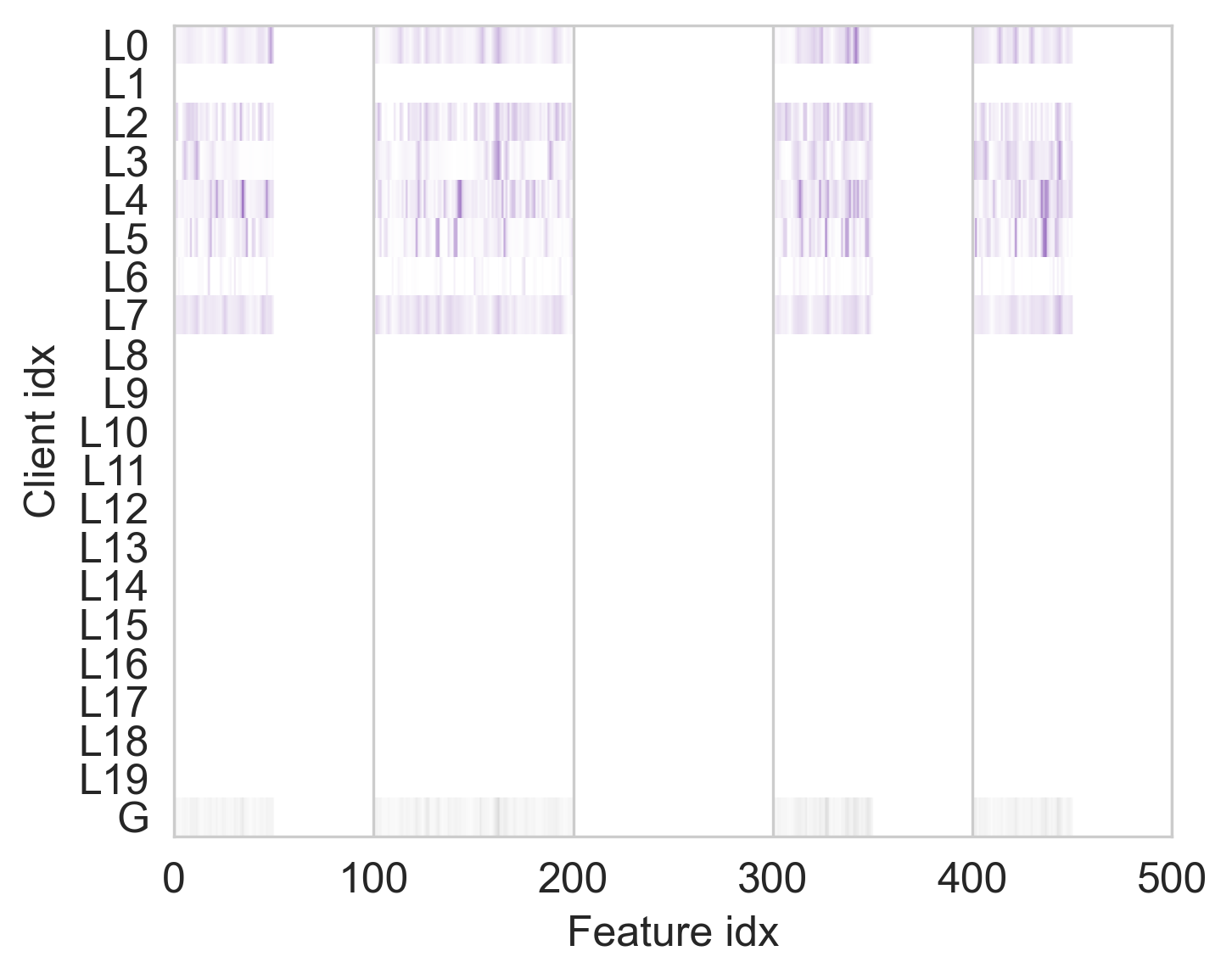}
        \caption{Prototypes of class \#4 (original)}
    \end{subfigure}
    \quad
    \begin{subfigure}{0.33\textwidth}
        \centering
        \includegraphics[width=\textwidth]{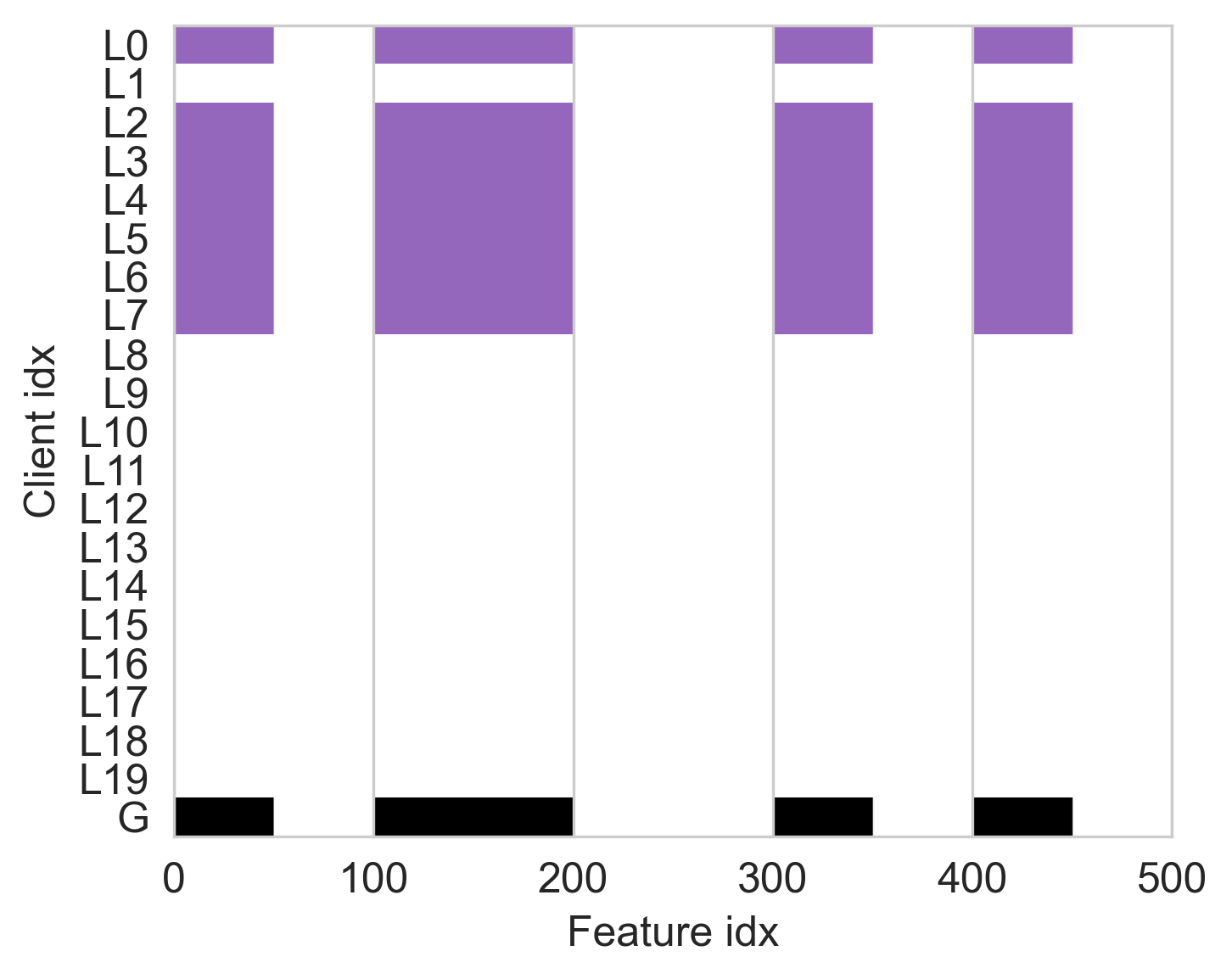}
        \caption{Prototypes of class \#4 (binary)}
    \end{subfigure}
    
    \begin{subfigure}{0.33\textwidth}
        \centering
        \includegraphics[width=\textwidth]{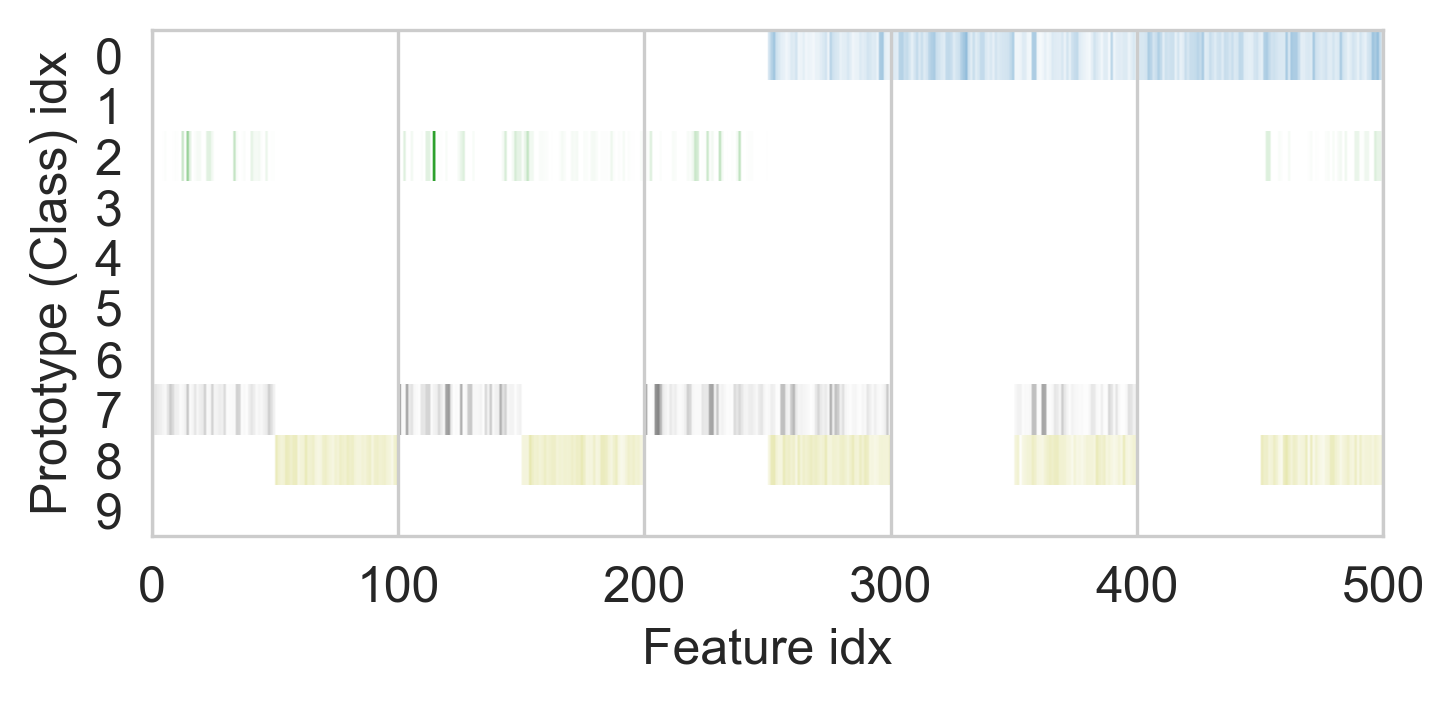}
        \caption{Prototypes of client \#10 (original)}
    \end{subfigure}
    \quad
    \begin{subfigure}{0.33\textwidth}
        \centering
        \includegraphics[width=\textwidth]{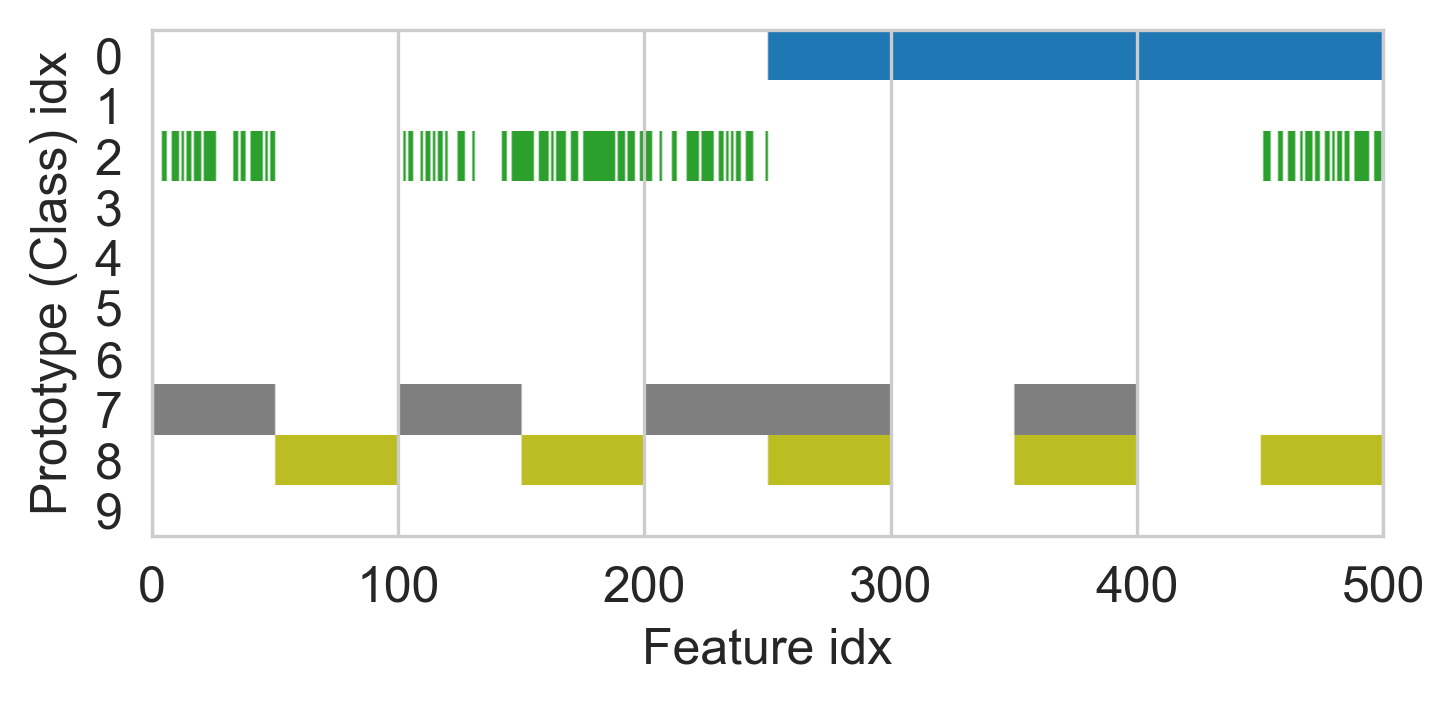}
        \caption{Prototypes of client \#10 (binary)}
    \end{subfigure}

    \begin{subfigure}{0.33\textwidth}
        \centering
        \includegraphics[width=\textwidth]{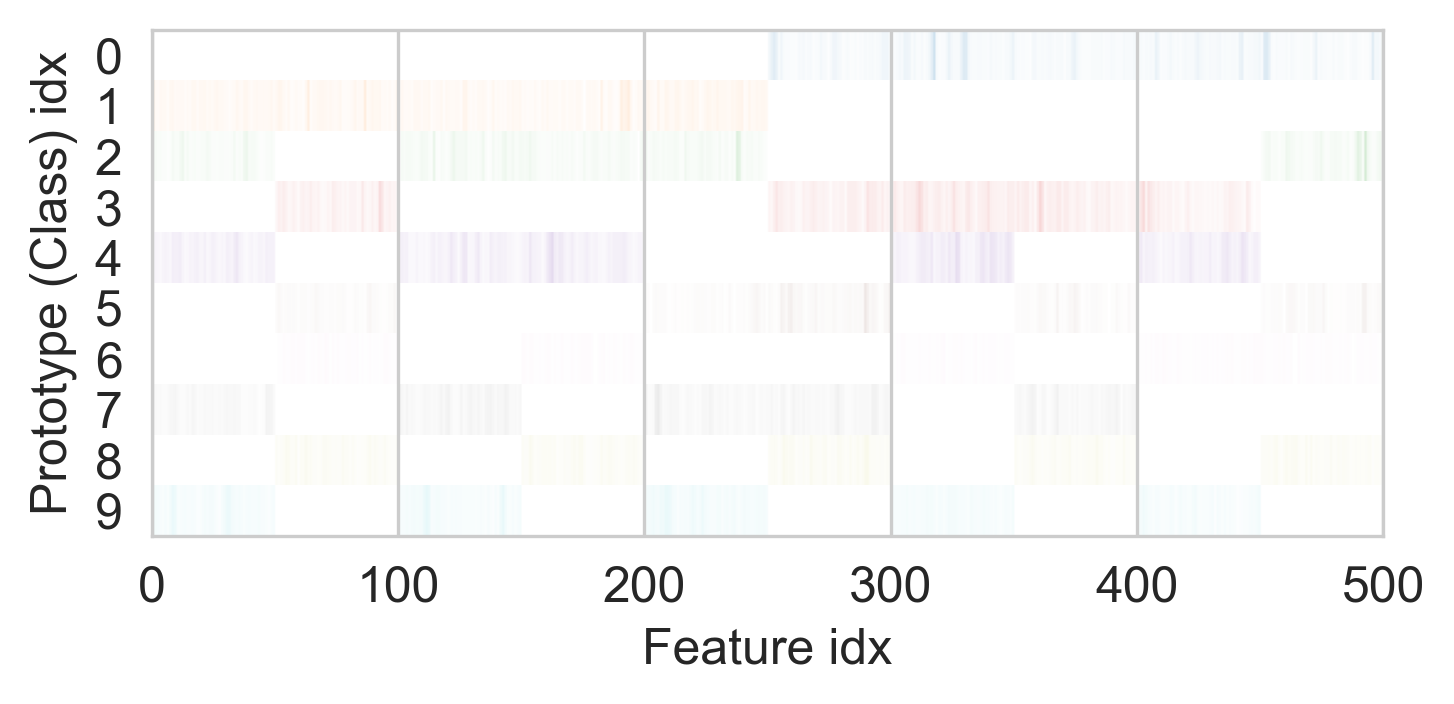}
        \caption{Global prototypes (original)}
    \end{subfigure}
    \quad
    \begin{subfigure}{0.33\textwidth}
        \centering
        \includegraphics[width=\textwidth]{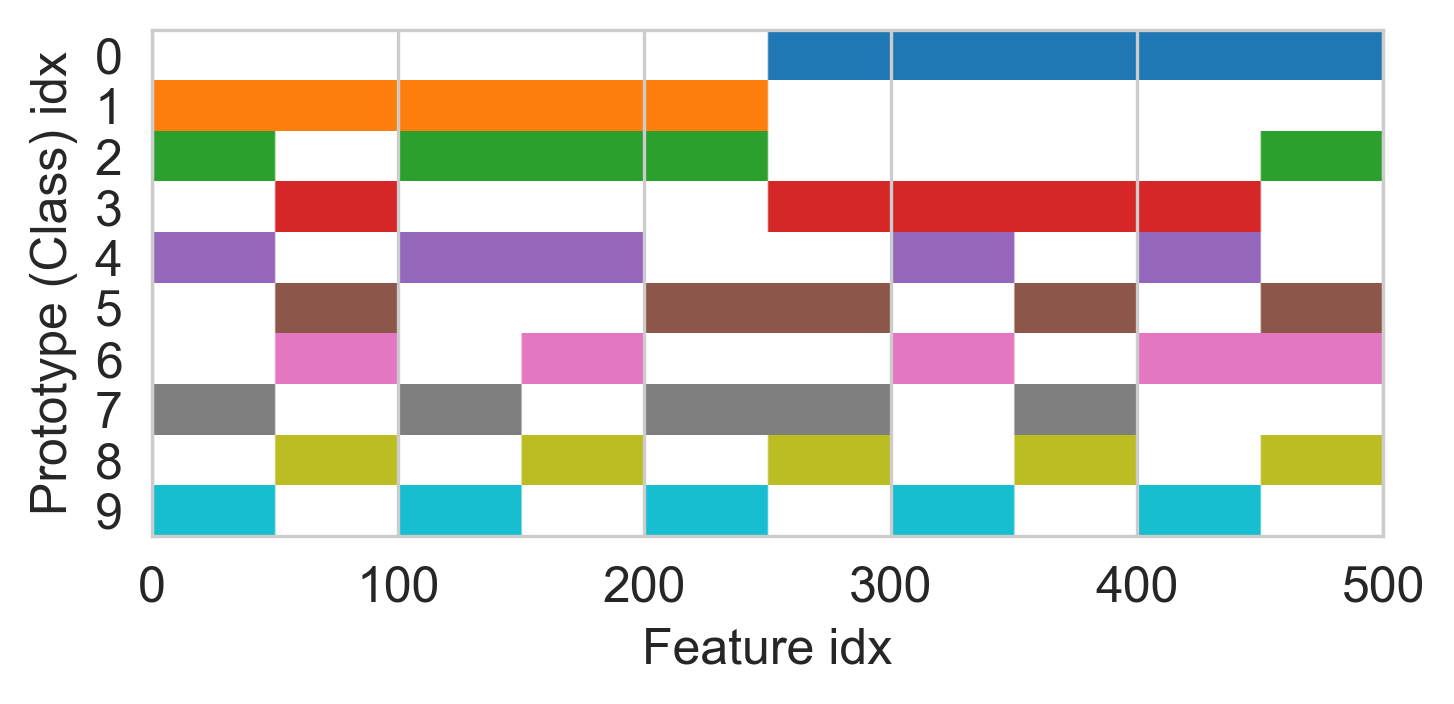}
        \caption{Global prototypes (binary)}
    \end{subfigure}
    \caption{Prototype comparison of FedProto with Class-wise Prototype Sparsification (CPS) for the CIFAR-10 dataset. The dimension $s$ is 250 for CPS.}
    \label{fig:heatmaps_prototype_cifar10_cps250}
    \vspace{-10pt}
\end{figure}

\begin{figure}[ht]
    \centering
    \begin{subfigure}{0.33\textwidth}
        \centering
        \includegraphics[width=\textwidth]{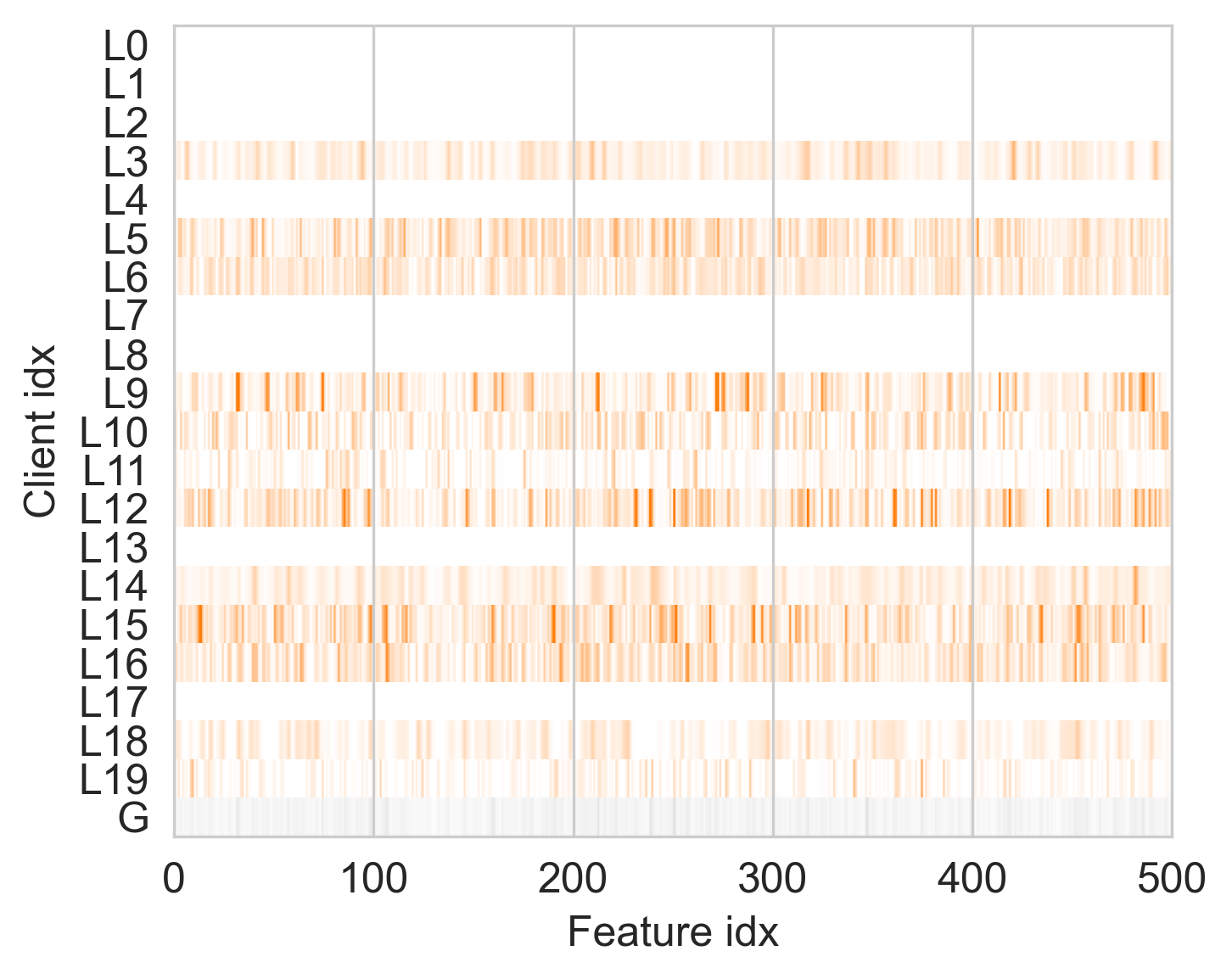}
        \caption{Prototypes of class \#22 (original)}
    \end{subfigure}
    \quad
    \begin{subfigure}{0.33\textwidth}
        \centering
        \includegraphics[width=\textwidth]{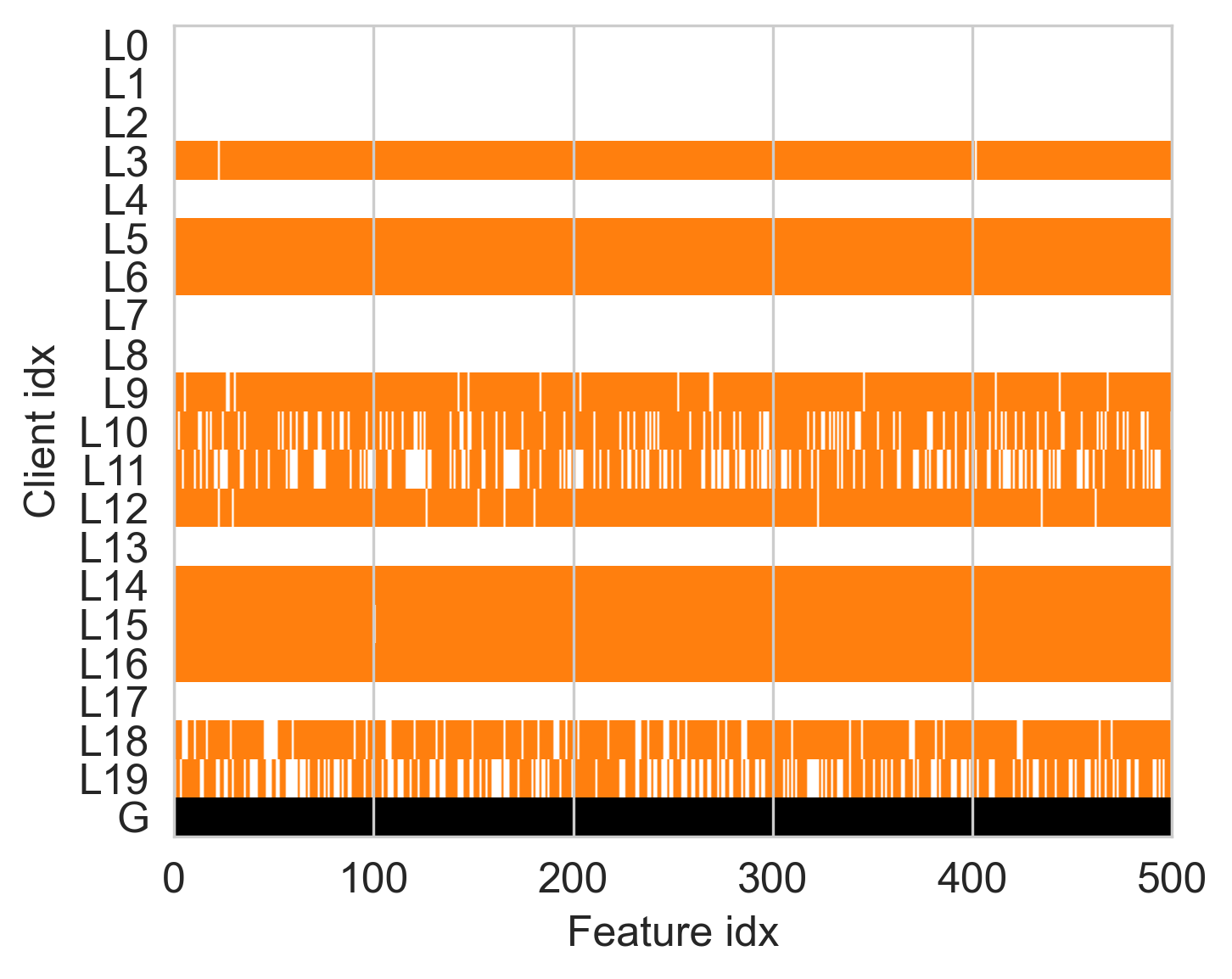}
        \caption{Prototypes of class \#22 (binary)}
    \end{subfigure}
    
    \begin{subfigure}{0.33\textwidth}
        \centering
        \includegraphics[width=\textwidth]{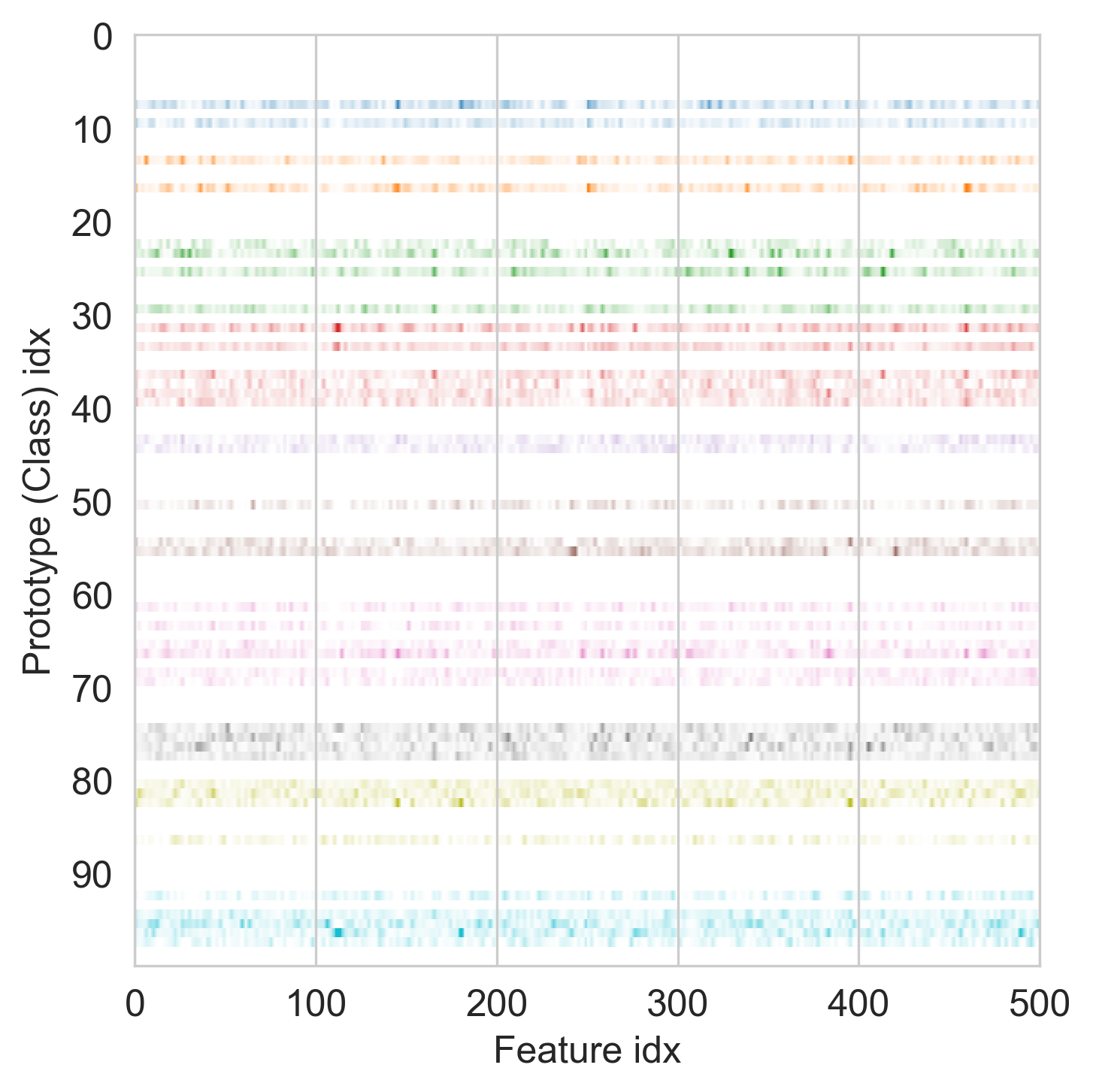}
        \caption{Prototypes of client \#8 (original)}
    \end{subfigure}
    \quad
    \begin{subfigure}{0.33\textwidth}
        \centering
        \includegraphics[width=\textwidth]{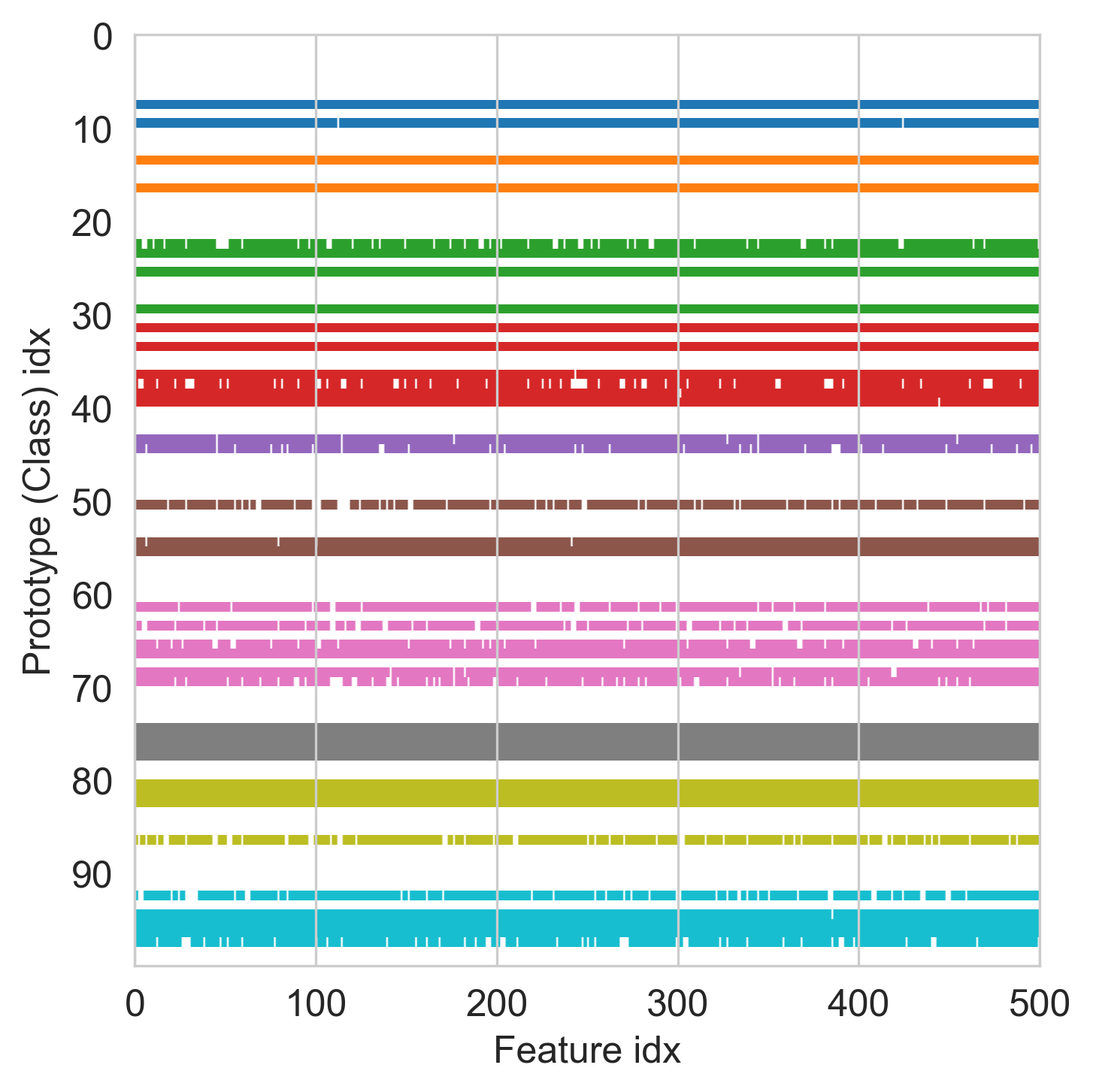}
        \caption{Prototypes of client \#8 (binary)}
    \end{subfigure}

    \begin{subfigure}{0.33\textwidth}
        \centering
        \includegraphics[width=\textwidth]{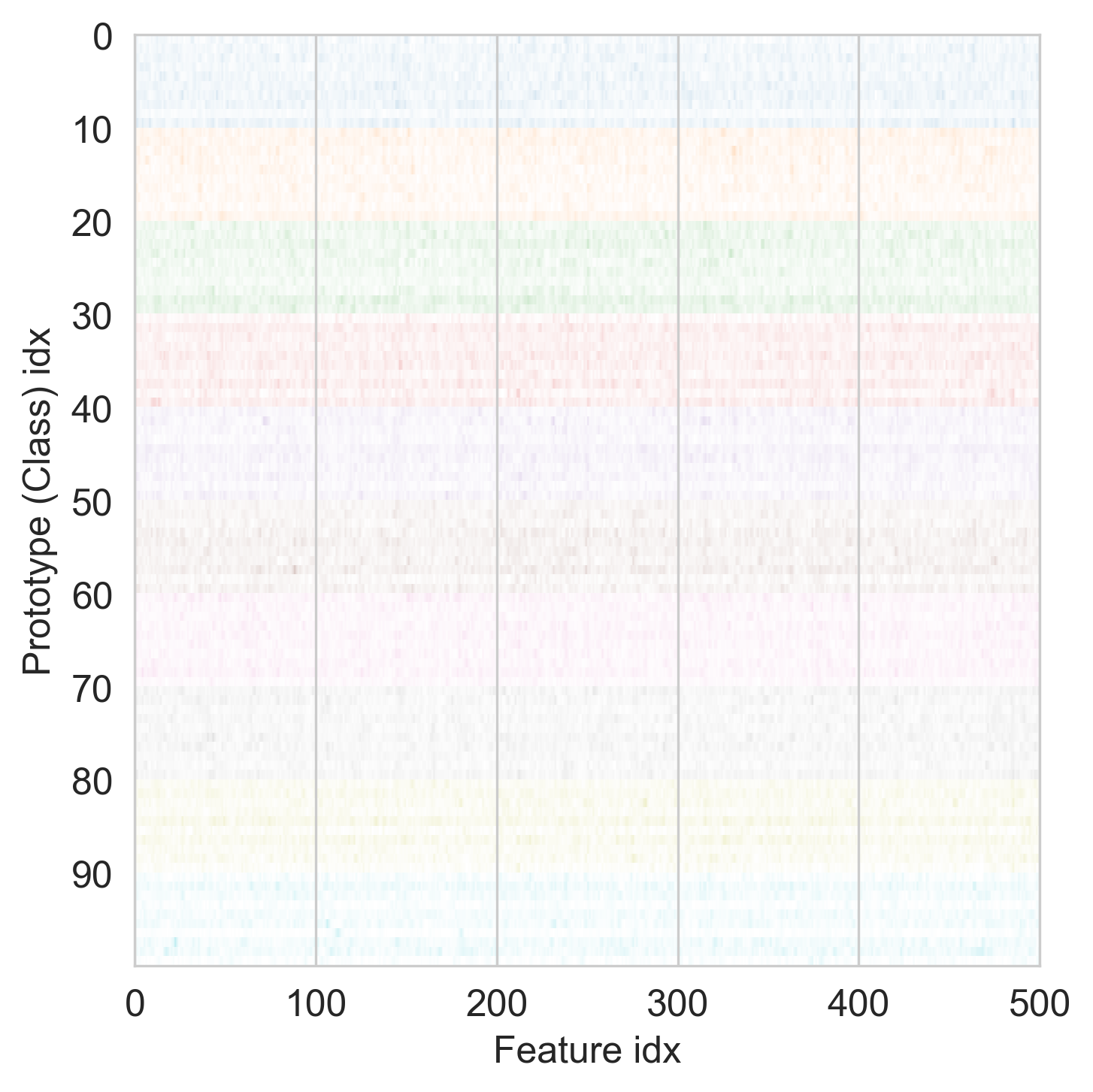}
        \caption{Global prototypes (original)}
    \end{subfigure}
    \quad
    \begin{subfigure}{0.33\textwidth}
        \centering
        \includegraphics[width=\textwidth]{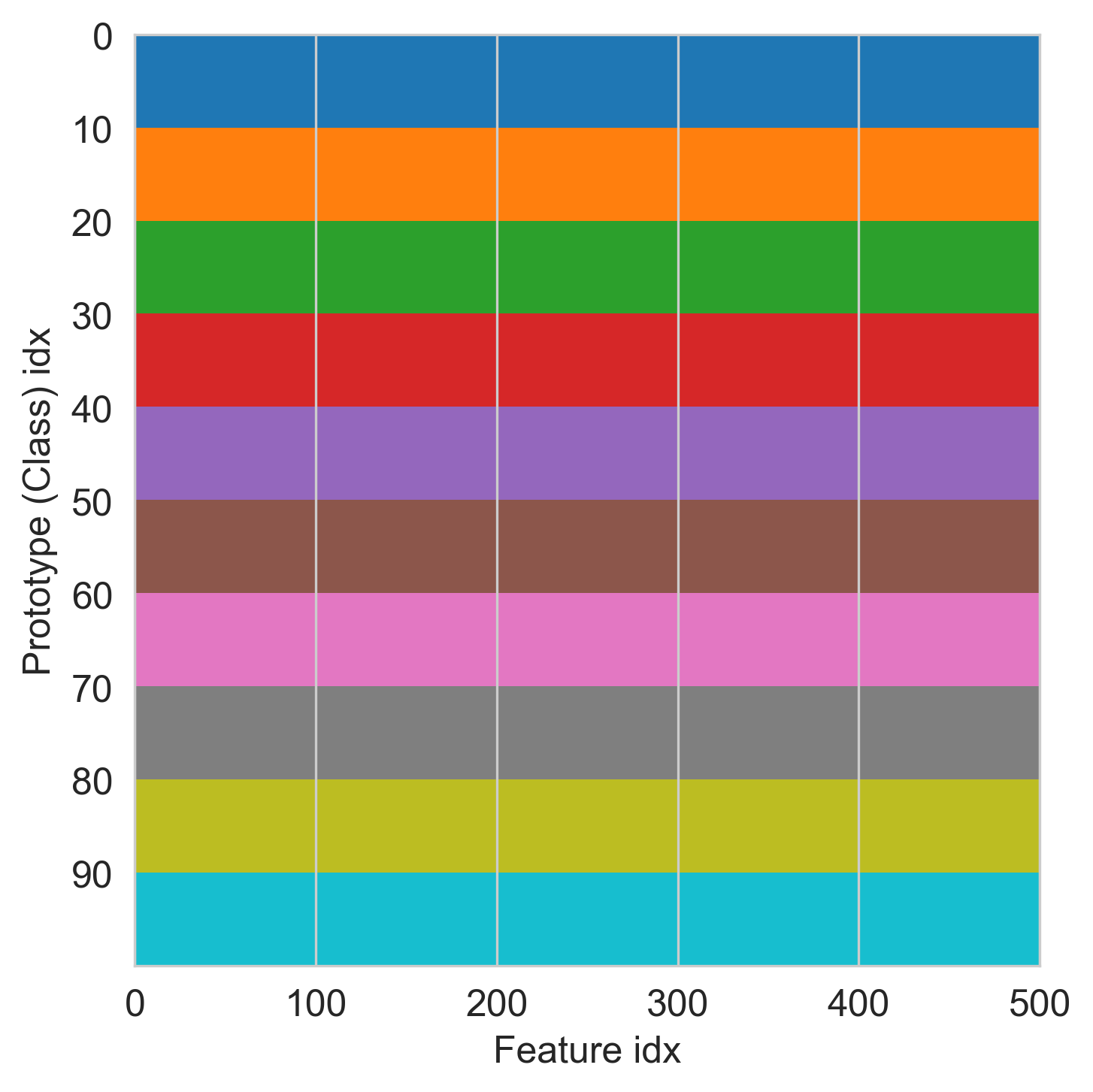}
        \caption{Global prototypes (binary)}
    \end{subfigure}
    \caption{Prototype comparison of FedProto without Class-wise Prototype Sparsification (CPS) for the CIFAR-100 dataset.}
    \label{fig:heatmaps_prototype_cifar100_original}
    \vspace{-10pt}
\end{figure}

\begin{figure}[t]
    \centering
    \begin{subfigure}{0.33\textwidth}
        \centering
        \includegraphics[width=\textwidth]{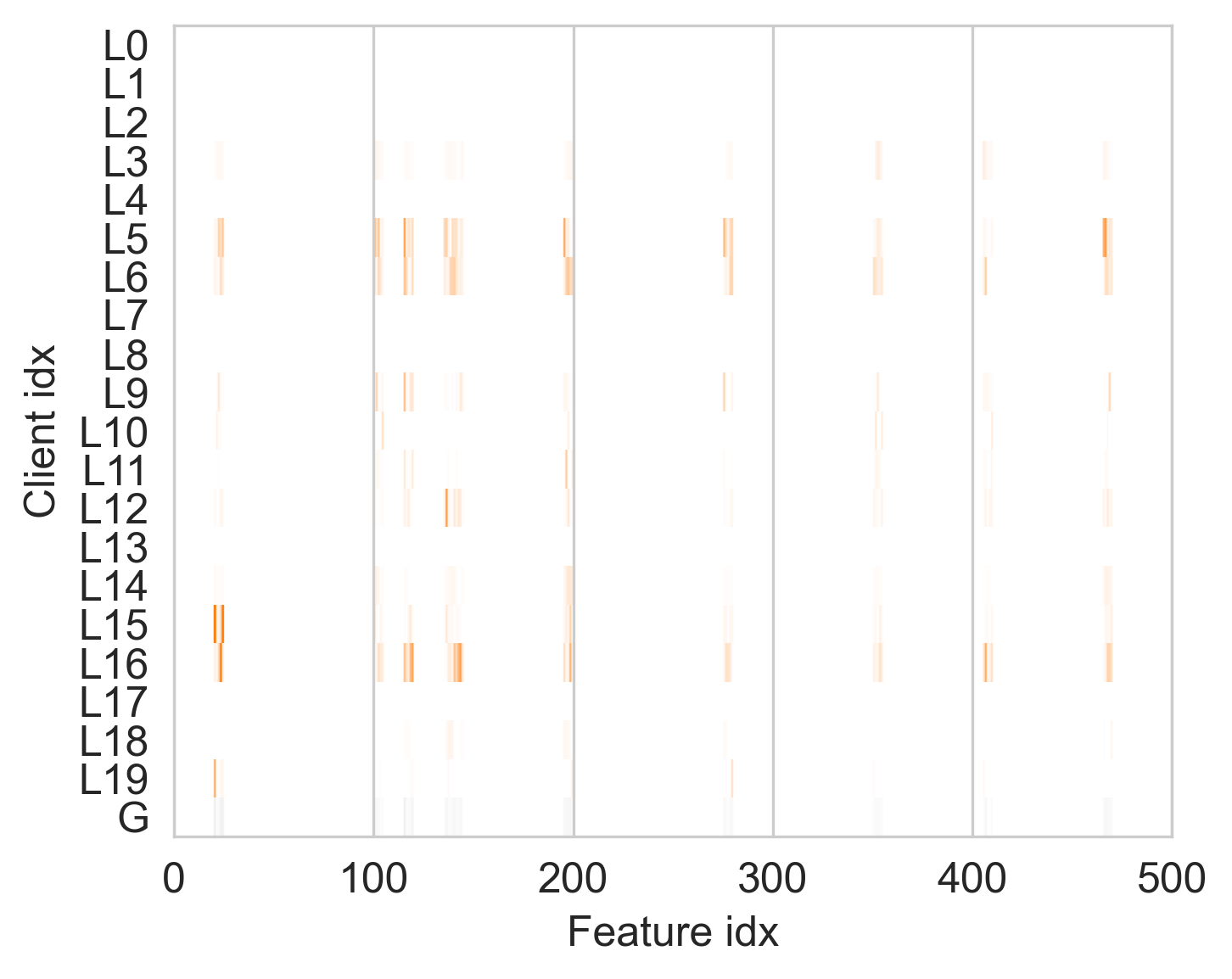}
        \caption{Prototypes of class \#22 (original)}
    \end{subfigure}
    \quad
    \begin{subfigure}{0.33\textwidth}
        \centering
        \includegraphics[width=\textwidth]{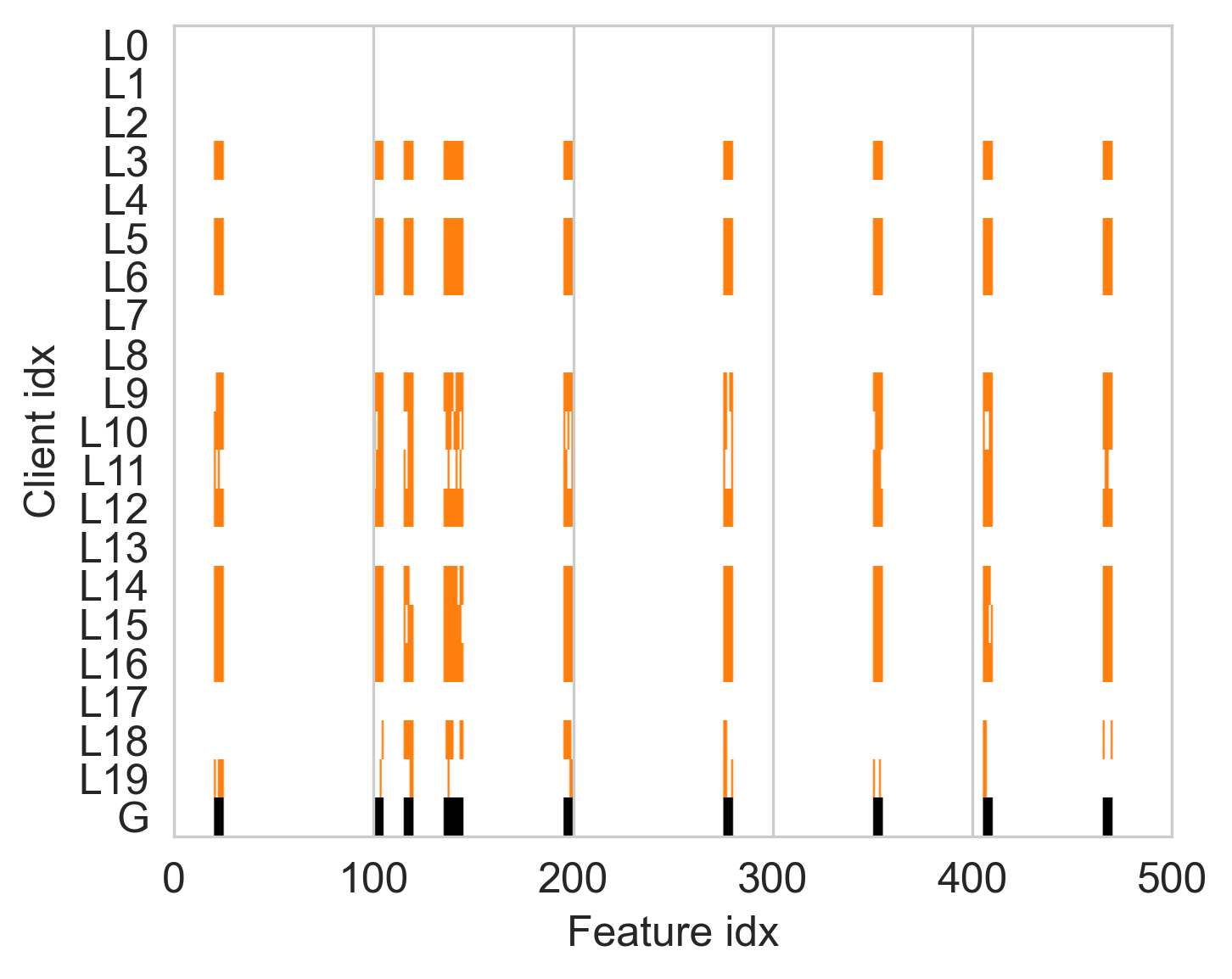}
        \caption{Prototypes of class \#22 (binary)}
    \end{subfigure}
    
    \begin{subfigure}{0.33\textwidth}
        \centering
        \includegraphics[width=\textwidth]{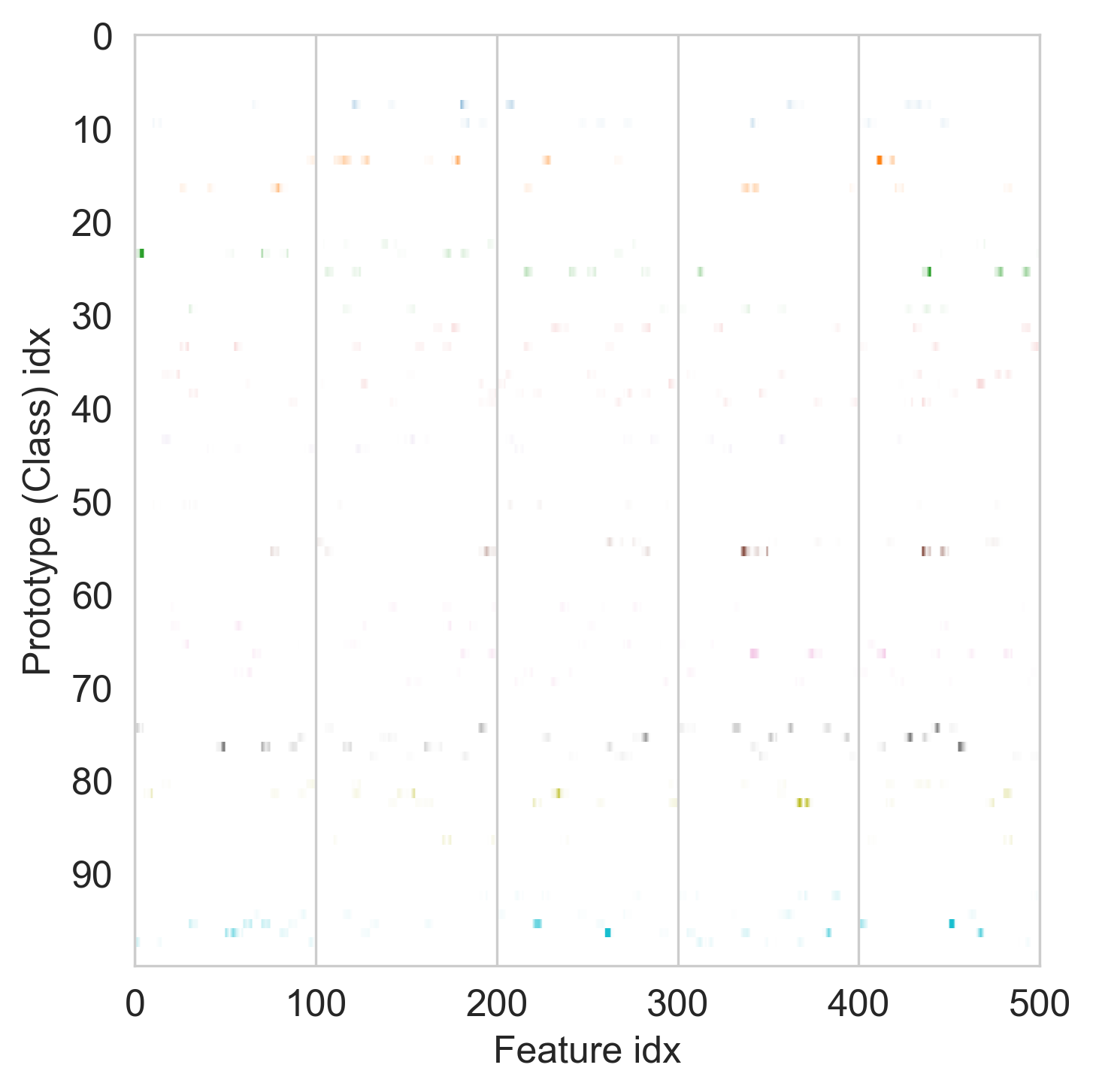}
        \caption{Prototypes of client \#8 (original)}
    \end{subfigure}
    \quad
    \begin{subfigure}{0.33\textwidth}
        \centering
        \includegraphics[width=\textwidth]{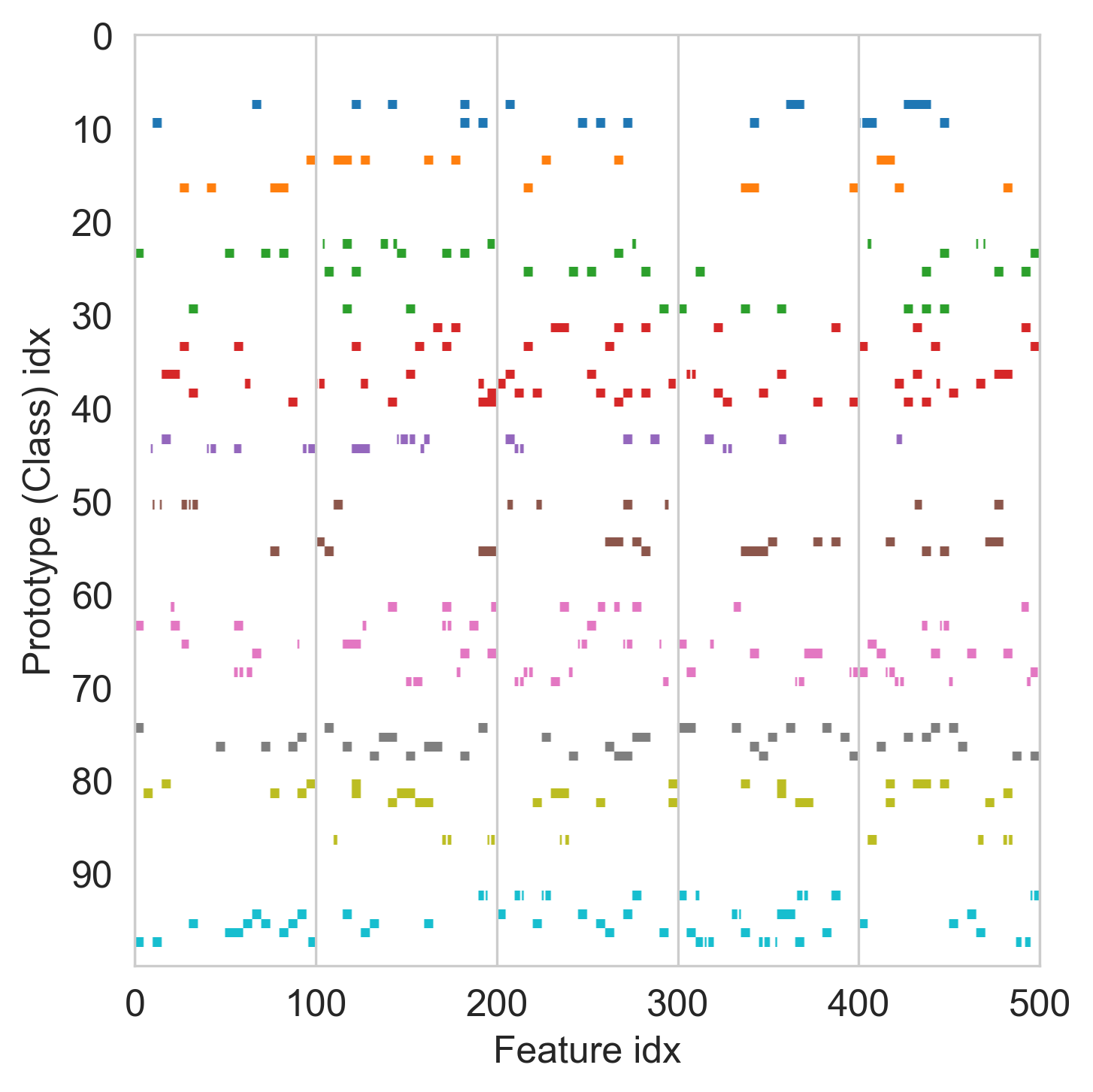}
        \caption{Prototypes of client \#8 (binary)}
    \end{subfigure}

    \begin{subfigure}{0.33\textwidth}
        \centering
        \includegraphics[width=\textwidth]{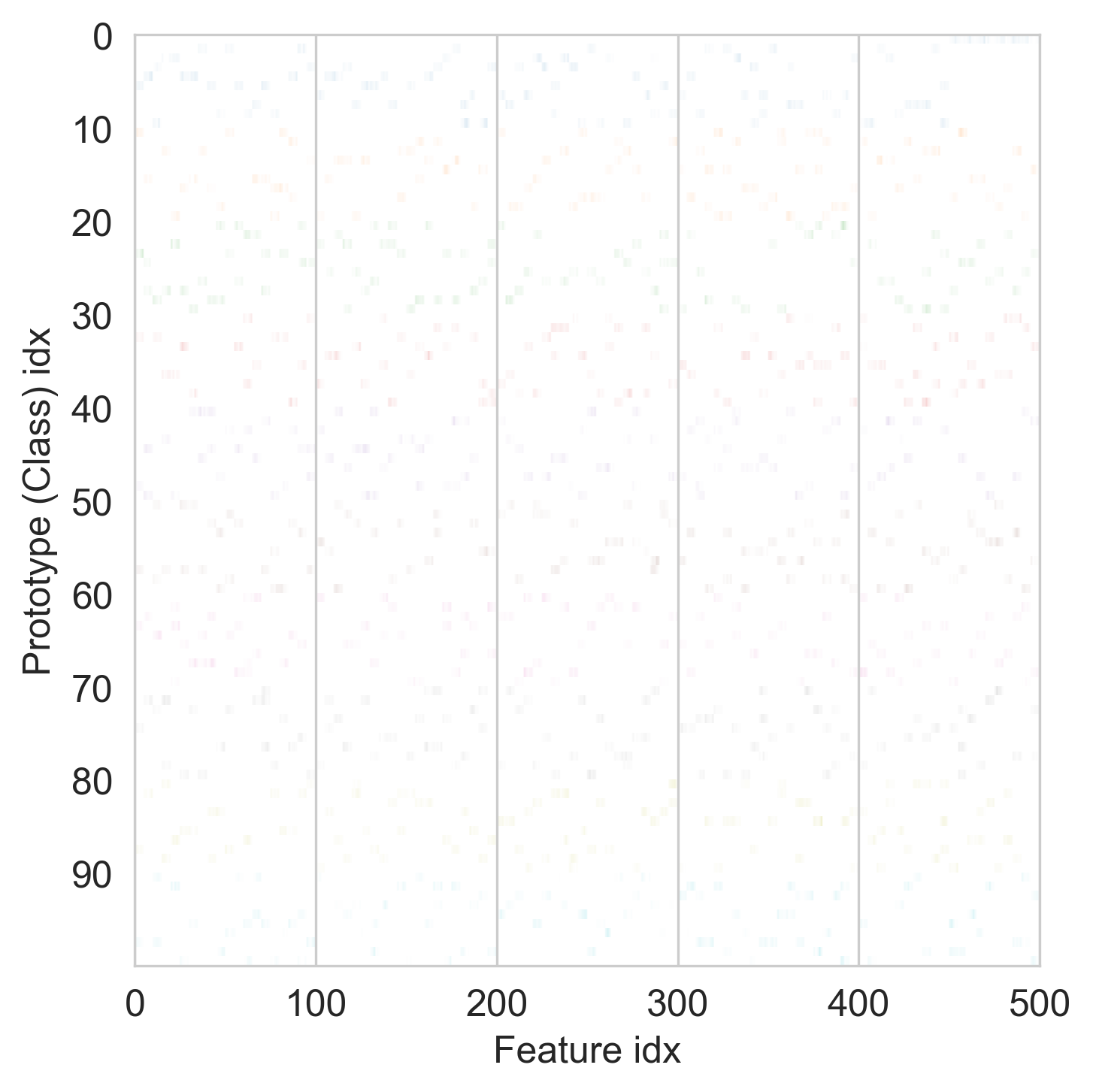}
        \caption{Global prototypes (original)}
    \end{subfigure}
    \quad
    \begin{subfigure}{0.33\textwidth}
        \centering
        \includegraphics[width=\textwidth]{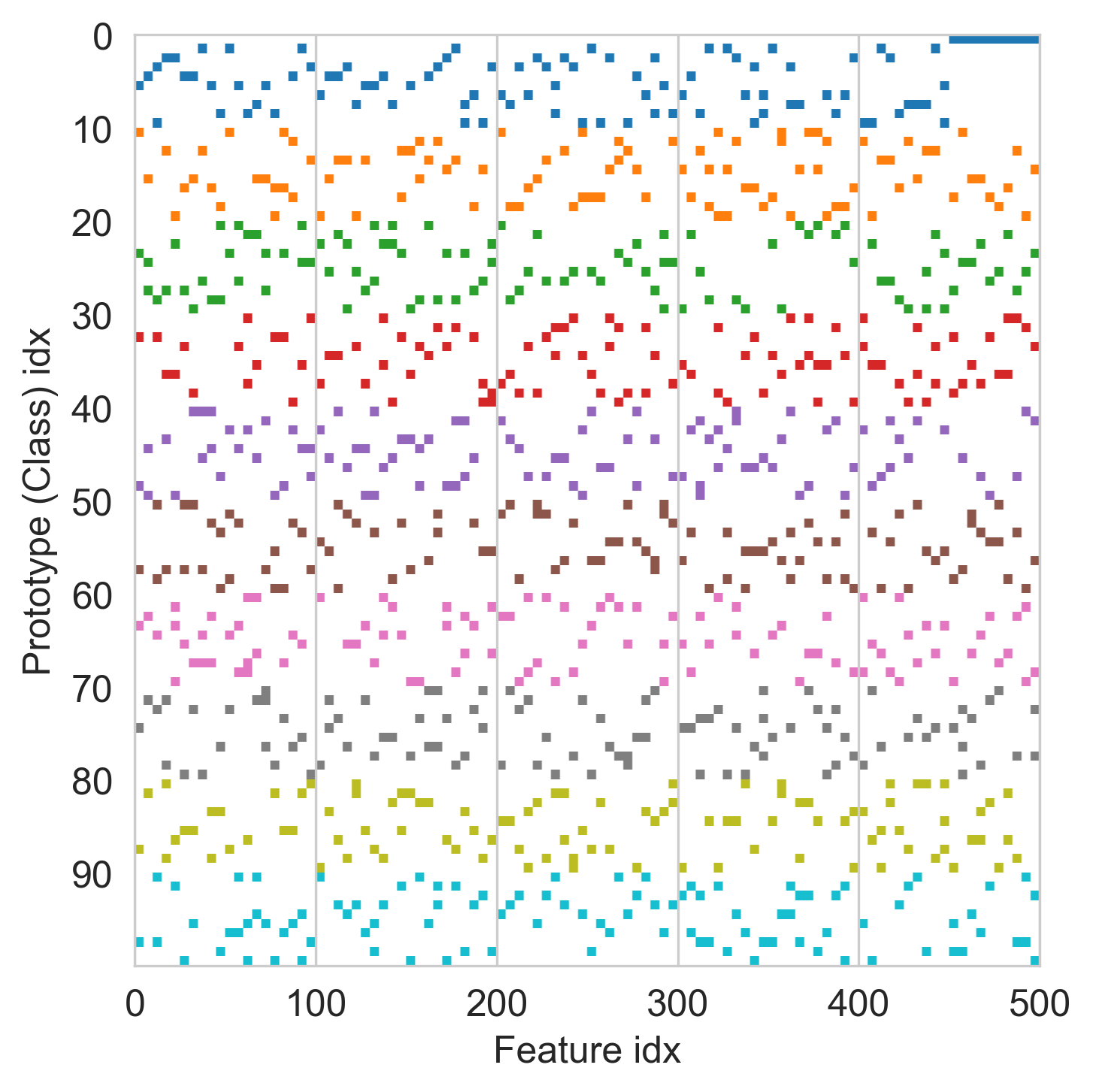}
        \caption{Global prototypes (binary)}
    \end{subfigure}
    \caption{Prototype comparison of FedProto with Class-wise Prototype Sparsification (CPS) for the CIFAR-100 dataset. The dimension $s$ is 50 for CPS.}
    \label{fig:heatmaps_prototype_cifar100_cps50}
\end{figure}

\begin{figure}[t]
    \centering
    \begin{subfigure}{0.33\textwidth}
        \centering
        \includegraphics[width=\textwidth]{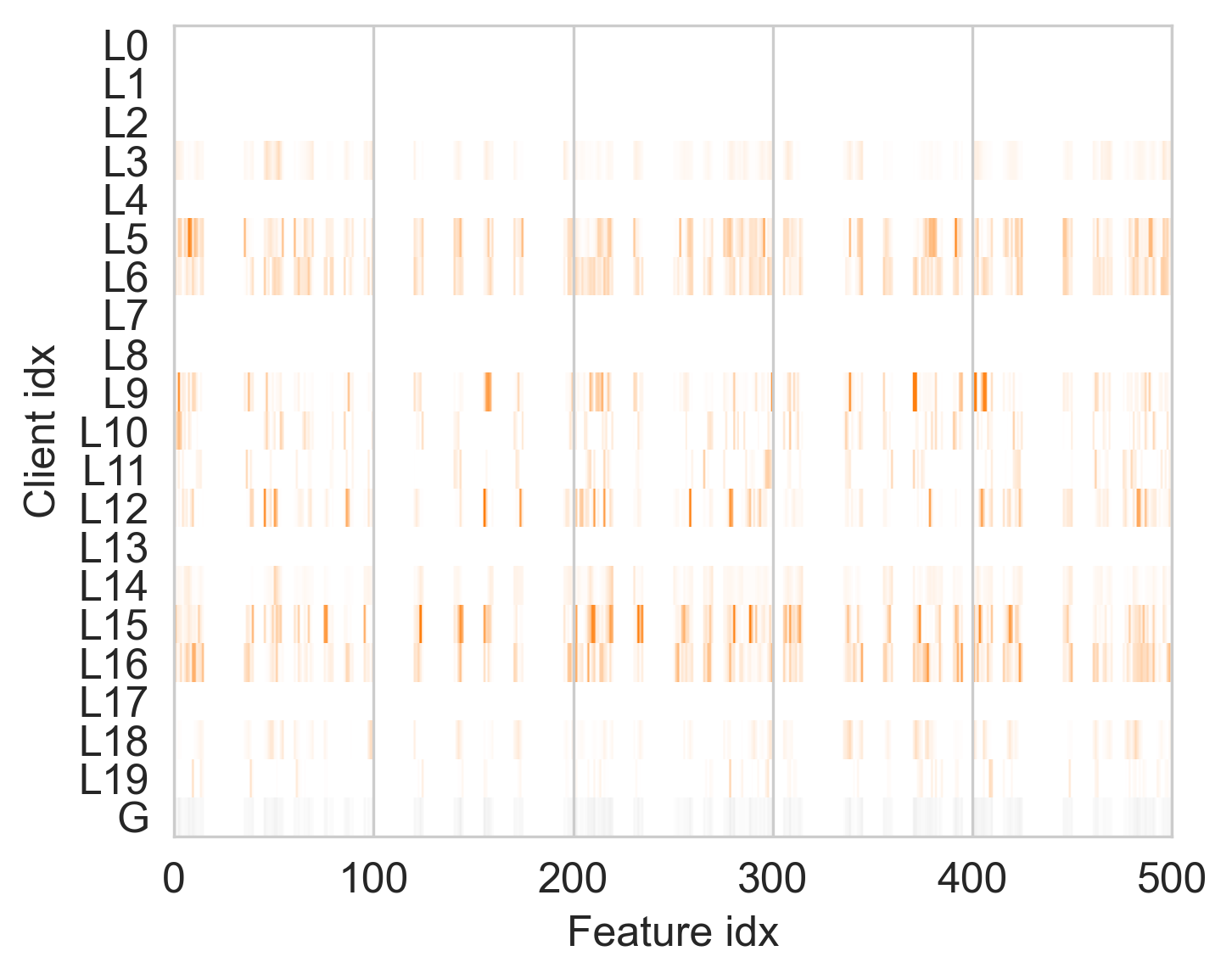}
        \caption{Prototypes of class \#22 (original)}
    \end{subfigure}
    \quad
    \begin{subfigure}{0.33\textwidth}
        \centering
        \includegraphics[width=\textwidth]{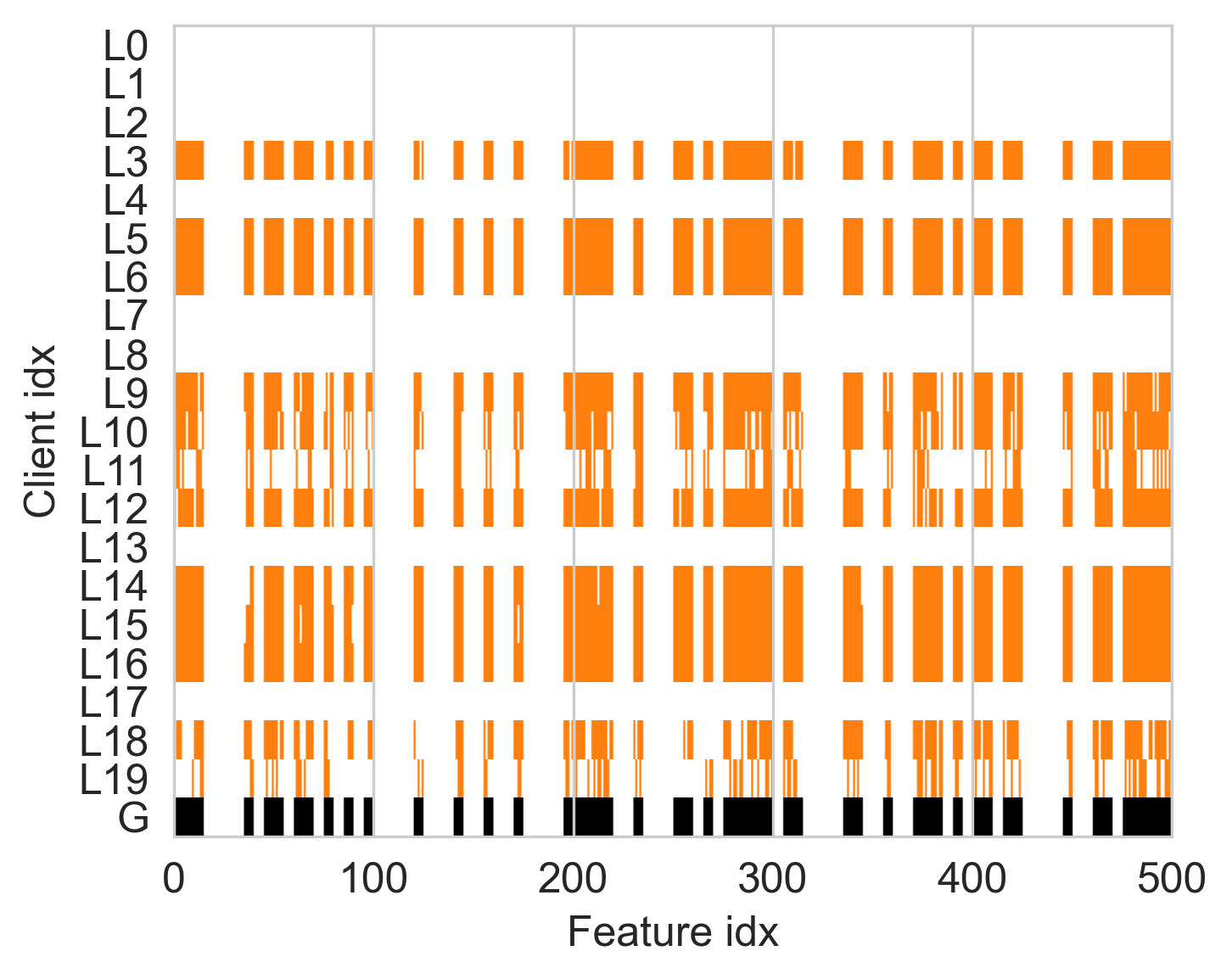}
        \caption{Prototypes of class \#22 (binary)}
    \end{subfigure}
    
    \begin{subfigure}{0.33\textwidth}
        \centering
        \includegraphics[width=\textwidth]{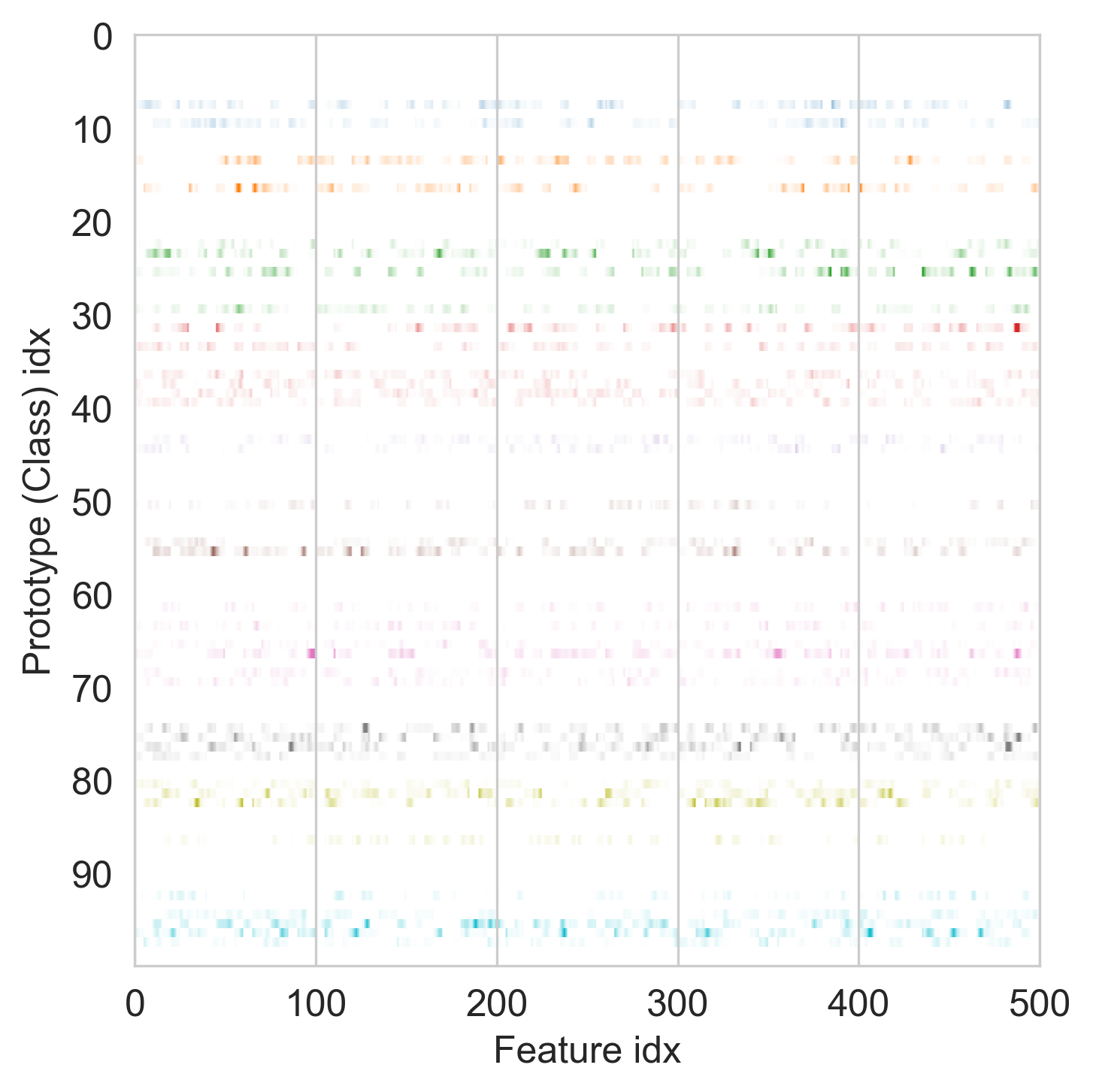}
        \caption{Prototypes of client \#8 (original)}
    \end{subfigure}
    \quad
    \begin{subfigure}{0.33\textwidth}
        \centering
        \includegraphics[width=\textwidth]{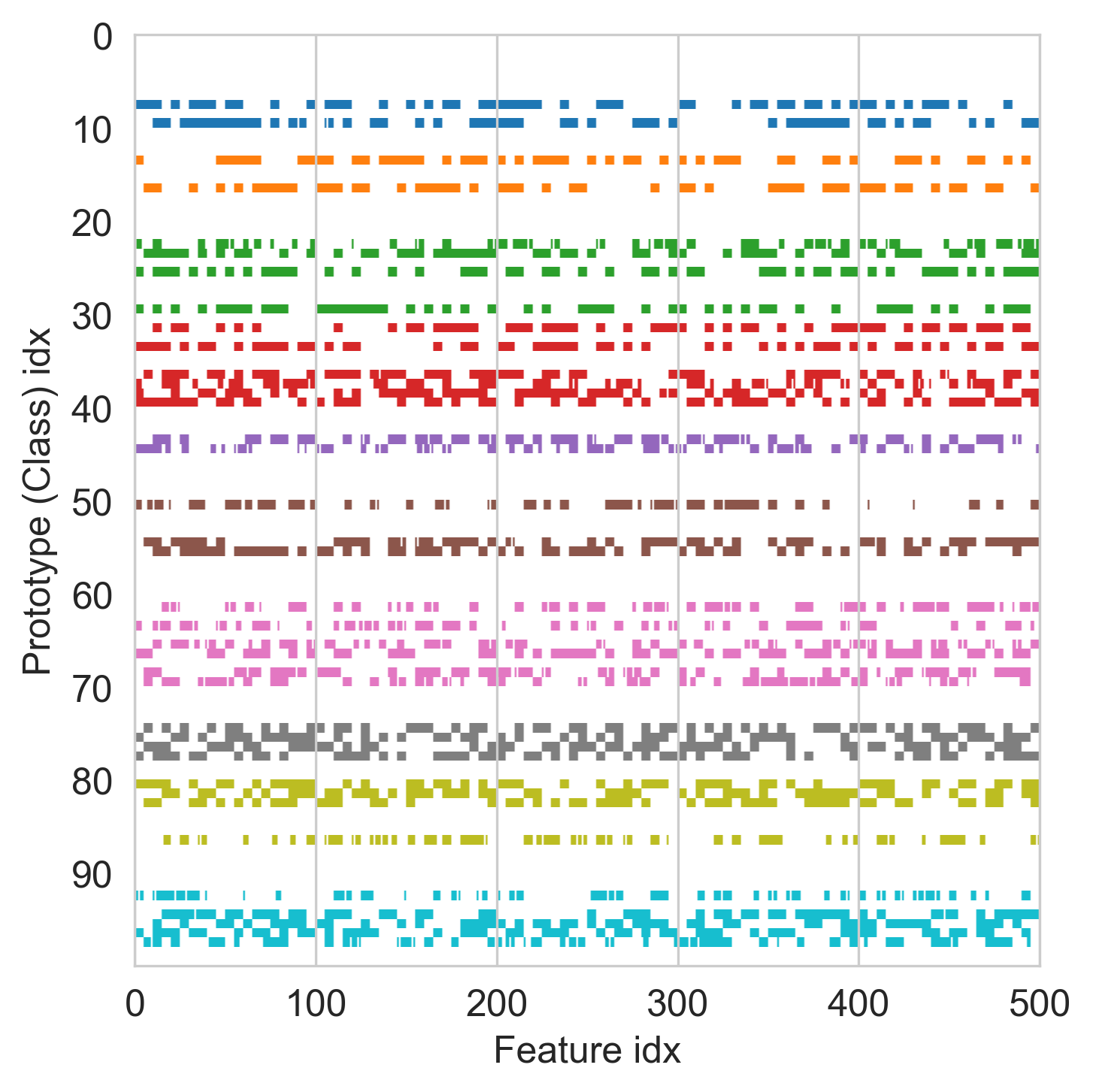}
        \caption{Prototypes of client \#8 (binary)}
    \end{subfigure}

    \begin{subfigure}{0.33\textwidth}
        \centering
        \includegraphics[width=\textwidth]{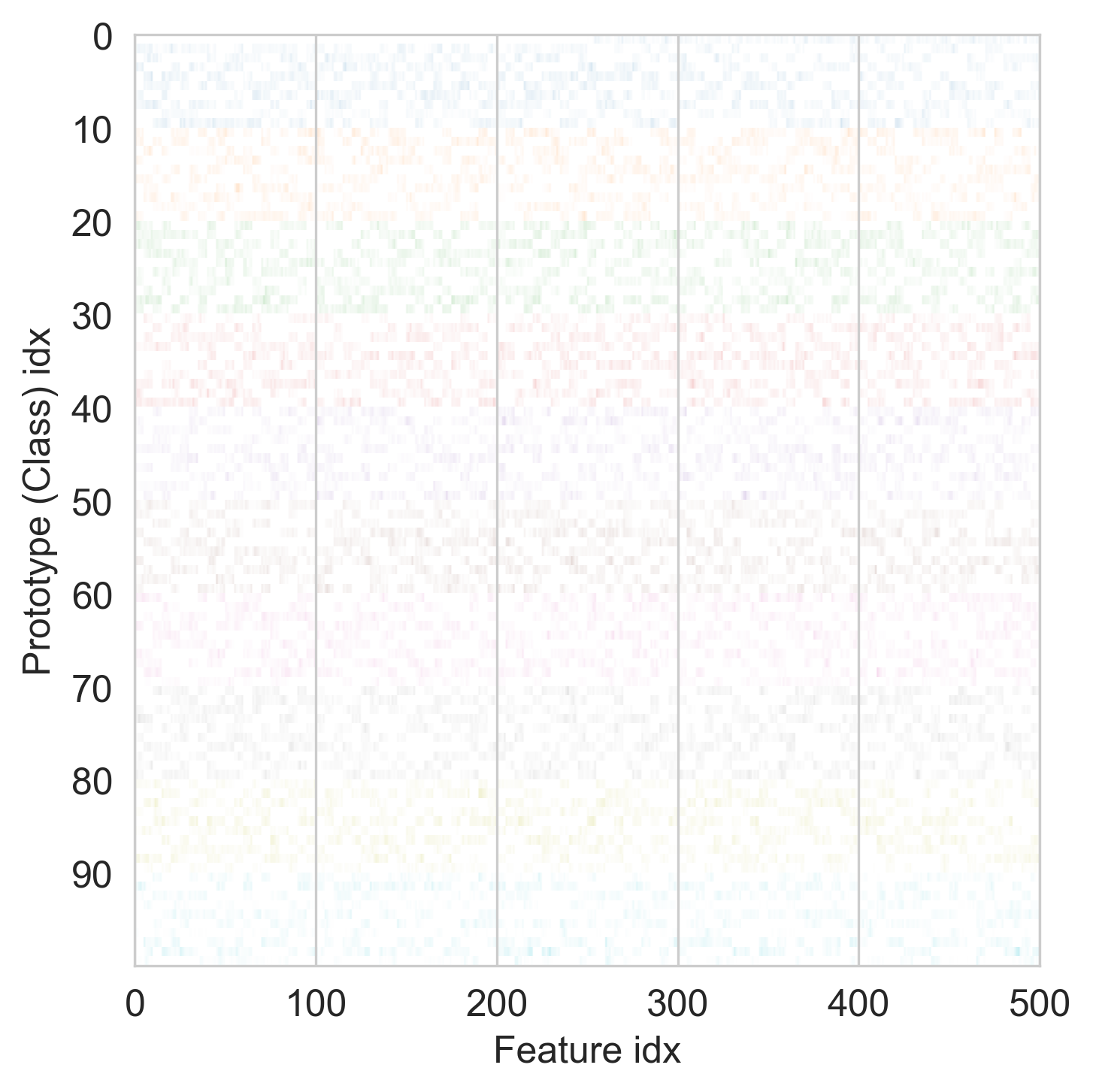}
        \caption{Global prototypes (original)}
    \end{subfigure}
    \quad
    \begin{subfigure}{0.33\textwidth}
        \centering
        \includegraphics[width=\textwidth]{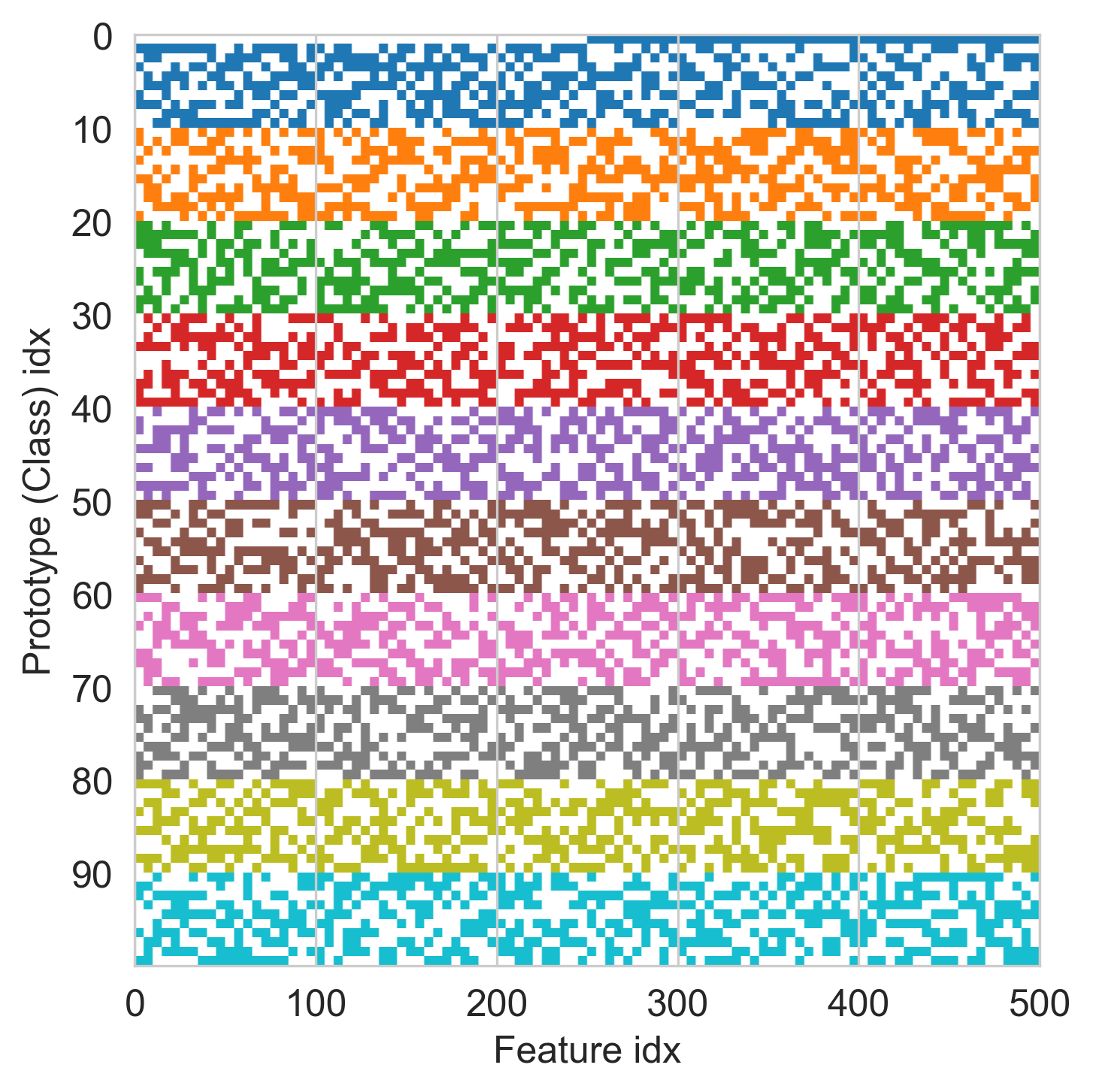}
        \caption{Global prototypes (binary)}
    \end{subfigure}
    \caption{Prototype comparison of FedProto with Class-wise Prototype Sparsification (CPS) for the CIFAR-100 dataset. The dimension $s$ is 250 for CPS.}
    \label{fig:heatmaps_prototype_cifar100_cps250}
\end{figure}

\clearpage

\section{Convergence Analysis of the CPS method}
The convergence guarantees established in Theorem 2 of FedProto can be preserved within our framework by showing that Lemma 2 of FedProto remains valid after applying Class-wise Prototype Sparsification (CPS). Specifically, CPS modifies the original framework by introducing a sparsification operator that acts on global prototypes. Therefore, we need to adapt the portions of Lemma 2 that concern global prototype operations.
To maintain consistency with FedProto's notation, we establish the correspondence $\bar{C}^{(j)} \equiv \bar{\vc}_{j}^{G}$, where $\bar{C}^{(j)}$ denotes FedProto's notation and $\bar{\vc}_{j}^{G}$ represents our global prototype for class $j$. Under this notation, we introduce the following assumption that characterizes the properties of our sparsification operator:
\begin{assumption}[Fixed Support Sparsity]
Let $S: \mathbb{R}^d \to \mathbb{R}^d$ be the sparsification operator defined in Definition \ref{df:sparse_prototype}. For each class $j$, $S$ satisfies:
\begin{enumerate}
\item[(a)] For any $\bar{C}^{(j)} \in \mathbb{R}^d$, the support of $S(\bar{C}^{(j)})$ equals a fixed index set $\Omega_j \subset \{1,\ldots,d\}$, i.e.,
$\text{supp}(S(\bar{C}^{(j)})) = \Omega_j$
\item[(b)] For any $\bar{C}^{(j)}_1, \bar{C}^{(j)}_2 \in \mathbb{R}^d$, the operator $S$ is non-expansive:
$||S(\bar{C}^{(j)}_1) - S(\bar{C}^{(j)}_2)||_2 \leq ||\bar{C}^{(j)}_1 - \bar{C}^{(j)}_2||_2$
\end{enumerate}
Here, $\text{supp}(S(\bar{C}^{(j)})) = \{i: [S(\bar{C}^{(j)})]_i \neq 0\}$ denotes the support (non-zero indices) of $S(\bar{C}^{(j)})$.
\label{as:fixed_support_sparsity}
\end{assumption}
Now, we extend Lemma 2 of FedProto by incorporating Assumption \ref{as:fixed_support_sparsity}. This assumption ensures that our sparsification operator maintains both fixed support structure and non-expansiveness properties. For notational simplicity, we omit the class label $j$ from global prototype $\bar{C}^{(j)}$ in the subsequent derivation, as this simplification does not affect the proof's validity. The modified equations that replace Equations (20)-(24) in FedProto's Lemma 2 are:
\begin{align}
    \mathcal{L}_{(t+1)E+1/2} 
        &= \mathcal{L}_{(t+1)E} + \mathcal{L}_{(t+1)E+1/2} - \mathcal{L}_{(t+1)E} \\
        &= \mathcal{L}_{(t+1)E} 
        + \lambda\|f_i(\phi_{i,(t+1)E}) - S(\bar{C}_{t+2})\|_2 
        - \lambda\|f_i(\phi_{i,(t+1)E}) - S(\bar{C}_{t+1})\|_2 \\
        &\leq \mathcal{L}_{(t+1)E} 
        + \lambda\|S(\bar{C}_{t+2}) - S(\bar{C}_{t+1})\|_2 \\
        &= \mathcal{L}_{(t+1)E} 
        + \lambda\left\|S\left(\sum_{i=1}^m q_iC_{i,(t+1)E}\right) 
        - S\left(\sum_{i=1}^m q_iC_{i,tE}\right)\right\|_2 \\
        &= \mathcal{L}_{(t+1)E} 
        + \lambda\left\|\sum_{i=1}^m q_i(C_{i,(t+1)E} - C_{i,tE})\right\|_2 
        \quad \text{(since $S$ is linear with fixed locations)}. \label{eq:assumption_operator}
\end{align}
The proof following Eq. (\ref{eq:assumption_operator}) proceeds identically to the derivations from Eq. (24) onward in FedProto, where the linearity property introduced in the final equation plays a crucial role. Although the extension is straightforward, our analysis reveals that our sparsification operator preserves the fundamental properties necessary for maintaining the original convergence bounds.

\section{TinyProto Integrated into FedTGP} \label{sec:algorithm}
The scaling method's application to FedTGP differs from FedProto. In FedTGP \citep{zhang2024fedtgp}, global prototypes are optimized through contrastive learning, which minimizes intra-class distances while maximizing inter-class distances between local and global prototypes. Direct application of scaled local prototypes $n_{i,j} \hat{\vc}_{i,j}^{L}$ would distort these distance calculations due to high variance in $n_{i,j}$ across clients and classes. To address this, we implement server-side normalization using a constant $\mu$, in contrast to FedProto's client-side scaling approach. The normalized prototypes are then reconstructed into structured sparse form. For proper feature alignment during contrastive learning, the structured sparse prototype $\tilde{\vc}_{i,j}^{L}$ is utilized instead of its compressed version.
The refined global prototypes from contrastive learning are subsequently compressed using operator $C$ to reduce communication costs. The remaining process follows the same procedure as FedProto's scaling method implementation.

\section{Experimental Details} \label{sec:experiment_details}
\subsection{Hyperparameters} 
For baseline algorithms, we adopt algorithm-specific hyperparameters as recommended in \cite{zhang2024fedtgp}. Table \ref{table:hp_settings} provides a comprehensive overview of these hyperparameter settings. It is important to note that the hyperparameter notations used in Table \ref{table:hp_settings} are specific to each baseline method and may differ from notations used elsewhere in our paper. 
\begin{table}[bht]
    \centering
    {\fontsize{9}{11}\selectfont
    \begin{tabular}{lp{0.76\linewidth}}
    \toprule
    Method      & Hyperparameter settings \\
    \midrule
    FML         & $\alpha$ (KD weight for local model) $= 0.5$, $\beta$ (KD weight for meme model) $= 0.5$ \\
    FedKD       & $T_\text{start}$ (energy threshold) $= 0.95$,  $T_\text{end}$ (energy threshold) $= 0.98$ \\
    FedDistill  & $\gamma$ (weight of logit regularizer) $= 1$ \\
    \bottomrule
    \end{tabular}
    }
    \caption{Hyperparameter settings for the compared methods.}
    \label{table:hp_settings}
    \vspace{-10pt}
\end{table}

\subsection{Experimental Environment} 
To ensure reproducibility and provide a clear understanding of our experimental environment, we detail our setup below. Our experiments were designed to rigorously test the proposed methods under controlled conditions. The following list outlines the key components of our experimental setup:

\begin{itemize}
    \item Framework: PyTorch 2.4
    \item Hardware:
    \begin{itemize}
        \item CPUs: 2 Intel Xeon Gold 6240R (96 cores total)
        \item Memory: 256GB
        \item GPUs: Two NVIDIA RTX A6000
    \end{itemize}
    \item Operating System: Ubuntu 22.04 LTS
\end{itemize}

\section{Additional Results} \label{sec:additional_results}

\begin{table*}[ht]
\centering
\setlength{\tabcolsep}{5.pt}
{\fontsize{9}{11}\selectfont
\begin{tabular}{lrcrrrrrr}
\toprule
\multirow{2}{*}{Algorithm} & \multirow{2}{*}{CPS dim.} & \multirow{2}{*}{Scaling} & \multicolumn{2}{c}{CIFAR-10} & \multicolumn{2}{c}{CIFAR-100} & \multicolumn{2}{c}{TinyImageNet} \\
\cmidrule(lr){4-5} \cmidrule(lr){6-7} \cmidrule(lr){8-9}
 & & & Acc. (\%) & Comm. (M) & Acc. (\%) & Comm. (M) & Acc. (\%) & Comm. (M) \\
\midrule
FedProto & \xmark & \xmark         & 82.90 $\pm$ 0.46 & 0.15 & 29.97 $\pm$ 0.18 & 1.46 & 13.30 $\pm$ 0.06 & 2.93 \\
TinyProto-FP   & 50  & \xmark & 84.18 $\pm$ 0.71 & 0.02 & 29.27 $\pm$ 0.28 & 0.15 & 10.02 $\pm$ 0.24 & 0.29 \\
TinyProto-FP   & 150 & \xmark & 84.24 $\pm$ 0.23 & 0.05 & 32.43 $\pm$ 0.52 & 0.44 & 14.25 $\pm$ 0.47 & 0.88 \\
TinyProto-FP   & 250 & \xmark & 84.30 $\pm$ 0.16 & 0.08 & 33.00 $\pm$ 0.28 & 0.73 & 15.70 $\pm$ 0.47 & 1.46 \\
TinyProto-FP   & 350 & \xmark & 84.07 $\pm$ 0.50 & 0.11 & 32.86 $\pm$ 0.34 & 1.02 & 16.45 $\pm$ 0.30 & 2.05 \\
TinyProto-FP   & 450 & \xmark & 83.32 $\pm$ 0.18 & 0.14 & 31.58 $\pm$ 0.42 & 1.31 & 15.48 $\pm$ 0.24 & 2.64 \\
\midrule
TinyProto-FP   & \xmark & \cmark & 83.83 $\pm$ 0.22 & 0.15 & 34.94 $\pm$ 0.17 & 1.46 & 19.00 $\pm$ 0.03 & 2.93 \\
TinyProto-FP   & 50  & \cmark    & 84.52 $\pm$ 0.06 & 0.02 & 31.82 $\pm$ 0.24 & 0.15 & 16.01 $\pm$ 0.21 & 0.29 \\
TinyProto-FP   & 150 & \cmark    & 85.14 $\pm$ 0.17 & 0.05 & 34.87 $\pm$ 0.29 & 0.44 & 18.34 $\pm$ 0.25 & 0.88 \\
TinyProto-FP   & 250 & \cmark    & 85.07 $\pm$ 0.14 & 0.08 & 35.61 $\pm$ 0.10 & 0.73 & 18.74 $\pm$ 0.20 & 1.46 \\
TinyProto-FP   & 350 & \cmark    & 85.43 $\pm$ 0.24 & 0.11 & 35.25 $\pm$ 0.31 & 1.02 & 18.83 $\pm$ 0.10 & 2.05 \\
TinyProto-FP   & 450 & \cmark    & 85.26 $\pm$ 0.17 & 0.14 & 35.13 $\pm$ 0.16 & 1.31 & 19.07 $\pm$ 0.23 & 2.64 \\
\midrule
FedTGP & \xmark & \xmark           & 86.32 $\pm$ 0.49 & 0.15 & 36.92 $\pm$ 0.16 & 1.46 & 19.44 $\pm$ 0.12 & 2.93 \\
TinyProto-FT   & 50 & \xmark  & 85.91 $\pm$ 0.23 & 0.02 & 35.26 $\pm$ 0.25 & 0.15 & 18.51 $\pm$ 0.11 & 0.29 \\
TinyProto-FT   & 150 & \xmark & 86.05 $\pm$ 0.20 & 0.05 & 36.05 $\pm$ 0.06 & 0.44 & 18.68 $\pm$ 0.19 & 0.88 \\
TinyProto-FT   & 250 & \xmark & 86.03 $\pm$ 0.16 & 0.08 & 35.74 $\pm$ 0.69 & 0.73 & 18.74 $\pm$ 0.58 & 1.46 \\
TinyProto-FT   & 350 & \xmark & 86.27 $\pm$ 0.09 & 0.11 & 35.57 $\pm$ 0.28 & 1.02 & 18.83 $\pm$ 0.15 & 2.05 \\
TinyProto-FT   & 450 & \xmark & 85.93 $\pm$ 0.49 & 0.14 & 35.07 $\pm$ 0.25 & 1.31 & 16.45 $\pm$ 0.02 & 2.64 \\
\midrule
TinyProto-FT   & \xmark & \cmark & 88.40 $\pm$ 0.11 & 0.15 & 46.10 $\pm$ 0.24 & 1.46 & 26.82 $\pm$ 0.12 & 2.93 \\
TinyProto-FT   & 50 & \cmark     & 88.47 $\pm$ 0.21 & 0.02 & 45.94 $\pm$ 0.40 & 0.15 & 27.29 $\pm$ 0.21 & 0.29 \\
TinyProto-FT   & 150 & \cmark    & 88.56 $\pm$ 0.06 & 0.05 & 46.34 $\pm$ 0.05 & 0.44 & 27.41 $\pm$ 0.10 & 0.88 \\
TinyProto-FT   & 250 & \cmark    & 88.35 $\pm$ 0.27 & 0.08 & 46.36 $\pm$ 0.25 & 0.73 & 25.90 $\pm$ 0.09 & 1.46 \\
TinyProto-FT   & 350 & \cmark    & 88.55 $\pm$ 0.06 & 0.11 & 46.53 $\pm$ 0.17 & 1.02 & 25.59 $\pm$ 0.09 & 2.05 \\
TinyProto-FT   & 450 & \cmark    & 88.21 $\pm$ 0.21 & 0.14 & 45.59 $\pm$ 0.09 & 1.31 & 25.54 $\pm$ 0.04 & 2.64 \\
\bottomrule
\end{tabular}
}
\caption{Classification test accuracy (Acc.) and communication cost (Comm.) across datasets. The CPS dim. column shows the compressed prototype dimension $s$. Communication cost is measured by the number of parameters shared per FL round (M: millions). TinyProto-FP and TinyProto-FT denote TinyProto integrated into FedProto and FedTGP, respectively.}
\label{table:supp_total_result}
\end{table*}

\end{document}